\documentclass[hidelinks]{article} 
\usepackage{iclr2027_conference,times}

\makeatletter
\newcommand{\iclrpreprint}{%
    \iclrfinalcopy
    \let\iclrpreprint@originallhead\lhead
    \def\lhead##1{%
        \iclrpreprint@originallhead
        {Under review as a conference paper at ICLR 2027}%
    }%
}
\makeatother

\usepackage{amsmath,amsfonts,bm}

\def\eqref#1{equation~\ref{#1}}

\def\1{\bm{1}}

\DeclareMathAlphabet{\mathsfit}{\encodingdefault}{\sfdefault}{m}{sl}
\SetMathAlphabet{\mathsfit}{bold}{\encodingdefault}{\sfdefault}{bx}{n}

\usepackage{amsmath, amssymb, amsthm}
\usepackage{algorithm}
\usepackage{algpseudocode}
\usepackage{subcaption}
\usepackage{hyperref}
\usepackage{url}
\usepackage{mathtools}
\usepackage[dvipsnames]{xcolor}
\usepackage{booktabs}
\usepackage{float}
\usepackage{tipa}

\usepackage{cleveref}
\Crefname{section}{Section}{Sections}
\crefname{section}{Section}{Sections}
\Crefname{appsec}{Appendix}{Appendices}
\crefname{appsec}{Appendix}{Appendices}
\Crefname{figure}{Figure}{Figures}
\crefname{figure}{Fig.}{Figs.}

\usepackage[inline,shortlabels]{enumitem}
\usepackage{multirow}

\definecolor{ibmmagenta}{HTML}{DC267F}
\definecolor{ibmblue}{HTML}{648FFF}

\usepackage{colortbl}

\newif\ifhighlights
\highlightstrue         

\theoremstyle{plain}

\theoremstyle{definition}

\usepackage{tcolorbox}

\title{Online Class Posteriors for Interpretable and Controllable Discrete Diffusion}

\author{Felix Koulischer*, Gianni Van de Velde*, Francois Remy, Chris Develder, Thomas Demeester}

\author{%
  \begin{tabular}[t]{c} 
    \hspace{1.35cm} Felix Koulischer\textsuperscript{1,*} \hspace{1cm} Gianni Van De Velde\textsuperscript{1, *} \hspace{1cm} François Remy \textsuperscript{2}    \\ \hspace{1.35cm} Chris Develder \textsuperscript{1} \hspace{1cm} Thomas Demeester\textsuperscript{1} \\
  \end{tabular}
  \\
  \\
  \begin{tabular}[t]{c} 
    \hspace{2.25cm} \textsuperscript{1} Ghent University - imec\hspace{0.5cm} \hspace{1.25cm}\textsuperscript{2} Parallia Healthcare
  \end{tabular}
  \\
}

\iclrpreprint
\begin{document}

\maketitle

\renewcommand{\thefootnote}{}
\footnotetext{\textsuperscript{*} Joint First Authors: \textit{felix.koulischer@ugent.be}, \textit{gianni.vandevelde@ugent.be}}
\renewcommand{\thefootnote}{\arabic{footnote}}

\begin{abstract}
Discrete diffusion models remain far less understood than their continuous counterparts, despite recent breakthroughs across modalities, most notably in text.
Leveraging tools from continuous diffusion, we formulate a theoretical framework to track the class-conditional posterior in an online manner by combining conditional and unconditional model predictions. 
In masked language modeling, this posterior offers highly interpretable dynamics: tokens aligned with a class boost the posterior, while contradictory ones suppress it. We also show that the posterior can 
be used for logical guidance operations (\textsc{and}, \textsc{or}, \textsc{not}) in discrete diffusion. These operations enable precise control over generation, such as avoiding specific classes. 
In addition, we discuss 
a mismatch in the formulation of classifier-free guidance between discrete and continuous settings: 
discrete models mix the data distributions, while continuous models mix the marginals. 
We validate our methods and discuss their potential applications on synthetic data, with all quantities analytically tractable, as well as on MNIST, text, and antibody sequences.
%
%
\end{abstract}

\section{Introduction}

Diffusion models are the default generative paradigm for continuous data, from images to protein structures \citep{SohlDicksteinDMs, DDPM, StableDiffusion, AF3}. They learn the so-called score function $\nabla \log p_t(x_t)$, which gradually transports noise towards data.
Discrete Diffusion Models (DDMs) transfer this idea by operating directly on the discrete state space \citep{D3PM,MD4,CTMCDiscreteDiff}, as an alternative to embedding tokens into a continuous latent space and reusing continuous machinery \citep{dieleman2022continuous}. Like their continuous counterparts, DDMs progressively corrupt data, 
either by masking tokens or by replacing them with tokens drawn uniformly from the vocabulary.
In the discrete setting, the score function is replaced by a rate function $R_t(x,y)$, the instantaneous probability of token $y$ transitioning to token $x$. Integrating this rate over diffusion time defines a continuous time Markov chain. This process, like in the continuous case, carries the state from noise to data~\citep{CTMCDiscreteDiff, DFM}.

Emerging evidence in language and DNA modeling suggests that DDMs could rival autoregressive models, thanks to their bidirectional context and multi-token generation \citep{LLaDa, DFMforDNA, ScalingBeyhondMasked}. 
Yet, while the literature on continuous diffusion models has extensively explored noise schedules \citep{EDM, EntropicTimeSchedulers}, training dynamics \citep{EDM2}, and inference-time guidance \citep{CG, CFG}, the algorithmic tools for discrete diffusion, and in particular for guiding it, remain far less developed.
We address this gap by transferring to the discrete setting two guidance tools that have no discrete counterpart yet: the online tracking of the class posterior during generation, and the composition of conditional generative processes through logical operators.

Classifier-free guidance (CFG) \citep{CG,CFG} 
applies a constant push towards a condition throughout the denoising process, regardless of whether the considered sample needs it. Recent work in continuous diffusion has shown that guidance can instead adapt to the state of the trajectory, only being used if a trajectory truly requires it \citep{DNG,FBG,LearnToGuideDM}. 
Additionally, guidance can be used for composing several conditional generative processes, for instance to generate samples satisfying either or both of two conditions. This requires weighting each process according to how likely the current state is to satisfy its corresponding condition~\citep{ItoOrGate,LogicalGuidCont}. All of these constructions rely on the same quantity, the class posterior $p_t(c\mid x_t)$, which estimates the probability that the noisy state $x_t$ will eventually satisfy condition $c$. In this work, we focus on how conditional and unconditional model predictions, and their corresponding rate functions, can be combined to estimate this posterior and construct different conditional generative processes.

We show that the class posterior can be tracked during discrete diffusion directly from the conditional and unconditional predictions, for both uniform and masking noise. 
The estimation is exact when analytical quantities are available and exhibits interpretable token-level dynamics: because the state changes one token at a time, transitions that support a class increase its posterior, while transitions supporting a different class decrease it. This provides a direct interpretation of the posterior that has no continuous analogue. In masked language modeling, for example, it reveals which generated tokens move a sample towards or away from a class.

Building on this posterior estimation, we extend CFG to the logical \textsc{and}, \textsc{or}, and \textsc{not} operators for discrete rate functions, drawing on recent advances in the continuous domain \citep{LogicalGuidCont,ItoOrGate,DNG}. These operators combine conditional rate functions through posterior-dependent weighting: the \textsc{or} gate arbitrates between the conditions according to their posterior probabilities, while the \textsc{not} gate detects the undesired condition and activates a repulsive term only when necessary. We verify these constructions analytically and validate them on discrete image and antibody sequence generation. Our derivation also reveals that standard discrete CFG \citep{SimpleGuidMechDDMs,NisonoffGuidDDMs} differs from its continuous counterpart in a way that has gone unnoticed. Continuous CFG mixes marginals multiplicatively, which is realized as an additive mixture of predictions, whereas discrete CFG instead mixes data distributions multiplicatively. These operations are not equivalent because the noising process does not commute with multiplicative mixing~\citep{GuidIsPredCorr}.

Our contributions are threefold:
\begin{enumerate*}[(i)]
\item we cast the online class posterior to the discrete domain and demonstrate its interpretable dynamics in language generation;
\item we derive posterior-weighted \textsc{or} and \textsc{not} operators for discrete rate functions, verify these on an analytical example, and validate on discrete image generation and antibody sequence generation; and 
\item we identify and formalize the multiplicative/additive mismatch between continuous and discrete CFG.
\end{enumerate*}
\begin{figure}[t]
    \centering
    \begin{subfigure}[t]{0.48\textwidth}
        \centering
        \includegraphics[width=\linewidth]{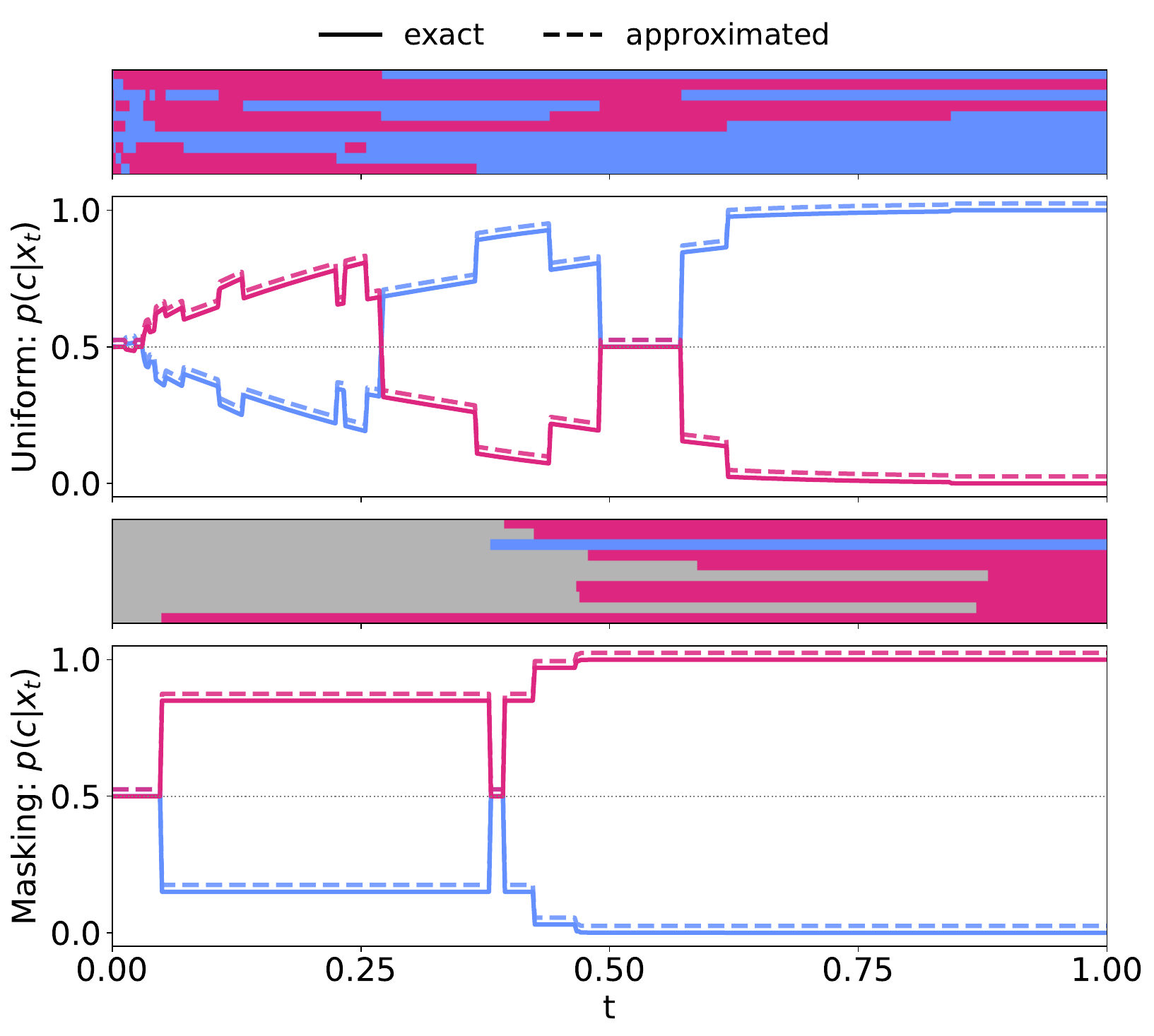}
        \caption{Synthetic data}
        \label{subfig: Illustrative posterior estimations analytical}
    \end{subfigure}
    \hfill
    \begin{subfigure}[t]{0.48\textwidth}
        \centering
        \includegraphics[width=\linewidth]{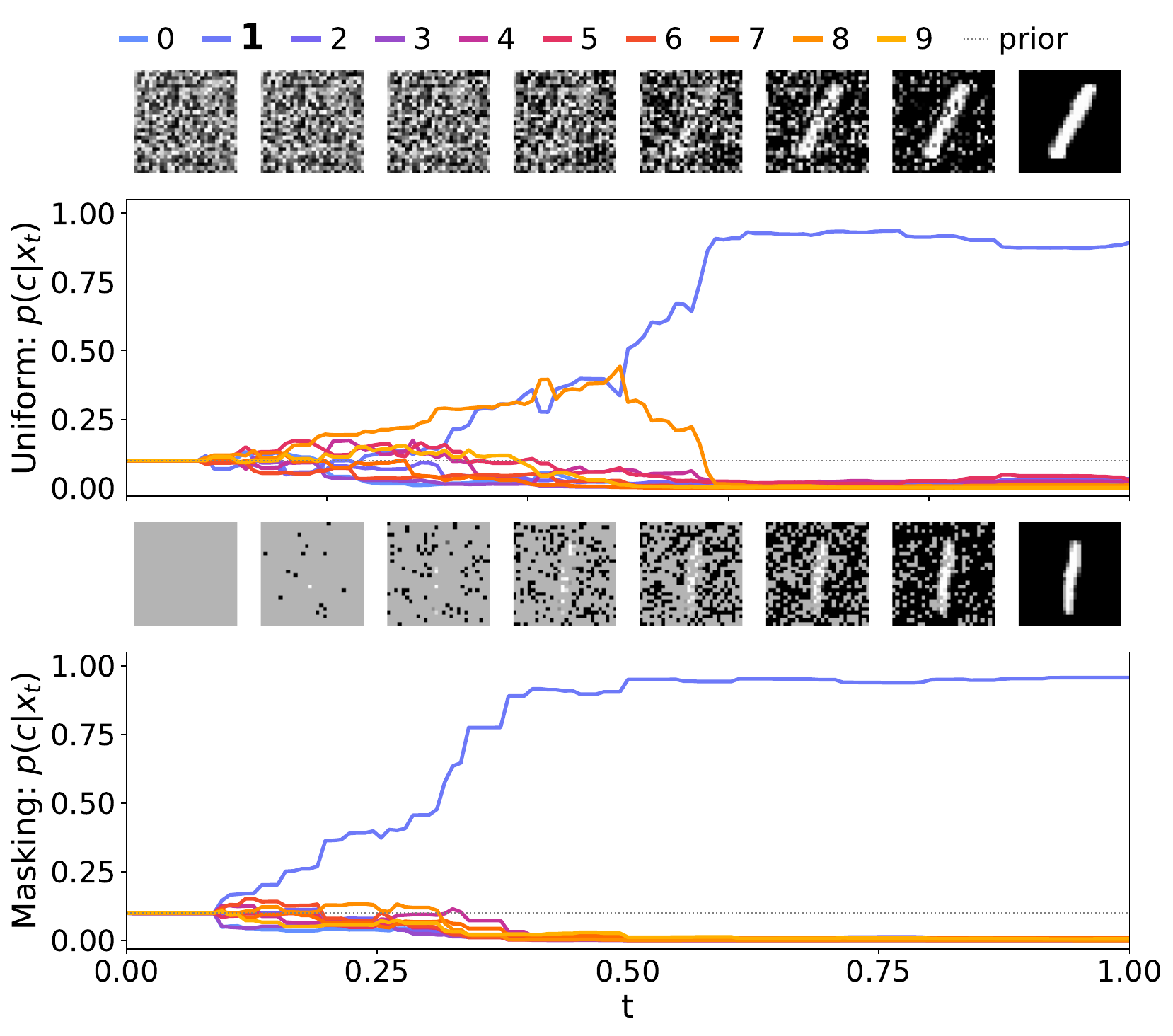}
        \caption{MNIST}
        \label{subfig: Illustrative posterior estimations MNIST}
    \end{subfigure}
    \caption{Illustration of the online posterior evaluation. 
    In (a), evaluated on an analytically tractable setting in which data points consist of two tokens represented by blue and red with each class having an 80\% majority of one of the two tokens, trajectories are shown above the posteriors. 
    In (b), the posterior is evaluated on MNIST digit generation. On top for uniform, below for masked diffusion. The increase in the posterior occurs simultaneously with the emergence of the digit.} 
    \label{fig: Illustrative posterior approx}
\end{figure}

\section{Background and Related Work}
\label{sec: Background and related work}
\subsection{Discrete Diffusion Models}
\label{subsec: Discrete Diffusion Basics}
We consider sequences $\mathbf{x}_t=(x_{t}^1,\ldots,x_{t}^L)\in\mathcal{V}^L$ of $L$ discrete tokens, with $t\in[0,1]$ running from noise ($t=0$) to data ($t=1$). Let $q_t(x_t\mid x_1)$ denote the forward noising kernel for a single token. We consider the mixture path
\begin{equation}
q_t(x_t\mid x_1)=\kappa(t)\delta_{x_1}(x_t)+\bigl(1-\kappa(t)\bigr)\sigma(x_t),\qquad \kappa(0)=0,\quad \kappa(1)=1,
\end{equation}
where $\sigma$ is a base noise distribution. The corresponding sequence-level corruption process factorizes over tokens, with classical choices including masking, in which tokens converge to an absorbing mask token, and uniform noising, in which tokens are replaced by samples from a uniform distribution \citep{D3PM}.

The forward process defines a continuous-time Markov chain (CTMC) with marginal distribution $p_t$. Let $R_t(x,y)$ denote the instantaneous rate of transitioning from state $y$ to state $x$. 
With this convention, the marginal distribution evolves according to the master equation
\begin{equation}
\frac{\mathrm{d}p_t(x)}{\mathrm{d}t} = \sum_{y\in\mathcal{V}}R_t(x,y)p_t(y) \equiv R_t p_t
\label{eq:ctmc}
\end{equation}
with standard notation for the rate matrix $R_t$.
Equivalently, for a small time step $h>0$, the transition kernel, which can be used to construct a sampler, is given by
\begin{equation}
P_{t,t+h}(x\mid y) =\delta_{xy}+hR_t(x,y)+o(h),
\end{equation}
Typically, a neural network is trained to approximate the denoiser
$p_{\theta}(x_1\mid x_t)$ using a time-weighted cross-entropy loss \citep{MD4}. This posterior can subsequently be converted into an instantaneous rate function. Importantly, the rate function is not unique: multiple generators can induce the same prescribed marginal path. One such construction is Discrete Flow Matching \citep{DFMKinOpt}, which, for the marginal path above, defines;
\begin{equation}
R_{\theta,t}(x,y) =
\frac{\dot{\kappa}(t)}{1-\kappa(t)}\left[p_{\theta}(x_1=x\mid x_t=y)-\delta_{xy}\right].
\label{eq: velocity fields}
\end{equation}
A different valid construction is given by the Markovian ancestral sampler \citep{D3PM}.

In the language domain, Discrete Diffusion Models, or Diffusion Language Models (DLMs), already come with a range of decoding strategies for improving generation. 
LLaDA \citep{LLaDa}, a commonly used DLM, uses a fixed decoding schedule and confidence unmasking. The schedule fixes how many tokens are unmasked at each diffusion step, while confidence unmasking determines which ones: rather than choosing at random, it unmasks the tokens with the highest confidence. 

\subsection{Diffusion Guidance}
\label{subsec: Diffusion Guidance}

\textbf{Guidance in continuous diffusion models:} Sampling 
on a class label or prompt $c$ alone often does not yield satisfactory samples. In continuous diffusion models, guidance corrects this by tilting the noisy marginal with an exponentiated classifier term \citep{CG},
\begin{equation}
    \tilde{p}_t(x_t|c) \propto p_{t,\theta}(x_t)\, p_{t,\theta}(c|x_t)^\lambda ,
\end{equation}
where $\lambda>1$ controls the strength of the conditioning signal. Rewriting $p_{t,\theta}(c|x_t)$ via Bayes' rule replaces the classifier with a ratio of a conditional and an unconditional model, $\tilde{p}_t(x_t|c) \propto p_{t,\theta}(x_t)^{1-\lambda}\,p_{t,\theta}(x_t|c)^{\lambda}$, so that both terms can be obtained from a single jointly-trained network \citep{CFG}. Taking the gradient of the log of the 
tilted marginal gives the familiar rules of Classifier Guidance and Classifier-Free Guidance, respectively: 
\begin{align}
    \nabla \log \tilde{p}_t(x_t|c) & = \nabla \log p_t(x_t) + \lambda \nabla \log p_t(c|x_t) \\
     & = \nabla \log p_t(x_t) + \lambda \big(\nabla \log p_t(x_t|c)-\nabla \log p_t(x_t)\big) ,
\end{align}
an \emph{additive} combination of the conditional and unconditional score predictions.

\textbf{Guidance in discrete diffusion models:} The discrete setting admits a similar decomposition, although not equivalent. 
Guidance was introduced for discrete diffusion with the same goal of sharpening an under-confident conditional signal \citep{NisonoffGuidDDMs, SimpleGuidMechDDMs}, realized by combining the logits of the predictions instead of the predictions themselves:
\begin{equation}
    \log \tilde{p}(x_1|x_t,c) = \log p(x_1|x_t) + \lambda \big(\log p(x_1|x_t,c)-\log p(x_1|x_t)\big) ,
    \label{eq: discrete CFG}
\end{equation}
a \emph{multiplicative} mixture of the predicted token distributions, in contrast to the additive mixture of scores used above. 

\textbf{Recent progress on guidance:} Both formulations have since been refined. 
In the continuous case it was proposed to schedule the guidance scale $\lambda$ over time rather than hold it fixed, aiming to 
target the noise levels that benefit most from it \citep{LIG, FBG, LearnToGuideDM}. Recent work has also shown that deliberately weakening the unconditional model pushes samples away from low-quality regions of the data space as well as from unconditionality~\citep{Autoguid}. Analogous ideas, including 
time-dependent schedules, have since begun to be carried over to the discrete formulation as well \citep{GuidSchedulesDiscreteDiff}. More recently, guidance has been reframed as one instance of a broader family of operators for combining diffusion processes, including \textsc{and} (both conditions), \textsc{or} (either condition) and \textsc{not} (avoid a condition) \citep{ItoOrGate, DNG, LogicalGuidCont}. The bottleneck of the latter two is the requirement of a class-conditional posterior $p_t(c|x_t)$ needed to weight this combination, which can be trained separately \citep{LogicalGuidCont} or approximated from existing conditional/unconditional models \citep{DNG, ItoOrGate}. Discrete diffusion, however, has lacked both ingredients: the logical operators and a posterior to weight them. In this work, we close this gap, formulating the full family of operators for the discrete setting and showing that the posterior can be tracked during inference with high interpretability. 

\section{Posterior Estimation and Combining Multiple 
Processes}
\label{sec: Theory}

This section describes our key methodological contributions for discrete diffusion models.
\cref{subsec: online posterior estimation} shows how the class-conditional posterior can be estimated during inference by combining conditional and unconditional model predictions.
Building on this, \cref{subsec: logical operators} introduces the three logical operators \textsc{and}, \textsc{or} and \textsc{not}, beginning with the notation they require and an important note on the \mbox{(non-)commutativity} of mixing and noising operations, which, for the \textsc{and} operator, reveals that discrete and continuous CFG are not equivalent~\citep{GuidIsPredCorr}.

\subsection{Online Posterior Estimation}
\label{subsec: online posterior estimation}

The conditional posterior $p_t(c|x_t)$, which evaluates how likely a noisy state $x_t$ ends up satisfying the condition $c$, is often mistakenly considered intractable during inference. 
In the continuous case, its tractability when sampling using a reversible Markov chain has been demonstrated~\citep{DNG,ItoOrGate}, but to the best of our knowledge, this has not been explored for 
discrete diffusion models. 
We here present how $p(c\mid x_t)$ can be tracked iteratively through inference and what specifics arise in the discrete case as opposed to the continuous case.
The idea is to track the posterior $p(c\mid x_{0:t})$ throughout inference using the conditional and unconditional models' predictions, with $x_{0:t}$ a shorthand notation for all states leading up to $t$:
\begin{align}
    \log p(c\mid x_{0:t}) = \log p(c\mid x_{0:t-1}) + \log p(x_t\mid x_{t-1},c) - \log p(x_t\mid x_{t-1}). \label{eq:iter_post}
\end{align}
In practice, we follow \citet{DNG} and introduce two hyperparameters $\tau$ and $\delta$:
\begin{align}
    \log p(c\mid x_{0:t}) = \log p(c\mid x_{0:t-1}) + \tau \big(\log p(x_t\mid x_{t-1},c) - \log p(x_t\mid x_{t-1})\big) + \delta. \label{eq:iter_post_with_hyperparam}
\end{align}
If a Markovian sampler is used, the current state is a sufficient statistic for the class: $p(c|x_t) = p(c\mid x_{0:t})$. 
In the masking case, this identity always holds since a token never changes once unmasked and $x_t$ therefore retains every revealed token.
In the uniform case, the Markov property only holds when sampling using an ancestral sampler specifically constructed to be Markovian.
However, in practice we find that using the transition kernel predicted by a discrete flow matching sampler is more stable, which is why we use it in the results section.\footnote{For more detail, see  \cref{app: theory - posterior history dependence,app: synth data - posterior validation}.}

A key advantage of the discrete case, compared to its continuous analogue, is that one can see which token flips caused the posterior to change, therefore allowing the user to understand what might have caused the model to opt for one class or another.
This is particularly interesting in masked language modeling, where the posterior reveals which words commit the sample to a class, long before denoising completes.

\subsection{Logical operators: \textsc{and}, \textsc{or}, \textsc{not}}
\label{subsec: logical operators}

We now show how logical mixture operators $\Lambda \in\{\textsc{and}, \textsc{or}, \textsc{not}\}$ over conditional distributions $\{p(x|c_i)\}_{i=1}^N$, defined by their rate matrices $R_t^{i}$ or predictors $p^i(x_1|x_t)$, 
combine into a joint rate $R^{\Lambda}_t$ or predictor $p^{\Lambda}(x_1|x_t)$. 
Crucially, an operator can be applied at two levels: 
\begin{enumerate*}[(i)]
\item before noising, thus mixing the data distributions (which we show is what discrete CFG does \citep{SimpleGuidMechDDMs}), or\label{it:before-noising}
\item after, which amounts to mixing the marginals (as in continuous CFG \citep{CFG}).\label{it:after-noising}
\end{enumerate*}
Because the noising convolution commutes with additive but not multiplicative mixing, the two levels \ref{it:before-noising}--\ref{it:after-noising} coincide for the additive \textsc{or} and \textsc{not} operators, but differ for the multiplicative \textsc{and}.
Hence, we provide separate derivations for \textsc{and}, but jointly present them for \textsc{or} and \textsc{not} (see \cref{app: theory - C vs D-CFG} for a detailed treatment).

\paragraph{\textsc{and} gate. \textit{(data-level mixture)}}
A data-level mixture targets the mixture distribution exactly.
For \textsc{and}, this is the multiplicative mixture $p^{\Lambda}(x_1|x_t)\propto \prod_{i=1}^N p^i(x_1|c_i)$, with as predictor:
\begin{align}
    p^{\Lambda}(x_1|x_t) &= p(x_t|x_1) \frac{p^{\Lambda}_1(x_1)}{p^{\Lambda}_t(x_t)}\propto p(x_t|x_1) \prod_{i=1}^N p^i(x_1|c_i)\\
    &\propto p(x_t|x_1)^{1-N}\prod_{i=1}^N p^i(x_1|x_t,c) 
\end{align}
Thus the predicted distributions should be multiplied with each other, which is commonly implemented as adding logits together:
\begin{align}
    \log p(x_1|x_t) &\propto \sum_{i=1}^N \log p^i(x_1|x_t,c) \label{eq:and data level}
\end{align}
From this predictor a rate function can be constructed using \cref{eq: velocity fields}.
This sum of logits strongly resembles how CFG is implemented in the discrete diffusion literature
, i.e., through \cref{eq: discrete CFG}. 

\paragraph{\textsc{and} gate. \textit{(marginal-level mixture)}}
On the other hand, an approach reminiscent of the continuous diffusion literature can be applied: 
considering a multiplicative mixture of the marginals defined as $p^{\Lambda}_t(x_t) \propto \prod_{i=1}^N p_t(x_t|c_i)$.\footnote{Note the change in subscript from $_1$ to $_t$.}
The joint rate function can be directly derived through:
\begin{align}
    \frac{\mathrm{d}p^{\Lambda}_t(x_t)}{\mathrm{d}t} & = \sum_{i}\frac{\mathrm{d}p_t(x_t|c_i)}{\mathrm{d}t}\prod_{j\neq i}^N p(x_t|c_j)= \Big(\sum_{i}R_t^i\Big)p^{\Lambda}_t(x_t) 
\label{eq: AND gate derivation}
\end{align}
Hence, the rate matrices of the joint process satisfy:
\begin{equation}
    R^{\Lambda}_t = \sum_{i=1}^N R^i_t. \label{eq:and marginal level}
\end{equation}
The shape of this combination is also intuitive: as long as one of the conditions is not satisfied, the corresponding rate will try to strongly modify the outcome.
Only once $x_t$ satisfies both conditions will both rate functions remain reasonably small. 
This is analogous to continuous CFG, which similarly combines two infinitesimal generators, with the score functions being replaced by the discrete rate functions.

\paragraph{\textsc{or} gate. \textit{(both)}}
For the additive operation behind the \textsc{or} gate, $p^{\Lambda}_{t}(x_t) \equiv p_t(x_t) = \sum_{i=1}^N p(c_i)p_t(x_t|c_i)$,
with mixture weights given by the priors $p(c_i)$, $p_t^\Lambda$ is simply the unconditional marginal induced by the partition ${c_i}_{i=1}^N$. The joint rate function can again be computed directly:
\begin{align}
    \frac{\mathrm{d}p^{\Lambda}_t(x_t)}{\mathrm{d}t} &= \sum_{i} p(c_i)\>  \frac{\mathrm{d}p_t(x_t|c_i)}{\mathrm{d}t} = \left( \sum_{i}p(c_i) \> \frac{p_t(x_t|c_i)}{p^{\Lambda}(x_t)} \> R^i_t \right) \> p^{\Lambda}_t(x_t) 
\label{eq: OR gate derivation1}
\end{align}
Recognizing the resulting Bayesian posterior gives:
\begin{equation}
    R_t^{\Lambda} = \sum_{i=1}^N p(c_i\mid x_t) \> R_t^{i}. \label{eq:or}
\end{equation}
Thus, the global rate is obtained by averaging the conditional rates according to the posterior probability of each component given the current noisy state.

This perspective closely mirrors the construction underlying Flow Matching \citep{FM,DFM}: rather than learning the global dynamics directly, one can define simple conditional dynamics and learn how to identify their contribution from the current state.
The global dynamics are then obtained by posterior-weighted averaging of the conditional dynamics.
In particular, taking each datapoint as a separate condition yields datapoint-specific rates that can be combined through $p(c_i| x_t)$ to recover the global rate.
Our construction therefore extends this conditional-to-global principle from datapoint-level conditioning to arbitrary partitions of the data.

Beyond this theoretical connection, the \textsc{or} gate is not strictly necessary for class-conditional generation, since sampling a condition according to its prior $c_i\sim p(c)$, and subsequently using the corresponding conditional model, yields the same desired distribution.
\paragraph{\textsc{not} gate. \textit{(both)}}
     Removing, or avoiding, a class from generation can be expressed as subtracting its conditional distribution from the unconditional one, $p^{\Lambda}_t(x_t) \propto p_t(x_t) - p(c)\,p_t(x_t|c)$. The joint rate is then given by:
\begin{align}
    \frac{\mathrm{d}p^{\Lambda}_t(x_t)}{\mathrm{d}t} &= \frac{R_t \> p_t(x_t) - p(c) \> R^c_t \> p_t(x_t|c)}{1-p(c)}  = \left( R_t - \frac{p_t(c|x_t)}{1-p_t(c|x_t)} \> \left( R_t^{c} -R_t \right) \right) \> p^{\Lambda}_t(x_t)
\label{eq: NOT gate derivation}
\end{align}
And therefore:
\begin{equation}
    R_t^{\Lambda} = R_t - \frac{p(c\mid x_t)}{1-p(c\mid x_t)}\bigl(R_t^c - R_t\bigr). \label{eq:not}
\end{equation}
This equation resembles a classifier-free guidance equation, in which $\frac{p(c \mid x_t)}{1-p(c\mid x_t)}$ can be seen as a \textit{dynamic} guidance scale that is small if the undesired class is unlikely to be generated and that becomes asymptotically large if that class becomes likely. 
Again, this is entirely analogous to what is found in the continuous case, with as single difference that the continuous generators, or score-functions, are replaced by their discrete analogues, the rate functions \citep{DNG}. 
\section{Results}
\label{se:results}
\Cref{sec:posterior_estimation} first validates {the posterior approximation} on 
\begin{enumerate*}[(i)]
\item synthetic data for which ground-truth quantities are available before being tested on MNIST images \citep{MNIST}, 
\item text, and
\item antibody sequence generation \citep{IgLM}.
\end{enumerate*}
Next, in \cref{sec:not_gate}, we evaluate {the \textsc{not} gate} for image and antibody sequence generation. 
\subsection{Online Posterior Estimation}
\label{sec:posterior_estimation}
\paragraph{Synthetic data.}
Our {synthetic dataset} consists of binary sequences of length 10, with red and blue tokens.
Each sequence is assigned to one of two classes based on a majority token: one token (red or blue) is sampled with probability $p_{\text{maj}}$, and the other with $1-p_{\text{maj}}$.
\Cref{subfig: Illustrative posterior estimations analytical} shows that the inferred posterior matches the analytically tractable ground-truth quantity computed using Bayes' rule with the known marginal likelihoods.
Importantly, the posterior inference remains exact across both masking and uniform noise processes,\footnote{Uniform noise requires the Markovian ancestral transition kernel. Details are provided in \cref{app: synth data - posterior validation}.}
despite their substantially different transition dynamics.

\begin{figure}[t]
    \centering
    \begin{subfigure}[b]{0.55\textwidth}
        \centering
        \includegraphics[width=\linewidth]{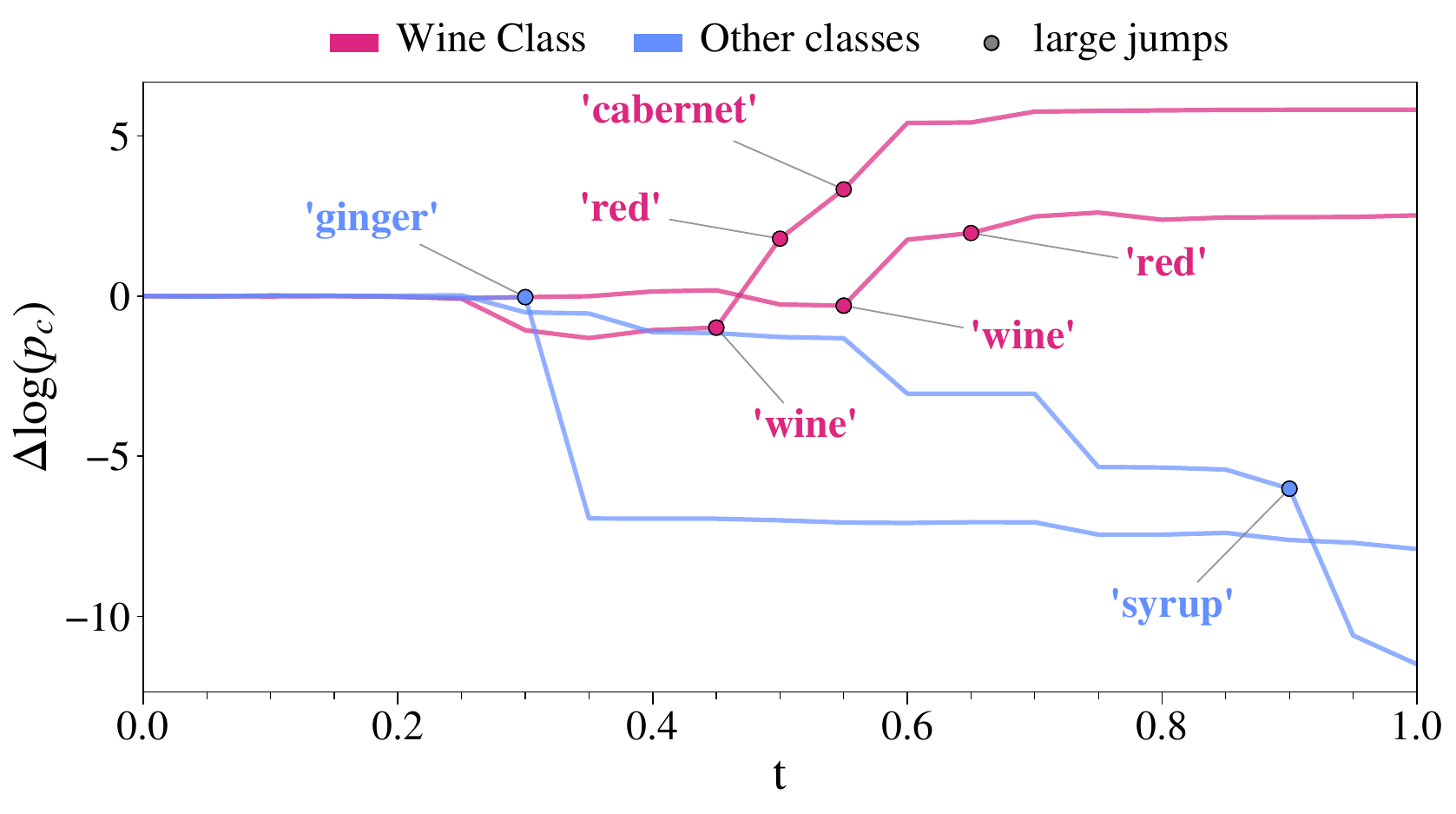}
        \caption{Illustrative posterior trajectories}
        \label{fig:posterior_trajectories}
    \end{subfigure}
    \hfill
    \begin{subfigure}[b]{0.41\textwidth}
        \centering
        \includegraphics[width=\linewidth]{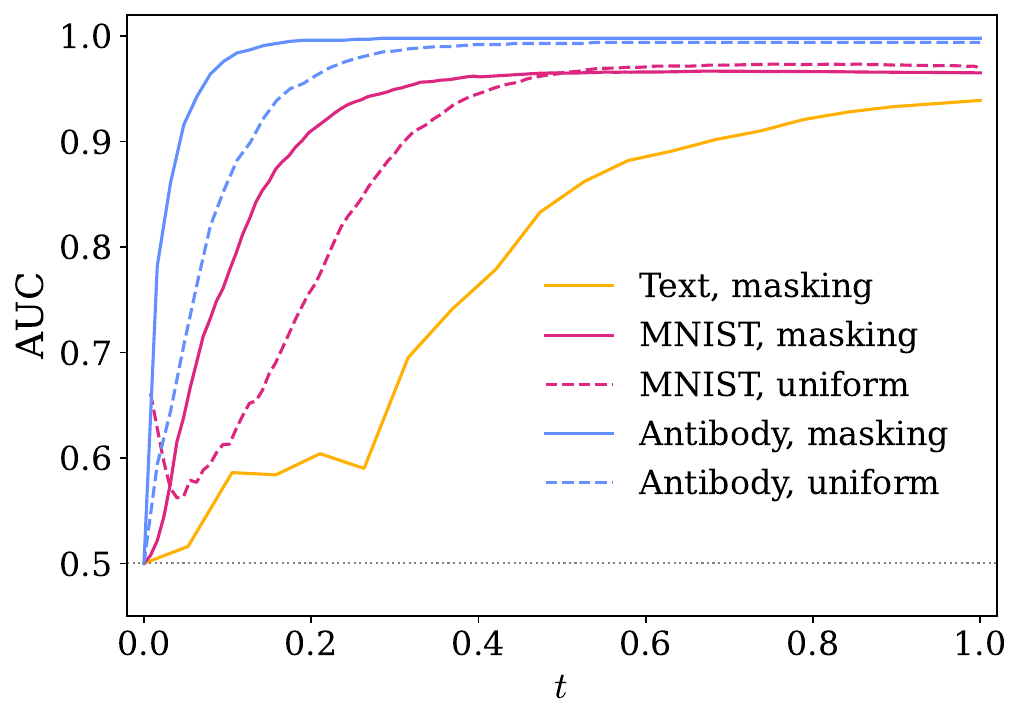}
        \caption{AUC during denoising}
        \label{subfig: AUC}
    \end{subfigure}
    \caption{
    In (a), the progress of the posterior approximation is shown throughout generation of four sentences on ``\textit{beverage}'' with as class ``\textit{wine}''. Highlighted words show large jumps in the posterior, corresponding to those given in \Cref{tab:intuitive_token_posterior}. In (b), one-vs-rest AUC is evaluated throughout generation, using a modality-specific external classifier to determine the final class.}
    \label{fig: posterior approx}
\end{figure}

\paragraph{Image generation.}
For the {image} modality, we train 
conditional models on MNIST using both masking and uniform noising schemes (training details and ablations in \cref{app: image - training details}).
During sampling, the posterior over the 10 MNIST classes is estimated online by applying a softmax to the class-specific log-posterior quantities computed with \cref{eq:iter_post_with_hyperparam}.
To evaluate the posterior throughout denoising, trajectories are classified using a pretrained ViT applied to the final state.
At each noise level \(t\), we evaluate the posterior scores in a one-vs-rest classification setting and report the AUC averaged over the ten classes, measuring how well the online posterior discerns the class eventually expressed by the generated sample.
As shown in \cref{subfig: AUC}, the AUC increases throughout denoising, indicating that the posterior becomes progressively more informative as the sample develops class-specific structure.
Representative trajectories and their corresponding posterior estimates are shown in \cref{subfig: Illustrative posterior estimations MNIST}, with additional examples in \cref{app: image - posterior estimation}.
In both noise processes, the posterior increases as class-specific structure emerges; compared to masking, the increase tends to occur later under uniform noising, caused by the ability of uniform diffusion to revise previously generated tokens and therefore defer commitment to a particular class.

\begin{table}[b!]
    \centering
    \setlength{\tabcolsep}{4pt}
    \begin{tabular}{llc cc cc cc}
    \toprule
    Topic & Sub-topic & & \multicolumn{2}{c}{1st}
        & \multicolumn{2}{c}{2nd}
        & \multicolumn{2}{c}{3rd} \\
    \cmidrule(lr){4-5}\cmidrule(lr){6-7}\cmidrule(lr){8-9}
    & (class c) & & Token & $\Delta\log p_c$ & Token & $\Delta\log p_c$
      & Token & $\Delta\log p_c$ \\
    \midrule
    \multirow{2}{*}{beverage} & \multirow{2}{*}{wine} & $+$ 
        & wine & +2.3 & \textcolor{gray}{cab}ernet & +1.3 & red & +1.1 \\
     & & $-$ 
        & syrup & -5.5 & ginger & -3.9 & espresso & -3.2 \\
    \midrule
    \multirow{2}{*}{sweet} & \multirow{2}{*}{candy} & $+$ 
        & candy & +2.8 & \textcolor{gray}{lollip}op & +0.8 & are & +0.5 \\
     & & $-$ 
        & \textcolor{gray}{mar}zip\textcolor{gray}{an} & -2.1
        & cookies & -0.7 & sweet & -0.6 \\
    \cmidrule(lr){1-9}
    \multirow{2}{*}{dairy product} & \multirow{2}{*}{cheese} & $+$ 
        & cheddar & +2.4 & Swiss & +1.4 & cheese & +0.5 \\
     & & $-$ 
        & yogurt & -6.8 & cream & -0.5 & product & -0.5 \\
    \bottomrule
    \end{tabular}
    \caption{Tokens pushing most towards ($+$) and away from ($-$) the class, estimated by our posterior estimation.
    (Sub)words in \textcolor{gray}{gray} provide context for the token shown in black. Each gray context corresponds to the actual context for that token in all samples containing that token.}
    \label{tab:intuitive_token_posterior}
\end{table}
\paragraph{Text generation.}
Here, evaluation is based on the Cognac \citep{chenControllableTextGeneration2022} benchmark, with as task generating a Wikipedia-style sentence around a certain topic.
The two models are obtained by prompting LLaDA-8B-Instruct \citep{LLaDa} with two different prompts.
The unconditional model is asked to generate about a category (e.g., ``beverage''), while the conditional model is prompted to generate about a sub-category (e.g., ``wine'') which we refer to as the classes. Full details of the experimental setup are in \cref{sec:experimental_setup_details_for_text_evaluation}.
Importantly, the online posterior estimation allows analyzing which unmasking events result in the generation of a sentence of a particular sub-category.
\Cref{tab:intuitive_token_posterior} quantifies this, by showing the tokens that have the strongest impact on the posterior across 1000 samples for different tasks.
Words with strong positive impact on the posterior are either a word that belongs to the class (e.g., `cheddar' for class cheese), or words that strongly implicate a class word (e.g., `Swiss' for class cheese).
Conversely, words with strong negative impact on the posterior often exemplify subcategories of the topic that are unrelated to the constraint (e.g., `yogurt'). 
In \Cref{fig:posterior_trajectories}, four trajectories are shown, illustrating how the posterior moves strongly in the right direction after words are generated exemplifying the class or the complement of a class.
\paragraph{Antibody sequence generation.}
For antibody sequence generation, we train a 50M-parameter Transformer based on the architecture of \citet{AMPLIFY} on the OAS dataset \citep{OAS,IgLM}, restricted to human antibodies. Antibody sequences are partitioned by V-gene type as classified by IgBLAST \citep{IgBLAST}. The V-gene largely determines the antibody framework, providing a strong conditioning signal. 
Our posterior estimation accurately identifies the V-gene from the partially generated sequence, achieving near-perfect AUC 
as shown in \Cref{subfig: AUC}. 
Furthermore, the posterior becomes informative after only a small number of token flips, allowing the conditioning variable to be identified early in generation. Further details on this modality are provided in \cref{app: antibody}.
\begin{figure}[t]
    \centering
    \begin{subfigure}[t]{0.45\textwidth}
        \centering
        \includegraphics[width=\linewidth]{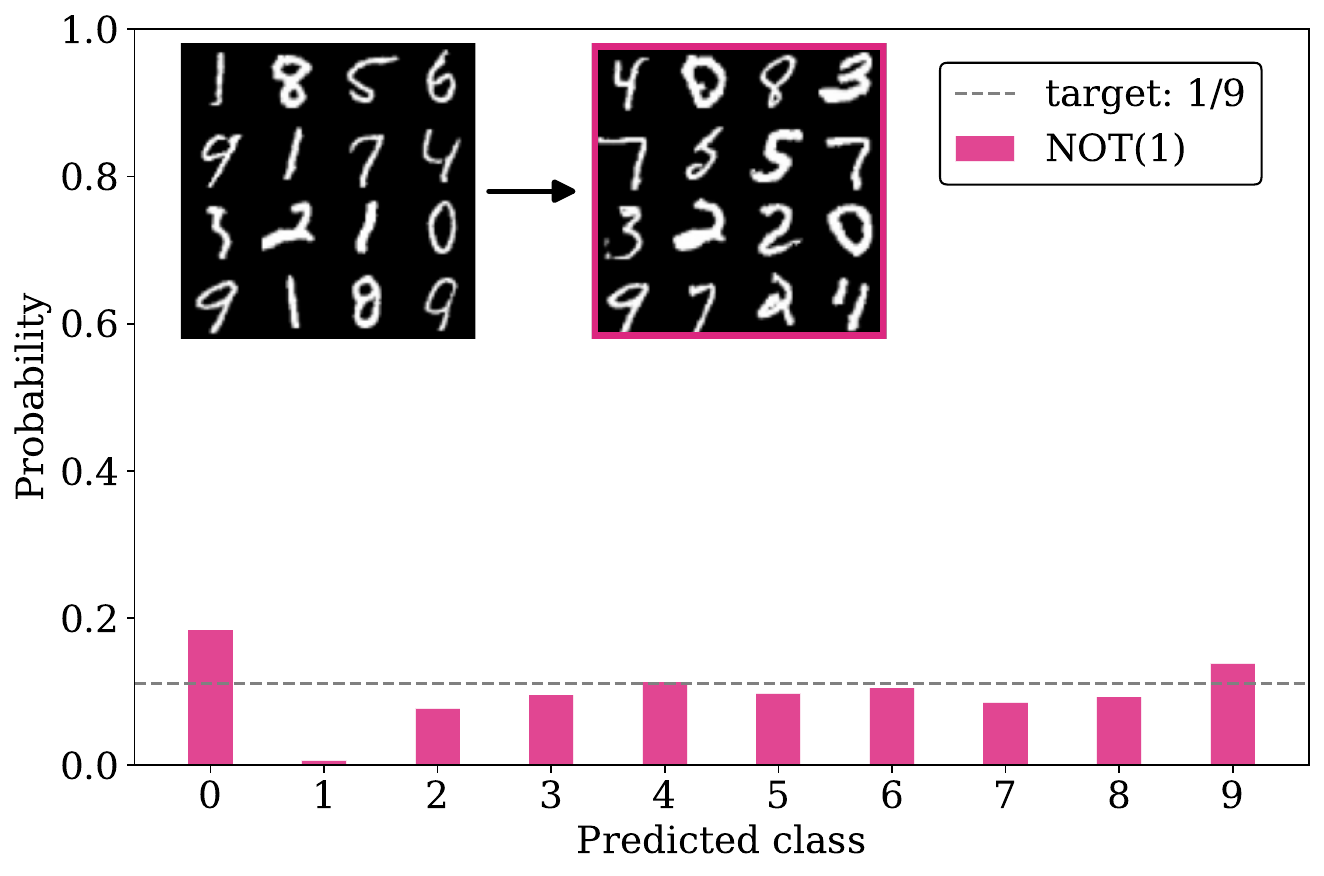}
        \caption{Samples under \textsc{not} gate}
        \label{subfig: Illustrative NOT gate MNIST}
    \end{subfigure}
    \hfill
    \begin{subfigure}[t]{0.45\textwidth}
        \centering
        \includegraphics[width=\linewidth]{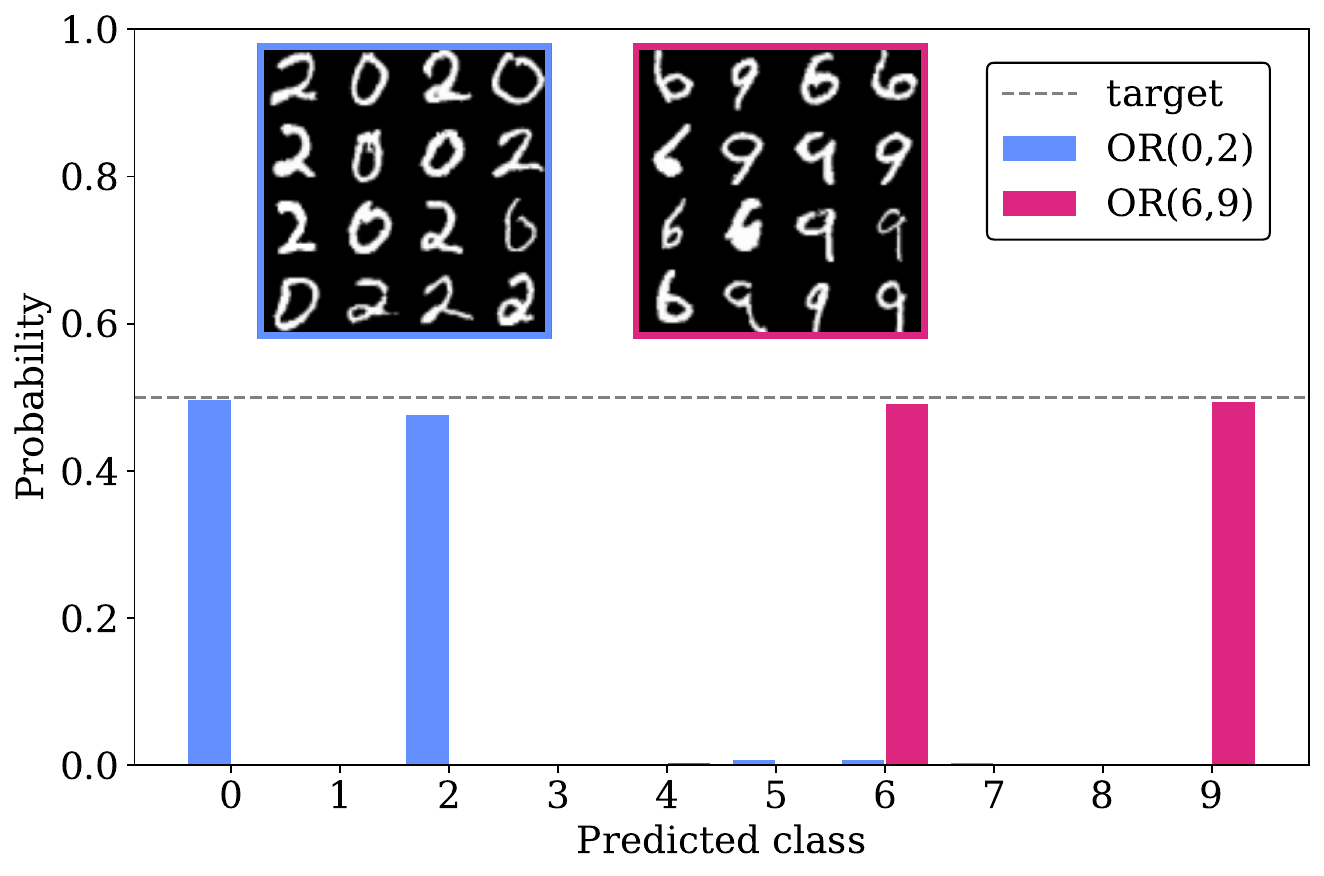}
        \caption{Samples under \textsc{or} gate}
        \label{subfig: Illustrative OR gate MNIST}
    \end{subfigure}
    \caption{Illustration of the \textsc{not} and \textsc{or} gates on MNIST under a uniform noise model. (a) The \textsc{not} gate, showing how unconditional samples containing the undesired number \textit{``1''} are transformed into other, predominantly round numbers such as \textit{``0''} and \textit{``9''}. (b) The \textsc{or} gate evaluated on two distinct pairs, with corresponding samples shown above.}
    \label{fig: samples under gates}
\end{figure}
\begin{figure}[b!]
    \centering
    \includegraphics[width=\linewidth]{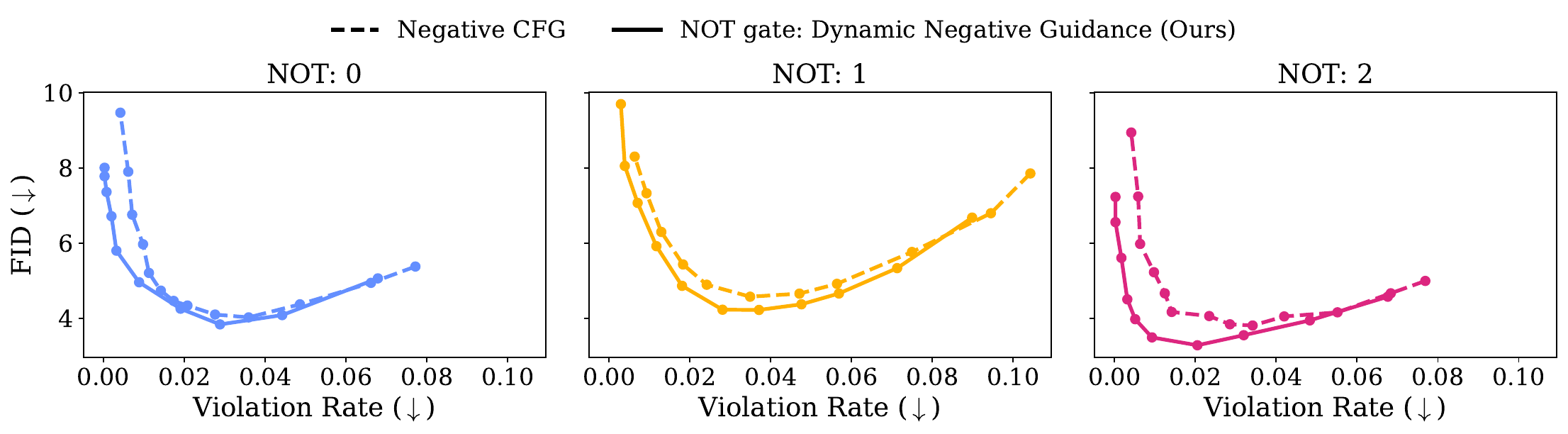}
    \caption{Pareto curves comparing the quality of the generated samples to the safety of the approach, respectively through the FID and the violation rate. Both negative CFG, as well as the theoretically justified dynamic \textsc{not} gate display a trade-off between quality and safety, with the dynamic approach outperforming the static one, albeit at the cost of an additional hyperparameter.}
    \label{fig: mnist NOT paretto}
\end{figure}
\subsection{Logical Operators}
\label{sec:not_gate}
Of the three gates, the \textsc{not} gate is most interesting, combining two rate functions into an otherwise unavailable novel rate function.
Note that in \textbf{text}, functionality of the \textsc{not} gate can be achieved adapting the prompt by stating which class to avoid, a baseline our method matches as shown in \cref{app:text_not}.
Next, we consider the more interesting cases of applying \textsc{not} for image and antibody sequence generation. For validation of the theory, rather than intended for practical application, we also evaluate the \textsc{or} gate in the image modality.
\paragraph{Image generation.} The \textsc{not} gate is evaluated using the Pareto front of safety and quality, measured respectively by the violation rate against a ViT classifier, and FID on 4096 samples~\citep{FID}.
In \Cref{fig: mnist NOT paretto}, the dynamic approach is compared to static CFG-style negative logit guidance, which samples from $\tilde{p}(x)\propto p_1(x_1)p_1(c\mid x_1)^{-\lambda}$. 
A grid search for the optimal hyperparameter is performed for the two approaches, with more details in \cref{app: image - logical gates}. Both methods exhibit the expected trade-off between violation rate and image quality. The dynamic \textsc{not} gate consistently achieves a better trade-off than negative CFG, at the cost of the additional hyperparameter $\tau$. \Cref{subfig: Illustrative NOT gate MNIST} shows the distribution resulting from excluding the class ``1'' and how it impacts samples generated from the same seed using the unconditional model. The \textsc{or} gate is evaluated similarly by restricting the posterior to the two desired conditions, visualized in \Cref{subfig: Illustrative OR gate MNIST}. 
\paragraph{Antibody sequence generation.}
In this setting, we evaluate 
how the \textsc{not} gate removes a V-gene family while preserving quality, measured respectively using violation rate and Fréchet Protein Distance (FPD) to the desired population in ESM2 embedding space \citep{ESM2}. The dynamic \textsc{not} gate is compared against unconditional generation and static negative CFG.

A key challenge in this setting is the strong imbalance between V-gene families: nearly 95\% of sequences belong to just three of the seven families. Removing a majority class requires substantially stronger guidance than removing a minority class. The dynamic formulation accounts for this imbalance as the initial guidance strength scales with the class prior. 
In \Cref{tab:not_gate_baselines}, we choose a setting such that both majority and minority classes are removed with only small FPD increases.\footnote{See \Cref{app: antibody - logical gates} for the full Pareto curves and guidance sweeps.} For comparison, 
negative CFG is calibrated separately to achieve a comparable violation rate for majority and minority classes. Majority-calibrated guidance is overly strong for minority classes, causing a substantial increase in FPD, while minority-calibrated guidance is too weak for majority classes. 
The dynamic \textsc{not} gate does not suffer the same flaw thanks to its prior-weighted initialization.
\begin{table}[t]
\centering
\begin{tabular}{ll cc cc}
\toprule
Method & Setting
& \multicolumn{2}{c}{Majority}
& \multicolumn{2}{c}{Minority} \\
\cmidrule(lr){3-4} \cmidrule(lr){5-6}
&& Viol. $\downarrow$ & FPD $\downarrow$
& Viol. $\downarrow$ & FPD $\downarrow$ \\
\midrule
Unconditional & 
& 0.278 & 0.174
& 0.032 & 0.007 \\

\multirow{2}{*}{Negative CFG}
& majority-calibrated
& 0.013 & 0.104
& \textcolor{gray}{0.002} & \textcolor{gray}{0.326} \\
& minority-calibrated
& \textcolor{gray}{0.186} & \textcolor{gray}{0.084}
& 0.017 & 0.020 \\

\textsc{not} gate & both
& 0.010 & 0.074
& 0.016 & 0.010 \\
\bottomrule
\end{tabular}
\caption{
\textsc{not} gate performance when removing majority and minority V-gene families. Static guidance is calibrated to match the dynamic method's violation rate in the corresponding regime, then transferred across regimes. \textcolor{gray}{Gray} entries show performance outside the calibration regime. The full Pareto front is given in \cref{app: antibody - logical gates}.}
\label{tab:not_gate_baselines}

\end{table}
\section*{Conclusion}
We have shown that the class-conditional posterior in discrete diffusion can be tracked online, at no cost beyond the conditional and unconditional predictions already required for guidance. In diffusion language modeling this posterior is highly interpretable: individual tokens can be seen to commit a sample toward a class or away from it during generation. Building on the posterior, we derived the \textsc{and}, \textsc{or}, and \textsc{not} operators over discrete rate functions, providing a principled means of composing conditional processes. The \textsc{not} gate is of particular practical value, enabling generation that steers away from a class in modalities such as images and antibody sequences, where prompting a model to avoid a concept is not possible. As the construction is agnostic to modality, noising process, data imbalances and schedules, it applies unchanged across the settings we consider. The derivation additionally revealed a previously unremarked discrepancy between classifier-free guidance in the continuous domain, an additive mixture of scores, and the discrete domain, a multiplicative mixture of predictions. Limitations are provided in \Cref{sec:limitations}.
\subsubsection*{Acknowledgments}
This research was partly funded by the Research Foundation - Flanders (FWO-Vlaanderen) under grant G0C2723N and by the Flemish Government (AI Research Program). We would like to thank Leander Gabriel, Florian Handke and Johannes Deleu for the interesting and insightful discussions that have led up to some of this work, particularly in understanding the posterior estimation algorithm. 

\subsubsection*{AI disclosure}
In this work, we used generative AI tools to 
implement methods used in this paper.
Additionally, we used generative AI tools to
plot results, 
create or edit software code, 
and edit the paper to improve readability.
We have not used generative AI tools for other purposes.
We have reviewed all AI-assisted work.  
We take responsibility for the final content of this work, including text, claims or artifacts produced with the aid of generative AI.



\bibliography{iclr2027_conference}
\bibliographystyle{iclr2027_conference}

\newpage

\appendix

\crefalias{section}{appsec}
\crefalias{subsection}{appsec}

\section{Limitations and future work}
\label{sec:limitations}
Due to the broad scope of the current paper, spanning several modalities and noising schemes, a detailed analysis of how diffusion-language-modeling choices such as confidence unmasking, block decoding, and remasking interact with the posterior is absent. We leave this to future work. A more fundamental limitation is that the posterior is updated only after the state has changed, so a token that strongly determines a class affects the estimate only once it has been unmasked. For the \textsc{not} gate this is a genuine shortcoming: ideally the posterior would reshape the transition probabilities before such tokens are generated, preventing the unwanted class rather than reacting to it. We view this anticipatory formulation as a promising direction.

\section{Additional derivations}
\label{app: theory}
In this appendix additional useful derivations are given. In \cref{app: theory - C vs D-CFG}, the difference between how the common classifier-freee guidance method works in the discrete and continuous domains is discussed. Thereafter, in \cref{app: theory - logical operators}, a detailed derivation of all logical operators introduced in the main body is given. Then, in \cref{app: theory - posterior history dependence} an important note regarding under which samplers our online posterior estimate is equivalent to the ground-truth marginal computed posterior.

\subsection{Continuous vs discrete CFG}
\label{app: theory - C vs D-CFG}
In the continuous diffusion literature guidance is seen as a multiplicative corruption/sharpening at the marginal level which results in an additive mixture of model predictions or score functions. In the discrete diffusion literature however it is common to see this corruption as a multiplicative tweaking of the predicitons themselves which results in an additive mixture of the predicted logits. This difference is not well known yet quite significant for our work as here we start with the same assumptions as in the continuous case, namely corruptions on the marginal level, and derive the rate function that would yield the discrete analogue of continuous CFG, which we show to be an additive mixture of the rate functions.\\
Let us frame this in more detail mathematically. Classifier-free guidance starts of from a sharpening of the marginals:
\begin{equation}
    p^{\text{CFG}}(x_t) = p(x_t) \frac{p(x_t|c)^{\lambda}}{p(x_t)^{\lambda}}
\end{equation}
When looking at the infinitesimal generator of that distribution, which in the continuous case is its score function, one finds:
\begin{equation}
    \nabla \log p^{\text{CFG}}(x_t) = \nabla \log p(x_t) + \lambda \Big(\nabla \log p(x_t|c) - \nabla \log p(x_t)\Big)
\end{equation}
Using tweedies formula the score function depends linearly on the denoiser prediction $\hat{x}_0 = \mathbb{E}[x_0|x_t]$ such that in practice CFG can be implemented as a linear combination of the denoisers prediction:
\begin{equation}
    \hat{x}_0^{\text{CFG}}(x_t) = \hat{x}_0(x_t) + \lambda \Big(\hat{x}_0(x_t|c) - \hat{x}_0(x_t)\Big)
\end{equation}
In our derivation of the \textit{and} operation we start from the exact same assumption and find that for a multiplicative mixture of the marginals one again needs to add their infinitesimal generators, which in the discrete case are given by the rate matrix $R_t$ instead of the score functions $\nabla \log p(x_t)$. This time, the rate functions linearly depend on the models token predictions $p(x_0|x_t)$, hence these should be added together to obtain CFG.

In the discrete diffusion literature, guidance is derived from a different assumption, namely a multiplicative mixture of the model predictions. This is erroneously described as the assumption used in the continuous case,\footnote{The error is most visible in the notation used for what they define as a score function $\nabla \log p(x_s|x_t,c)$, which is incorrect. The score function is the gradient of the log likelihood of the marginal, namely $\nabla \log p(x_t|c)$, and not of the transition kernel.}
but beyond that this is a perfectly fine modification of the generative process.
More precisely, the assumption commonly used in the discrete case is the following:
\begin{equation}
    p^{\text{D-CFG}}(x_0|x_t,c) = p(x_0|x_t) \frac{p(x_0|x_t,c)^{\lambda}}{p(x_0|x_t)^{\lambda}}
\end{equation}
Typically models are trained to predict logits, that is log probabilities, in which the decomposition becomes:
\begin{equation}
    \log p^{\text{D-CFG}}(x_0|x_t,c) = \log p(x_0|x_t) + \lambda \Big(\log p(x_0|x_t,c)-\log p(x_0|x_t)\Big)
\end{equation}
The main advantage of this modification is that by changing the distribution at the logits level, there is no risk to obtain an invalid distribution, because a softmax normalization will always be applied afterwards. 

In the following section we will try to unravel what the differences are between these two assumptions and what possible pitfalls might be. Let us first try to see how one assumption transfers into the second one:
\begin{align}
    p^{\text{D-CFG}}(x_0|x_t,c) & = p(x_0|x_t) \frac{p(x_0|x_t,c)^{\lambda}}
    {p(x_0|x_t)^{\lambda}}\\
    & = p(x_0)\frac{p(x_t|x_0)}{p(x_t)} \frac{\Big(p(x_0|c)\frac{p(x_t|x_0,c)}{p(x_t|c)}\Big)^{\lambda}}
    {\Big(p(x_0)\frac{p(x_t|x_0)}{p(x_t)}\Big)^{\lambda}}\\
    & = p(x_0) \frac{p(x_0|c)^{\lambda}}{p(x_0)^{\lambda}} \frac{p(x_t|x_0) \frac{p(x_t|x_0,c)^{\lambda}}{p(x_t|x_0)^{\lambda}}}{p(x_t) \frac{p(x_t|c)^{\lambda}}{p(x_t)^{\lambda}}}
\end{align}
Realizing that due to the conditional independence of the noising process we have $p(x_t|x_0) = p(x_t|x_0,c)$, and if we further note that $p^{\text{CFG}}(x_t) = p(x_t) \frac{p(x_t|c)^{\lambda}}{p(x_t)^{\lambda}}$ we find:
\begin{align}
    p^{\text{D-CFG}}(x_0|x_t,c) & = p(x_0) \frac{p(x_0|c)^{\lambda}}{p(x_0)^{\lambda}} \frac{p(x_t|x_0)}{p(x_t) \frac{p(x_t|c)^{\lambda}}{p(x_t)^{\lambda}}}\\
    & = p^{\text{CFG}}(x_0|c) \frac{p(x_t|x_0)}{p^{\text{CFG}}(x_t|c)}\\
    \label{eq: appendix DCFG vs CFG}
\end{align}
Now, one might think that we can invoke the conditional independence of the noising process to the sharpened distribution (i.e., setting $p(x_t|x_0)=p^{\text{CFG}}(x_t|x_0,c)$ to obtain through Bayes that $p^{\text{D-CFG}}(x_0|x_t,c)=p^{\text{CFG}}(x_0|x_t,c)$. This is however erroneous. Silently doing this would require $p^{\text{CFG}}(x_t|c) = \sum_{x_0'}p(x_t|x_0')p^{\text{CFG}}(x_0|c)$,\footnote{This needs to hold to apply Bayes' rule as a valid joint distribution should be obtained. Alternatively the sum over $x_0$ in \cref{eq: appendix DCFG vs CFG} should yield 1.}
i.e., that the tilted marginals correspond to the noised version of the tilted data distribution. When applying a multiplicative mixture at the distributional level this is not true, as the convolution behind the noising operation does not commute with the multiplicative mixing operation. In fact a similar result is the reason that CFG is heavily criticized in the continuous case, the sharpening of the marginals does not result in samples that correspond to the expected sharpened data distribution. 

\begin{figure}
    \centering
    \includegraphics[width=0.8\linewidth]{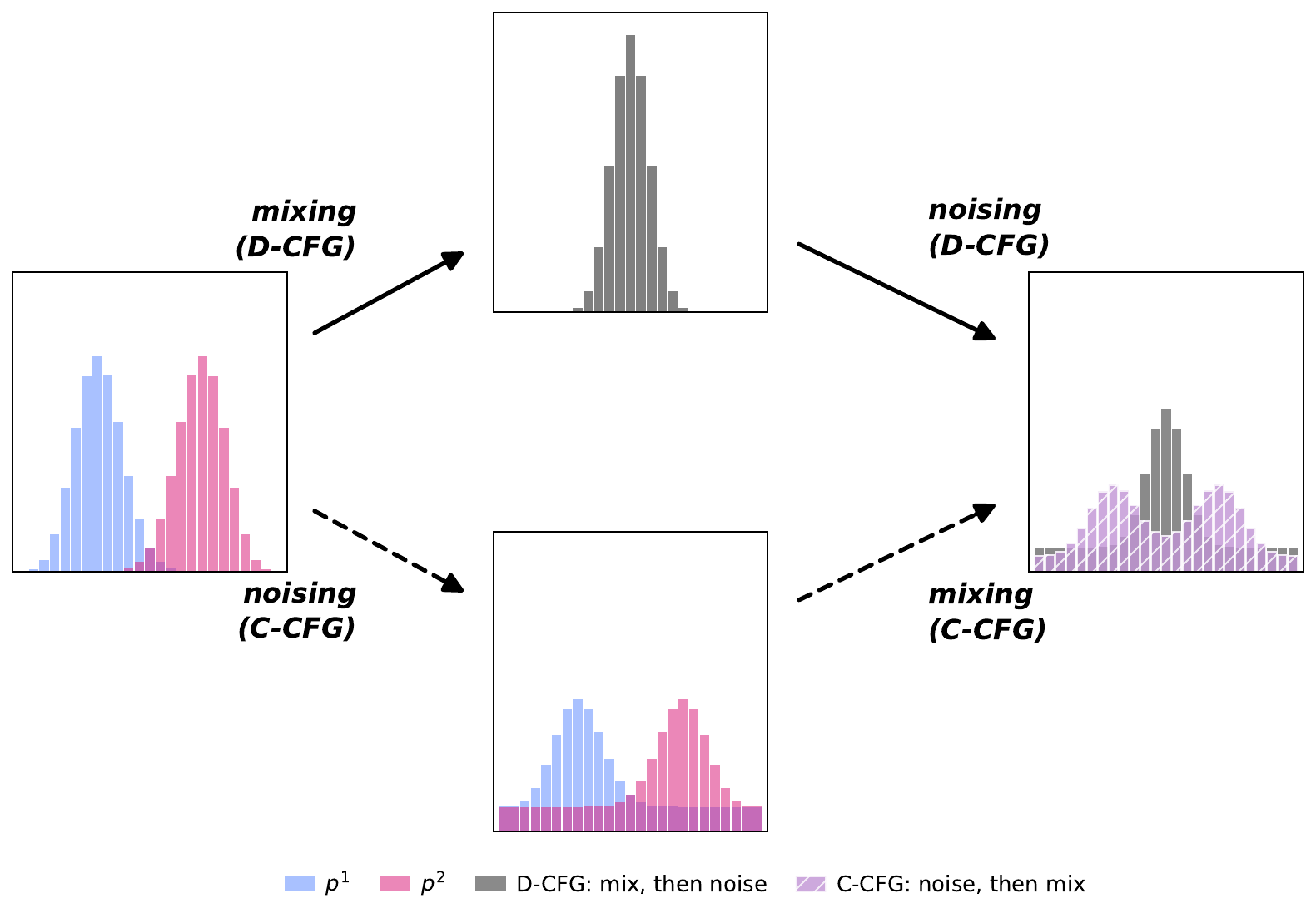}
    \caption{Illustration of the non-commutativity of noising and mixing in the case of a multiplicative mixture. Continuous CFG first noises the data distribution and mixes the marginals, whereas discrete CFG first mixes the distributions and then only noises.}
    \label{fig (app): D- vs C-CFG illustration}
\end{figure}

So we find that the difference between the two process lies in the non-commutativity of the noising and mixing operations, or mathematically by the non-commutativity of the convolution and the multiplication. Typical continuous CFG corresponds to sharpening all the marginals individually, while discrete CFG yields the denoiser that retrieves the sharpened data distribution. The difference between the two orders of mixing and noising are further illustrated in \Cref{fig (app): D- vs C-CFG illustration}. 

To prove this second fact let's derive which prediction $p^{\lambda}(x_0|x_t)$, or rate function, would yield a $\lambda$-sharpened data distribution $p^{\lambda}(x_0|c) = p^{\text{CFG}}(x_0|c) = p(x_0) \frac{p(x_0|c)^{\lambda}}{p(x_0)^{\lambda}}$. For this let us write out the true predictor of that distribution namely:
\begin{equation}
    p^{\lambda}(x_0|x_t,c) = \frac{p(x_t|x_0) p^{\lambda}(x_0|c)}{\sum_{x_0'}p(x_t|x_0')p^{\lambda}(x_0'|c)}
\end{equation}
Or by flipping the conditioning using Bayes and only looking at the $x_0$ dependence as the rest is simply a normalization constant that will be taken care off by the softmax:
\begin{align}
    p^{\lambda}(x_0|x_t,c) &\propto_{x_0} \big(p(x_t|x_0) p(x_0)\big)^{1-\lambda}\big(p(x_t|x_0,c) p(x_0|c)\big)^{\lambda}\\
    &=  \big(p(x_0|x_t) p(x_t)\big)^{1-\lambda}\big(p(x_0|x_t,c) p(x_t|c)\big)^{\lambda}\\
    &=  p(x_0|x_t)^{1-\lambda}p(x_0|x_t,c)^{\lambda} p(x_t)^{1-\lambda} p(x_t|c)^{\lambda}\\
    &\propto_{x_0}  p(x_0|x_t)^{1-\lambda}p(x_0|x_t,c)^{\lambda} \\
\end{align}
Hence we find that the exact denoiser, or generator, that will yield samples from the tilted data distribution is precisely the one implemented in the discrete version of CFG. Hence this proves that by adding the logits together, the generator sampling from the sharpened data distribution is obtained.

\subsection{Detailed derivation of Logical Operators: \textsc{and}, \textsc{or}, \textsc{not}}
\label{app: theory - logical operators}
In this section, the derivations leading up to the different logical operators outlined in the main paper is given in more detailed. The same ordering as in the main body is kept, starting with the derivation for the \textsc{and} operator, then the \textsc{or} operator and finally the \textsc{not} operator. For all cases the resulting process is analyzed when mixing the data distributions as well as when mixing the marginals themselves. Naturally as outlined in the main body for the \textsc{or} and \textsc{not} operators both result in the same process due to the commutativity of their specific additive mixing operations with noising. 
\paragraph{\textsc{and}-gate. \textit{(data-level mixture)}}
In this case we consider a multiplicative mixture on the data distribution level, i.e., $p^{\Lambda}_0(x_0) \propto \prod_{i=1}^N p_0(x_0|c_i)$. The key challenge is to express the predictor that leads to the mixture distribution, denoted by $p^{\Lambda}(x_0|x_t)$, as a function of the assumed known conditional predictors $p_0(x_0|c_i)$. Once this mixture predictor is known, we construct the according rate function using Eq.(\ref{eq: velocity fields}). The mixture denoiser can be computed by rewriting it as a data weighted function of the noising kernel through Bayes:
\begin{align}
    p^{\Lambda}(x_0|x_t) &= p(x_t|x_0) \frac{p^{\Lambda}_0(x_0)}{p^{\Lambda}_t(x_t)}\\
    &\propto p(x_t|x_0) \prod_{i=1}^N p^{\Lambda}_0(x_0|c_i) \\
    &\propto p(x_t|x_0)^{1-N}\prod_{i=1}^N p(x_t|x_0)  p^{\Lambda}_0(x_0|c_i) \\
    &\propto p(x_t|x_0)^{1-N}\prod_{i=1}^N p^{\Lambda}(x_0|x_t,c) 
\end{align}
Or thus the predictions distributions should be multiplied with each other, this is more commonly implemented as adding logits together:
\begin{align}
    \log p(x_0|x_t) &\propto \sum_{i=1}^N \log p(x_0|x_t,c) \label{eq:and data level (app)}
\end{align}
From this predictor a rate function can be constructed through Eq. (\ref{eq: velocity fields}). This is precisely how CFG is commonly implemented in the discrete diffusion literature, i.e., such as in eq (\ref{eq: discrete CFG}). 
\paragraph{\textsc{and} gate. \textit{(marginal-level mixture)}}
On the other hand, for the multiplicative mixing an approach reminiscent of the continuous diffusion literature can be applied, namely considering a multiplicative mixture of the marginals defined as $p^{\Lambda}_t(x_t) \propto \prod_{i=1}^N p_t(x_t|c_i)$. Here it's simplest to directly derive the joint rate function through:
\begin{align}
    \frac{\mathrm{d}p^{\Lambda}_t(x_t)}{\mathrm{d}t} & = \sum_{i}\frac{\mathrm{d}p_t(x_t|c_i)}{\mathrm{d}t}\prod_{j\neq i}^N p(x_t|c_j)\\
    & = \sum_{i}R_i p_t(x_t|c_i)\prod_{j\neq i}^N p(x_t|c_j)\\
    & = \sum_{i}R_i\prod_{j}^N p(x_t|c_j)\\
    &= \Big(\sum_{i}R_i\Big)p^{\Lambda}_t(x_t)
\label{eq: AND gate derivation (app)}
\end{align}
Hence the rate matrices of the joint process satisfies:
\begin{equation}
    R^{\Lambda}_t = \sum_{i=1}^N R^i_t. \label{eq:and marginal level (app)}
\end{equation}
By construction $R_t^{\Lambda}$ inherits the required properties of a valid rate matrix, namely non-negative off-diagonal entries and rows summing to zero, so the resulting process remains valid. The shape of this combination is also intuitive: as long as one of the two conditions is not satisfied, the corresponding rate will try to strongly modify the outcome. Only once $x_t$ satisfies both conditions will both rate functions remain reasonably small. This is entirely analogous to how CFG is implemented in the continuous diffusion literature, namely as an addition of two infinitesimal generators; in the latter a score function, and in this case a rate function.
\paragraph{\textsc{or}-gate. \textit{(data-level mixture)}}
In this context, mixing multiple distributions together in data space happens through a sum over the parts, i.e.
$p^{\Lambda}_1(x_1) = \sum_{i=1}^N p(c_i)p_1(x_1|c_i)$.
Looking at how the denoiser for this distribution can be expressed as a function of the conditional denoisers $p_1(x_1|x_t,c_i)$, one has:
\begin{align}
    p^{\Lambda}_1(x_1|x_t)
    &= p(x_t|x_1) \frac{p^{\Lambda}_1(x_1)}{p^{\Lambda}_t(x_t)}\\
    &= p(x_t|x_1) \sum_{i=1}^N p(c_i)
    \frac{p_1(x_1|c_i)}{p^{\Lambda}_t(x_t)}\\
    &= \sum_{i=1}^N p(c_i)
    \frac{p(x_t|x_1)p_1(x_1|c_i)}
    {p^{\Lambda}_t(x_t)}\\
    &= \sum_{i=1}^N p(c_i)
    \frac{p_t(x_t|c_i)}
    {p^{\Lambda}_t(x_t)}
    p_1(x_1|x_t,c_i)\\
    &= \sum_{i=1}^N p(c_i|x_t)
    p_1(x_1|x_t,c_i).
\end{align}
Hence, the denoiser of the mixture distribution is given by a posterior-weighted average of the conditional denoisers. The corresponding rate function can then be constructed from this mixture denoiser using Eq.~(\ref{eq: velocity fields}).

\paragraph{\textsc{or}-gate. \textit{(marginal level mixture)}}
For the additive operation behind the \textsc{or} gate $p^{\Lambda}_t(x) = \sum_{i=1}^N p(c_i)\,p_t(x_t|c_i)$ which can be interpreted as an unconditional model $p_t(x)$ in this setting, with as mixture weights the priors $p(c_i)$, one can again derive the joint rate function directly:
\begin{align}
    \frac{\mathrm{d}p^{\Lambda}_t(x_t)}{\mathrm{d}t} & = \sum_{i}p(c_i)\frac{\mathrm{d}p_t(x_t|c_i)}{\mathrm{d}t}\\
    &= \sum_{i} R^i_t p(c_i) p_t(x_t|c_i)\\
    &= \Big(\sum_{i}p(c_i) \frac{p_t(x_t|c_i)}{p(x_t)} R^i_t \Big)p_t(x_t)\\
    &= \Big(\sum_{i}p(c_i|x_t) R^i_t\Big)p^{\Lambda}_t(x_t)
\label{eq: OR gate derivation (app)}
\end{align}
Where in the second line we multiply and divide by the mixture marginal $p^{\Lambda}_t(x_t)$.
In other words,
\begin{equation}
    R_t^{\Lambda} = \sum_{i=1}^N p(c_i\mid x_t)\,R_t^{i}. \label{eq:or (app)}
\end{equation}
Importantly, the posterior quantity is interpreted here as a classifier \textit{restricted} to the specified conditions. This is precisely the mechanism that allows a dataset-specific predictor to be learned while training only on instances from that dataset. A predictor approximating the datapoint-specific distributions $p(x_0\mid x_t)$, with each datapoint viewed as a class, can be averaged over the corresponding deterministic datapoint-specific rates to recover a global rate, an idea closely related to the construction used in Flow Matching \citep{FM,DFM}. The results show that this construction extends to arbitrary partitions of the datapoints. Beyond this theoretical connection, the \textsc{or} gate is not strictly necessary for class-conditional generation, since sampling a condition according to its prior, and subsequently using the corresponding conditional model, yields the same desired distribution.

\paragraph{\textsc{not}-gate. \textit{(data-level mixture)}}
For the \textsc{not} gate, i.e., removing a particular part from a larger distribution, the resulting distribution corresponds to the complement of that part.
In particular, removing condition $c_i$ from the unconditional distribution gives
\begin{equation}
    p^{\Lambda}_1(x_1)=p_1(x_1| \neg c_i)=\frac{1}{1-p(c_i)}\big(p_1(x_1)-p(c_i)p_1(x_1| c_i)\big).
\end{equation}
Importantly, because the addition commutes with the convolution based noising kernel, the same identity holds at the marginal level:
\begin{equation}
    p^{\Lambda}_t(x_t)=p_t(x_t| \neg c_i)=\frac{1}{1-p(c_i)}\big(p_t(x_t)-p(c_i)p_t(x_t| c_i)\big).
\end{equation}
Deriving the predictor for this distribution gives:
\begin{align}
    p^{\Lambda}_1(x_1| x_t)&=p(x_t| x_1)\frac{p^{\Lambda}_1(x_1)}{p^{\Lambda}_t(x_t)}\\
    &=\frac{p(x_t|x_1)\big(p_1(x_1)-p(c_i)p_1(x_1| c_i)\big)}{p_t(x_t)-p(c_i)p_t(x_t| c_i)}\\
    &=\frac{p_t(x_t)p_1(x_1| x_t)-p(c_i)p_t(x_t| c_i)p_1(x_1| x_t,c_i)}{p_t(x_t)-p(c_i)p_t(x_t| c_i)}\\
    &=\frac{p_1(x_1| x_t)-p(c_i| x_t)p_1(x_1| x_t,c_i)}{1-p(c_i| x_t)}
\end{align}
Hence, the denoiser for the \textsc{not} operation can be written as a posterior-dependent modification of the unconditional denoiser:
\begin{equation}
    p^{\Lambda}_1(x_1| x_t) = p_1(x_1| x_t) - \frac{p(c_i| x_t)}{1-p(c_i | x_t)}\Big(    p_1(x_1| x_t, c_i)-p_1(x_1 | x_t)\Big).
\end{equation}
The corresponding rate function can then be constructed from this predictor using Eq.~(\ref{eq: velocity fields}).

\paragraph{\textsc{not}-gate. \textit{(marginal-level mixture)}}
The removal, or avoidance, of a certain class only given it's class-specific rate can be mathematically expressed as the subtraction of a conditional distribution from its unconditional distribution $p^{\Lambda}_t(x_t|c_i) \propto p_t(x_t) - p(c_i)\,p_t(x_t|c_i)$. The joint rate can again be derived as:\footnote{The same remark as for the \textsc{or} gate holds here.}
\begin{align}
    \frac{\mathrm{d}p^{\Lambda}_t(x_t)}{\mathrm{d}t} & = \frac{1}{1-p(c_i)} \Big( R_t p_t(x_t) -p(c)R^i_t p_t(x_t|c)\Big)\\
    & = \frac{1}{1-p(c_i)} \Big( R_t \big(p_t(x_t)-p(c_i)p_t(x_t|c_i)\big) -p(c_i)p_t(x_t|c)\big(R^i_t-R_t\big)\Big)\\
    & = \Big(R_t - \frac{p_t(c_i|x_t)}{1-p_t(c_i|x_t)}\big(R_t^{c}-R_t\big)\Big)p^{\Lambda}_t(x_t)
\label{eq: NOT gate derivation (app)}
\end{align}
We therefore find:
\begin{equation}
    R_t^{\Lambda} = R_t - \frac{p(c\mid x_t)}{1-p(c\mid x_t)}\bigl(R_t^c - R_t\bigr). \label{eq:not (app)}
\end{equation}
This equation strongly resembles a classifier-free guidance equation, in which the posterior ratio can be seen as a \textit{dynamic} guidance scale that is small if the undesired class if unlikely to be generated and that becomes asymptotically large if that class becomes likely to be generated. Again, this is entirely analogous to what is found in the continuous case, with as single difference that the continuous generators, or score-functions, are replaced by their discrete analogues, the rate functions~\cite{DNG}. 
\subsection{History independence of the posterior}
\label{app: theory - posterior history dependence}
An important point is that the posterior tracked throughout inference is the likelihood conditioned on the whole denoising trajectory and not only its least noisy state $x_t$. This is a consequence of the way the posterior estimation is constructed through the Markov chain:
\begin{align}
    p(c|x_0,..x_t) &= p(c) \frac{p(x_0,...,x_t|c)}{p(x_0,...,x_t)}\\
    &= p(c) \frac{p(x_0,...,x_{t-1}|c)}{p(x_0,...,x_{t-1})}\frac{p(x_t|x_{t-1},c)}{p(x_t|x_{t-1},c)}\\
    &= p(c|x_0,...,x_{t-1})\frac{p(x_t|x_{t-1},c)}{p(x_t|x_{t-1},c)}\\
\end{align}
Which corresponds to the log posterior update scheme described:
\begin{align}
    \log p(c|x_0,..x_t) = \log p(c|x_0,...,x_{t-1}|c)+ \log p(x_t|x_{t-1},c) - \log p(x_t|x_{t-1},c)\\
\end{align}
Crucially however, thanks to the conditional independence of the noising process, the following holds:
\begin{align}
    p(c|x_0,..x_t) & = p(c) \frac{p(x_0,...,x_t|c)}{p(x_0,...,x_t)}\\
    &=p(c) \frac{p(x_t|c)}{p(x_t)}\frac{p(x_0,...,x_{t-1}|x_t,c)}{p(x_0,...,x_{t-1}|x_t)}\\
    &=p(c|x_t)\\
\end{align}
Here the important part is that the dependence on the whole trajectory drops out due to the fact that conditional and unconditional sequences get noised in the same way i.e., the noising is independent on the condition. However, for this to hold the sampler needs to be memoryless in the sense that when the denoising process is reversed it should give rise to the noising process itself. This subtlety is not satisfied by all samplers, such as the ODE sampler of DFM that we use in this paper. An example of a moment where such a subtlety is relevant is when using the uniform model. Without a purely stochastic ancestral sampler the quantity tracked through inference is not exactly equal to the marginal computed posterior, as shown in \Cref{fig (app): posterior approx} . We find however that the DFM sampler works best in practice on trained models, such as the ones trained on MNIST or OAS. An advantage of the masking noising scheme is that this equality is always satisfied, meaning that the posterior tracking will always give rise to the true marginal computed posterior. This is because in the case of a masked model the least noisy state truly holds all the information regarding the final state when compared to it's history. In essence the older states contain the exact same words but with a higher masking rate. 

\section{Additional details regarding synthetic data evaluation}

To verify our theoretical framework, in particular the online posterior estimation and the logical operators, we first work in a synthetic regime in which everything is analytically tractable. \Cref{app: synth data - set up} describes the constructed dataset and how the conditional and unconditional predictors are obtained in closed form, the latter both by exhaustive enumeration and as a mixture of the conditionals. \Cref{app: synth data - posterior validation} compares the online posterior estimate to the exact ground-truth quantity. \Cref{app: synth data - logical operators} validates the logical operators \textsc{and}, \textsc{or} and \textsc{not}, alongside an illustration of the differences between continuous and discrete CFG. All experiments are run for both masking and uniform noise.
\subsection{Dataset construction}
\label{app: synth data - set up}
We use binary sequences of length $L=20$ over the tokens ``\textit{A}'' and ``\textit{B}'' visualized in red and blue, with either two or four classes specifiable. Classes are defined by the majority likelihood of one token in a certain region. For two classes the majority token dominates the entire sequence, while for four classes it dominates only the first or the second half, with the other half remaining uniform. The four-class set-up is mainly used to obtain non-mutually-exclusive classes, such that logical operations like \textsc{and} can be verified. This is summarized in \Cref{table: synthetic data likelihoods}.

\begin{table}
\begin{center}
\begin{tabular}{|l|c|c|}
\hline
class & $p_1(x_1^j=A|c)$ for $j\in\mathcal H_1$ & $p_1(x_1^j=A|c)$ for $j\in\mathcal H_2$ \\
\hline
$A$ & \multicolumn{2}{c|}{$p_{\mathrm{A}}$} \\
$B$ & \multicolumn{2}{c|}{$1-p_{\mathrm{A}}$} \\
\hline
$A_1$ & $p_{\mathrm{A}}$ & $1/2$ \\
$B_1$ & $1-p_{\mathrm{A}}$ & $1/2$ \\
$A_2$ & $1/2$ & $p_{\mathrm{A}}$ \\
$B_2$ & $1/2$ & $1-p_{\mathrm{A}}$ \\
\hline
\end{tabular}
\end{center}
\caption{Synthetic data construction. The sequences are split into two halves $\mathcal H_1,\mathcal H_2$ of 10 tokens each, filled in with a majority token whose strength is set by a tunable parameter $p_{\mathrm{A}}$. For two classes both halves share the same majority likelihood. For four classes only one half is dominated by a token while the other is uniform, which increases the overlap between classes and makes logical operations such as \textsc{and} more interesting.}
\label{table: synthetic data likelihoods}
\end{table}
Within a class all token positions are independent, so that $p_1(x_1|c)= p_A^{N_A}(1-p_A)^{N_B}$ with $N_A$ and $N_B$ the number of ``\textit{A}'' and ``\textit{B}'' tokens. With this set-up every sequence has a non-zero likelihood under every class: even a sequence consisting solely of ``\textit{B}'' can be sampled by the ``\textit{A}'' model. The corruption also acts independently per position, so the noisy marginals factorize as well. With $\kappa_t$ the noise schedule and $V=2$ data tokens, uniform noise gives
\begin{equation}
    p_t(x_t^j=A|c)=\kappa_t p_A+\tfrac{1-\kappa_t}{V}, \qquad p_t(x_t^j=B|c)=\kappa_t (1-p_A)+\tfrac{1-\kappa_t}{V},
    \label{eq: synth marginals}
\end{equation}
while for masking a position is $A$ with probability $\kappa_t p_A$, $B$ with probability $\kappa_t(1-p_A)$ and \texttt{MASK} with probability $1-\kappa_t$. The conditional marginal $p_t(x_t|c)$ is the product over positions. This also gives the conditional predictors: under masking a masked position simply predicts token ``\textit{A}'' with $p_A$ and token ``\textit{B}'' with $1-p_A$. Under uniform noise the observed token $x_t^j$ is not a reliable copy of the clean token, since with probability $1-\kappa_t$ it was replaced by a uniformly random one, so the conditional predictor weighs the class prior against how well each candidate explains the observation: $p(x_1^j=a|x_t^j,c)\propto p_1(x_1^j=a|c)\,p_t(x_t^j|x_1^j=a)$ with $p_t(x_t^j=v|x_1^j=u)=\kappa_t\mathbb{1}[v=u]+\tfrac{1-\kappa_t}{V}$, where $p_1(x_1^j=A|c)$ is the probability listed in \Cref{table: synthetic data likelihoods} and $p_1(x_1^j=B|c)$ its complement. The normalization is exactly the marginal of Eq.~\ref{eq: synth marginals}, and the prediction interpolates between the class prior $p_1(x_1^j|c)$ at $\kappa_t=0$ (pure noise, no information) and the observed token at $\kappa_t=1$ (no noise).
Now to obtain an unconditional model that correctly samples from this dataset, two options can be used. The first is a summation over the mixture dataset: construct all $2^{20}\approx1\text{M}$ sequences, assign each the average likelihood under the conditional models, $p_1(x_1)=\sum_c p(c)\,p_1(x_1|c)$, and loop over them. The noisy marginal and the unconditional denoiser are then obtained by explicit summation, with $p_t(x_t|x_1)$ the corruption kernel:
\begin{equation}
    p_t(x_t)=\sum_{x_1} p_1(x_1)\,p_t(x_t|x_1), \qquad p(x_1^j=a|x_t)=\frac{\sum_{x_1:\,x_1^j=a} p_1(x_1)\,p_t(x_t|x_1)}{p_t(x_t)}.
    \label{eq: synth enumeration}
\end{equation}
The second, far simpler option is to define the unconditional model as the mixture of all conditionals, i.e., an \textsc{or} gate over the classes, $p(x_1^j|x_t)=\sum_c p(c|x_t)\,p(x_1^j|x_t,c)$. The sole difficulty is the posterior $p(c|x_t)$, which is perfectly tractable through Bayes' rule and Eq.~\ref{eq: synth marginals}, since $p_t(x_t)=\sum_c p(c)\,p_t(x_t|c)$. Both constructions are exactly equivalent, which we also confirm numerically using the enumeration. In all further experiments we use the \textsc{or}-gate version, as it does not require an exhaustive summation over the $2^L$ sequences.

\subsection{Validation of posterior estimation}
\label{app: synth data - posterior validation}
The online posterior estimation can be compared to the exact quantity, given by the ratio of the conditional and unconditional marginals of Eq.~\ref{eq: synth marginals}:
\begin{equation}
    p_t(c|x_t) = p(c) \frac{p_t(x_t|c)}{p_t(x_t)}
\end{equation}
The posterior can also be tracked with the log-posterior update derived in the main body:
\begin{equation}
    \log p_t(c|x_0,...,x_t) = \log p_{t-1}(c|x_0,...,x_{t-1})+\log p_{t-1}(x_t|x_{t-1},c)-\log p_{t-1}(x_t|x_{t-1})
\end{equation}
This equation still allows for some freedom, namely in the choice of transition likelihoods, or sampler. In this work we mainly focus on two specific cases the kinetic optimal discrete flow matching sampler and the stochastic ancestral sampler. Also, note that it is perfectly sound to use one choice for sampling and another for tracking the posterior. As shown in \cref{app: theory - posterior history dependence}, if a memoryless stochastic sampler is used to compute the transition likelihoods, the following holds $p_t(c|x_0,...,x_t)=p_t(c|x_t)$: the entire posterior information is contained in the least noisy point of the trajectory, i.e., the current state. For \textit{masking} this always holds, as tokens remain unmasked once revealed, and the flow matching and ancestral samplers coincide. For \textit{uniform} noise it holds only if the denoising kernel is derived as the reverse of the corruption kernel, i.e., for an ancestral sampler.
\begin{figure}[H]
    \centering
    \begin{subfigure}[t]{0.48\textwidth}
        \centering
        \includegraphics[width=\linewidth]{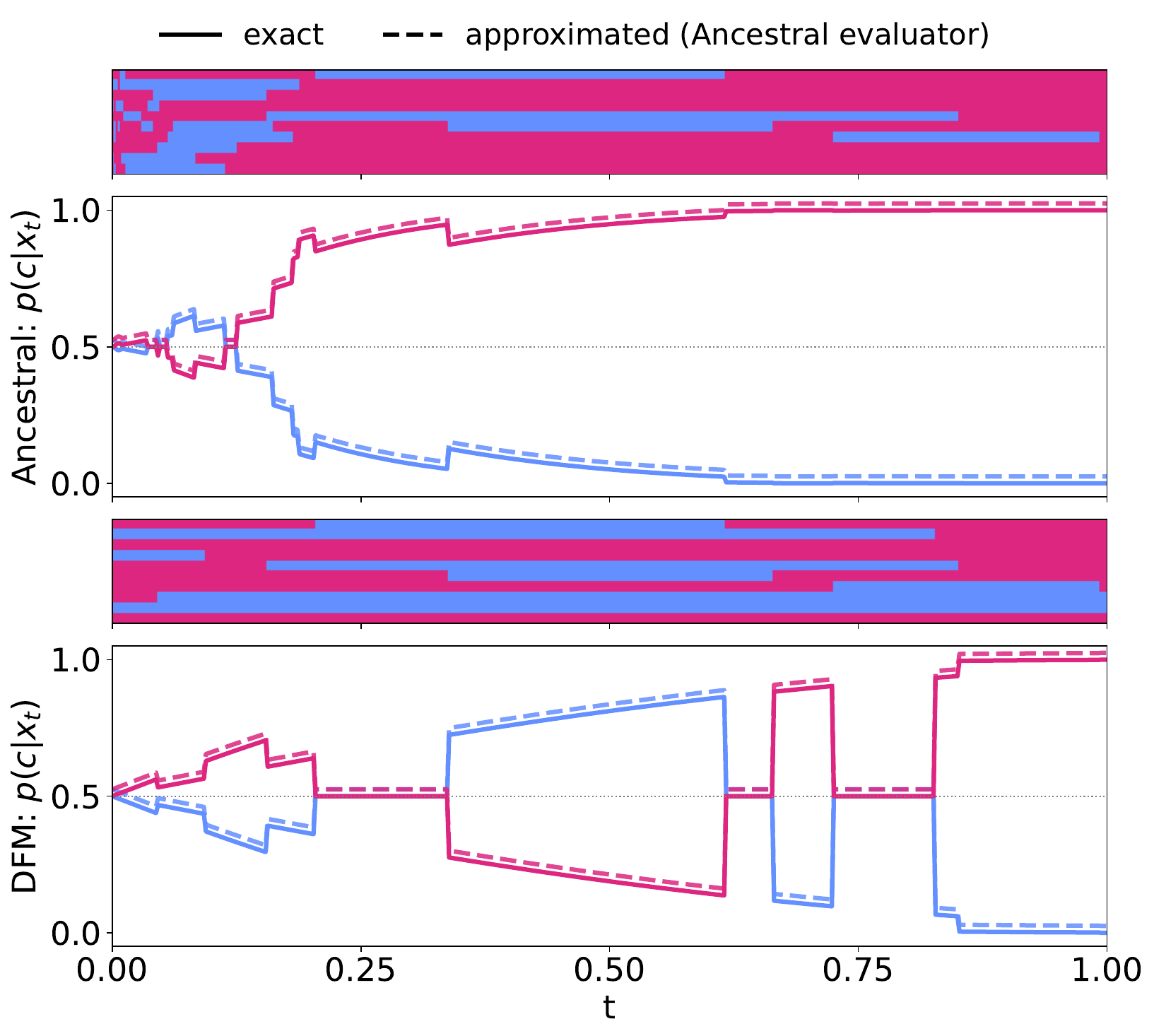}
        \caption{Using ancestral transition likelihoods}
        \label{subfig (app): posterior approx with ancestral transitions}
    \end{subfigure}
    \hfill
    \begin{subfigure}[t]{0.48\textwidth}
        \centering
        \includegraphics[width=\linewidth]{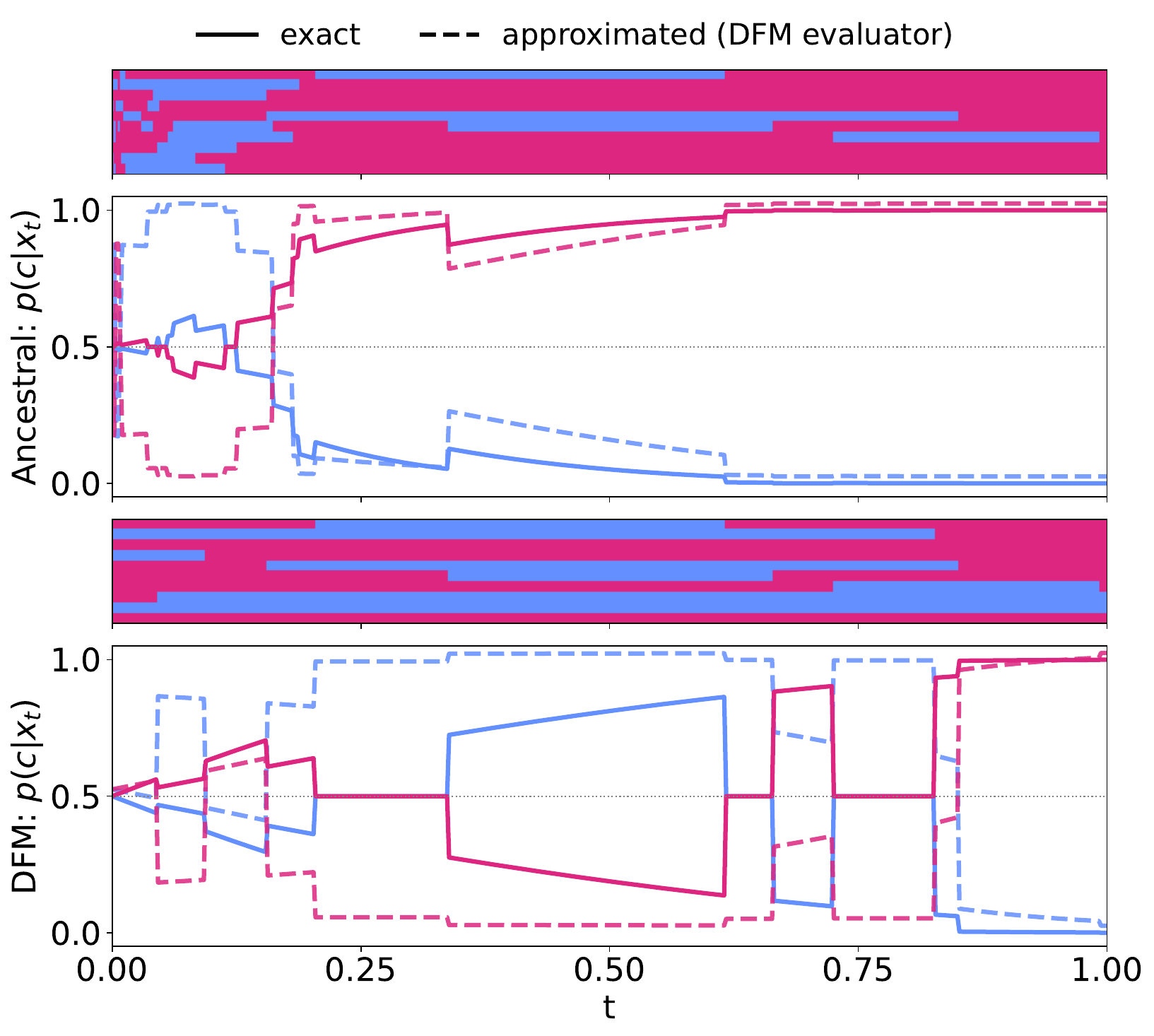}
        \caption{Using dfm transition likelihoods}
        \label{subfig (app): posterior approx with dfm transitions}
    \end{subfigure}
    \caption{Illustration of the online posterior evaluation under different samplers and evaluators. 
    In (a), samples are taken from an ancestral sampler at the top and a dfm sampler at the bottom and the posterior likelihoods are updated using the ancestral transition likelihoods. In this case the tracked posterior is equal to the ground truth posterior. In (b), the exact same samples are shown but with this time the DFM transition likelihoods used for the posterior estimation, in which case the tracked quantity is not equal to the exact one anymore.}
    \label{fig (app): posterior approx}
\end{figure}

\Cref{subfig: Illustrative posterior estimations analytical} shows that the tracked posterior is exact for both noise types when the ancestral sampler is used. In \Cref{subfig (app): posterior approx with ancestral transitions} we show that if the posterior likelihood is computed using the transition likelihoods predicted by the ancestral sampler, the tracked quantity is equal to the ground truth quantity regardless of whether an ancestral or dfm sampler is used to generate the sample. On the other hand in \Cref{subfig (app): posterior approx with dfm transitions}, we show that if the transition likelihoods are computed using the flow matching sampler the tracked and exact quantities are no longer equal, i.e., there is a real difference between $p_t(c|x_0,...,x_t)$ and $p_t(c|x_t)$. These do however remain strongly correlated, with for instance the dashed and full curves going up and down together, simply with the wrong scale. An important note is that in practice, on real trained models, we find that the flow matching sampler yields to more stable and accurate posterior estimation. We hypothesize that this discrepancy between theory and practice is also caused by the fact that the deterministic ODE shows smoother less stochastic trajectories, such that the posterior information is easier to discern for the trained model. A precise understanding of this phenomenon is however left as future work.

\subsection{Validation of the logical operators}
\label{app: synth data - logical operators}

\begin{figure}
\begin{center}
\includegraphics[width=\textwidth]{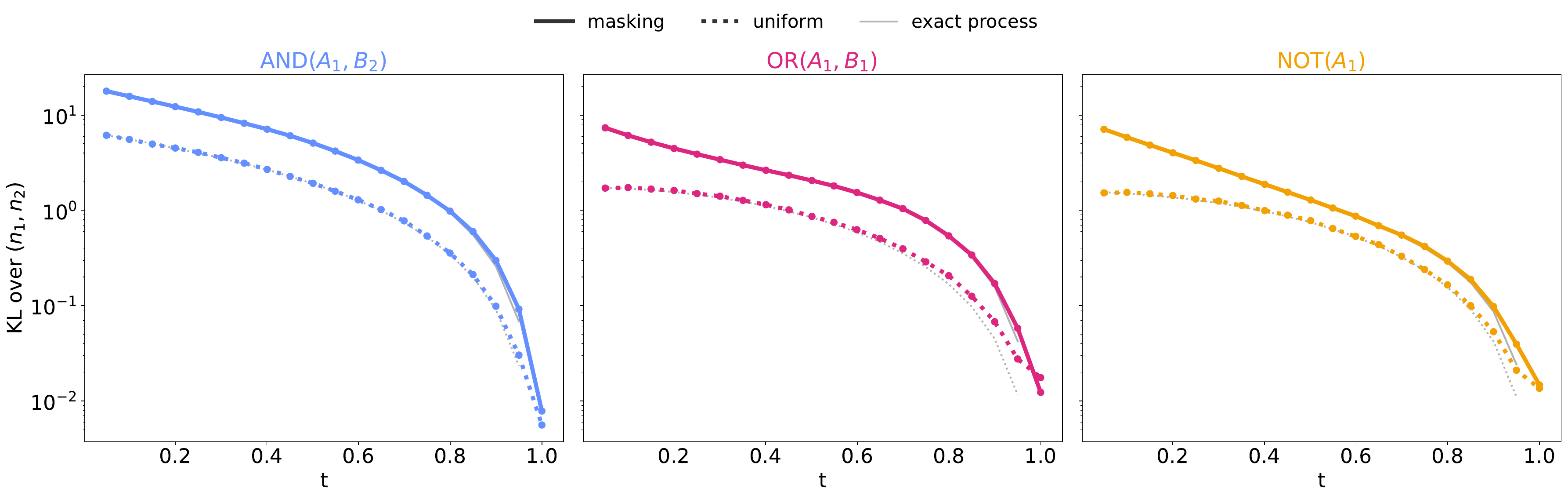}
\end{center}
\caption{Evolution of the KL divergence between the distribution of $(n_1,n_2)$ of the noisy state $x_t$ and that of the clean target, for \textsc{and}$(A_1,B_2)$, \textsc{or}$(A_1,B_1)$ and \textsc{not}$(A_1)$, under masking (full) and uniform (dotted) noise. In all cases the derived gates sample from the desired target distribution, illustrated by the KL distance between generated and reference samples going to zero.}
\label{fig: synth gates evolution}
\end{figure}

To validate the logical operators we compare the generated samples to the exact target distribution of each gate. Comparing full sequences would require far too many samples, so we instead count the number of tokens ``\textit{A}'' in the first and in the second half, $(n_1,n_2)$. As all classes are exchangeable within a half, each target (the \textsc{and}-law, and the mixtures underlying \textsc{or} and \textsc{not}) depends on a sequence only through these counts, such that its exact distribution $T(n_1,n_2)$ is easily computed. We then measure the KL divergence between the empirical distribution of $(n_1,n_2)$ over $M$ generated samples and $T$. We do this not only for the final samples, but also for the noisy states $x_t$ at intermediate noise levels, such that \cref{fig: synth gates evolution} shows how the generated distribution approaches the target as denoising progresses.

For the \textsc{and} gate it is important to also keep the forward kernel in the predictor. The exact data-level predictor for the \textsc{and} gate is
\begin{equation}
    \log p(x_1|x_t) = \sum_{i=1}^N \log p(x_1|x_t,c_i) + (1-N)\log p_t(x_t|x_1) + \mathrm{const},
    \label{eq: synth AND kernel}
\end{equation}
and not simply the sum of the logits of the conditional denoisers, as each of these already contains the forward kernel $p_t(x_t|x_1)$ once, while the target should contain it only once. For \textit{masking} the additional term does not depend on $x_1$ and drops out of the softmax. For \textit{uniform} noise however $p_t(x_t^j|x_1^j)$ favors the observed token, such that the term does depend on $x_1$ and has to be kept. As it only involves the forward kernel it is known exactly, without needing $p_1$ or the model. Importantly for a CFG-like mixture of conditional and unconditional distributions this factor does not exist as there, the base distribution is multiplied with a ratio of conditional and unconditional distributions. As such, this factor appears for both numerator and denominator and cancels out cleanly. The other two gates, \textsc{or} and \textsc{not}, are weighted averages of the conditional denoisers and require no such correction.

\section{Additional details regarding image generation}

In this section additional details and results on the image modality are provided. First, \cref{app: image - training details} describes the architecture and the training hyperparameters. Second, \cref{app: image - model ablations} shows ablations on the model performance, namely FID, precision, recall and C2ST as a function of the guidance scale and of the number of sampling steps. Then, \cref{app: image - posterior estimation} the posterior estimations is analyzed in detail. Finally, \cref{app: image - logical gates} contains an in depth discussion of the results obtained using the \textsc{not} gate.

\subsection{Training details}
\label{app: image - training details}

Our image models are based on the DiT architecture \citep{DiT}, adapted to discrete tokens. Every pixel value is a token from a vocabulary of 256 values (plus a mask token for masking diffusion), such that an MNIST image is a sequence of 784 tokens. The tokens are embedded, processed by transformer blocks and decoded into a categorical distribution over the 256 pixel values for every pixel. The noise level $t$ and the class label are provided through adaptive layer normalization (AdaLN-Zero), where the label embedding has an additional entry for the unconditional case. The model is trained to predict the clean image with a cross-entropy loss, which for masking is only computed on the masked tokens and for uniform noise on all tokens. Noise levels are sampled in a stratified way, using a linear noise schedule. Conditional models are trained with label dropout, while unconditional models are trained separately by always dropping the label. As is typical for diffusion models, an exponential moving average (EMA) of the weights is used for sampling. The hyperparameters, listed in \Cref{table: image training hyperparameters}, were set to standard values \citep{MD4} and not tuned per noise type.

\begin{table}[H]
\begin{center}
\begin{tabular}{|l|c|}
\hline
Architecture & DiT with AdaLN-Zero \\
Parameters & 7.6M \\
Depth / hidden size / heads & 6 / 256 / 4 \\
Sequence length / vocabulary & 784 / 256 (+ mask token) \\
\hline
Optimizer & AdamW \\
Learning rate & $4\cdot10^{-4}$ \\
Betas / weight decay & (0.9, 0.999) / 0.01 \\
Warmup steps / schedule & 1000 / cosine decay \\
Batch size & 512 \\
Training steps & 200k \\
Gradient clipping & 1.0 \\
EMA decay & 0.9999 \\
Label dropout & 0.15 \\
\hline
Noise schedule & linear \\
Noise level sampling & stratified in $[0,1]$ \\
Loss weighting & none \\
\hline
\end{tabular}
\end{center}
\caption{Hyperparameters of the MNIST models. The unconditional models are trained separately with label dropout 1.0, i.e., without ever seeing the label. The cosine decay is defined over a horizon longer than the training run, so that the learning rate only decays to about $3\cdot10^{-4}$ at the end of training.}
\label{table: image training hyperparameters}
\end{table}

\begin{figure}[H]
    \centering
    \includegraphics[width=0.9\linewidth]{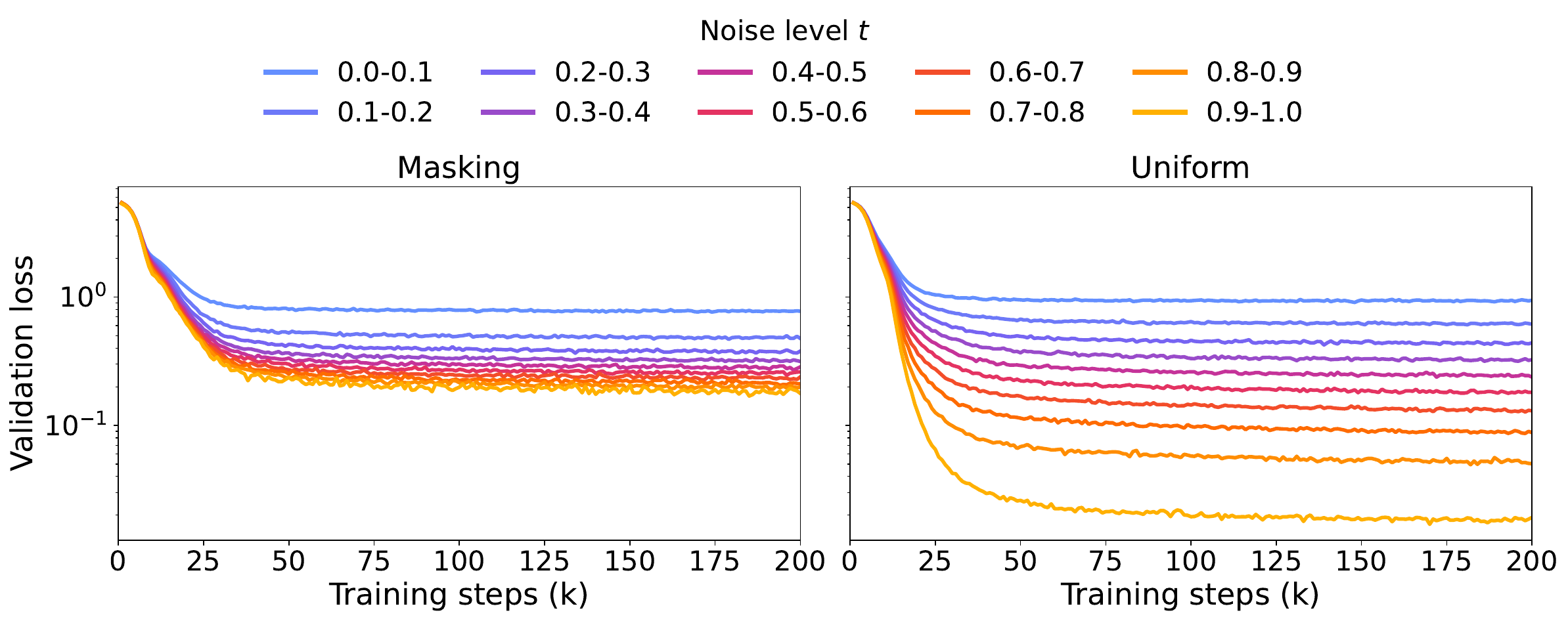}
    \caption{Validation loss during training of the MNIST models, averaged over bins of noise levels $t\in[0,0.1],[0.1,0.2],\dots$, for masking (left) and uniform noise (right).}
    \label{fig: mnist training curves}
\end{figure}

\Cref{fig: mnist training curves} shows the validation loss per noise level during training. In both cases the loss decreases with $t$, as less information has to be inferred at lower noise levels. The highest noise levels converge first, presumably because they are dominated by uncertainty that cannot be reduced (many images are consistent with a nearly empty input), while the low noise levels keep improving until the end of training. The two noise types are not directly comparable, as masking averages over the masked tokens only. At low noise the loss is only over masked tokens, so the loss stays high, whereas for uniform noise most tokens are already noise-free and thus easy to guess, so the loss goes to zero.
\subsection{Model ablations}
\label{app: image - model ablations}
\begin{figure}[H]
    \centering
    \includegraphics[width=0.98\linewidth]{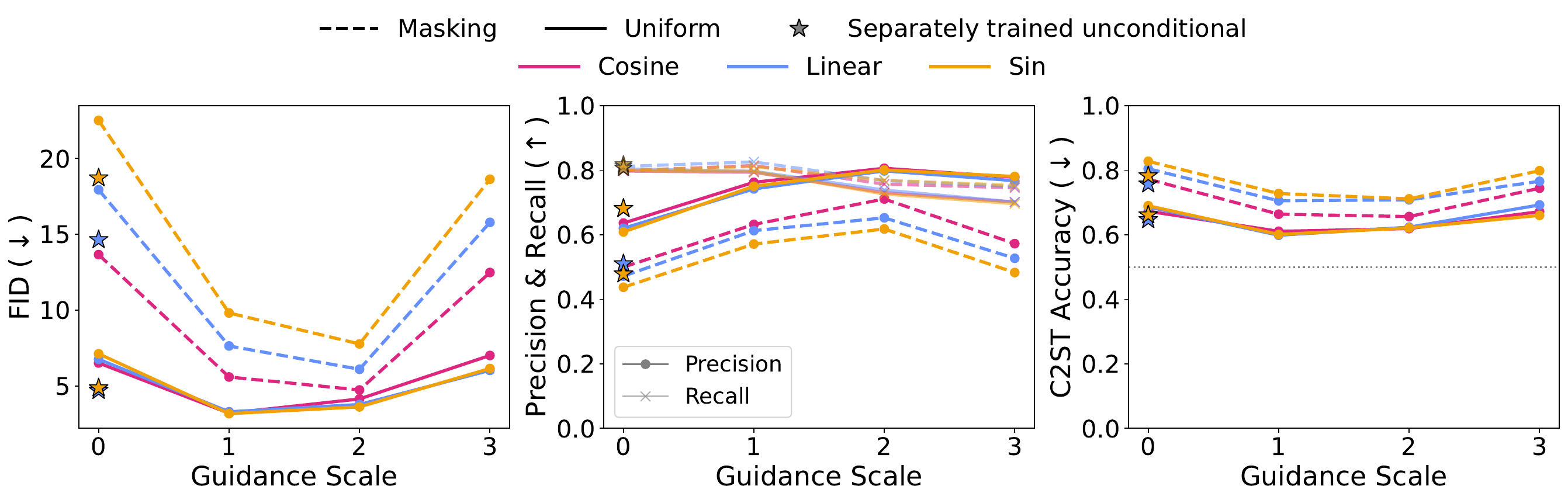}
    \caption{Sweep over the guidance scale for the MNIST-trained model. CFG is implemented as usual in the discrete case, i.e., a linear combination of logits. During inference 128 steps of the specified schedulers were taken. The stars correspond to the separately trained unconditional model.}
    \label{fig: mnist FID ablation guidance sweep}
\end{figure}

\begin{figure}[H]
    \centering
    \includegraphics[width=0.98\linewidth]{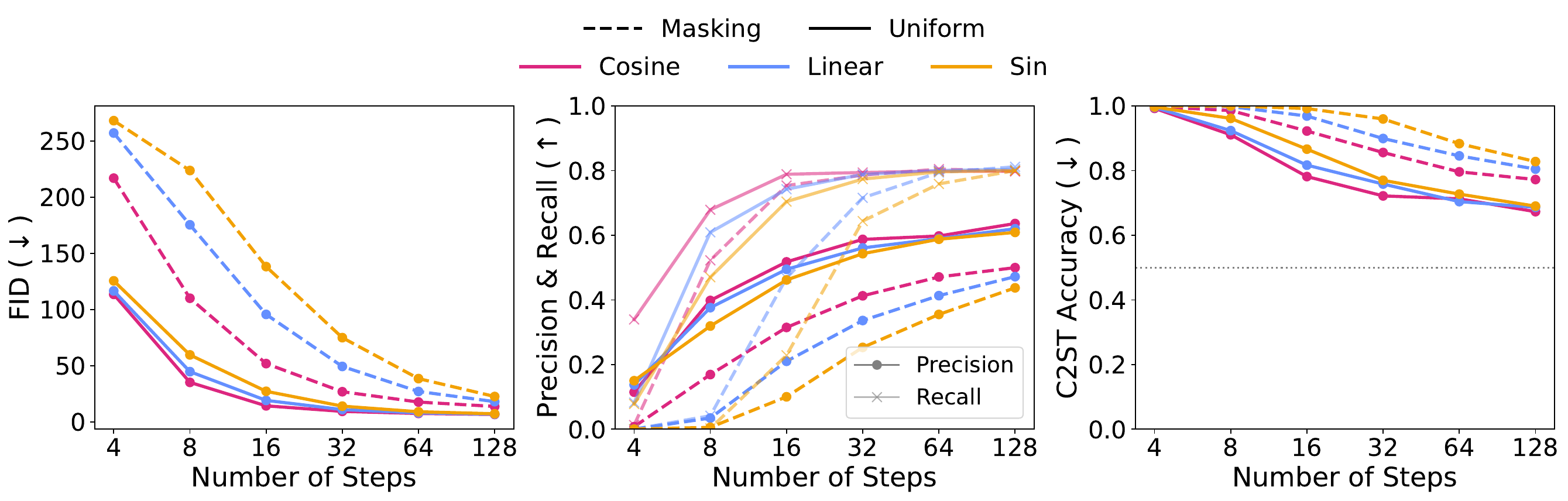}
    \caption{Sweep over the number of sampling steps for the unconditional MNIST-trained model. Masking and uniform models are evaluated under three schedulers; sine (fast at first and then slow), linear (constant sped) and cosine (slow at first and then fast). From left to right, the FID, the Precision and Recall and the C2ST accuracy. }
    \label{fig: mnist FID ablation num steps for uncond}
\end{figure}

\begin{figure}[H]
    \centering
    \includegraphics[width=0.98\linewidth]{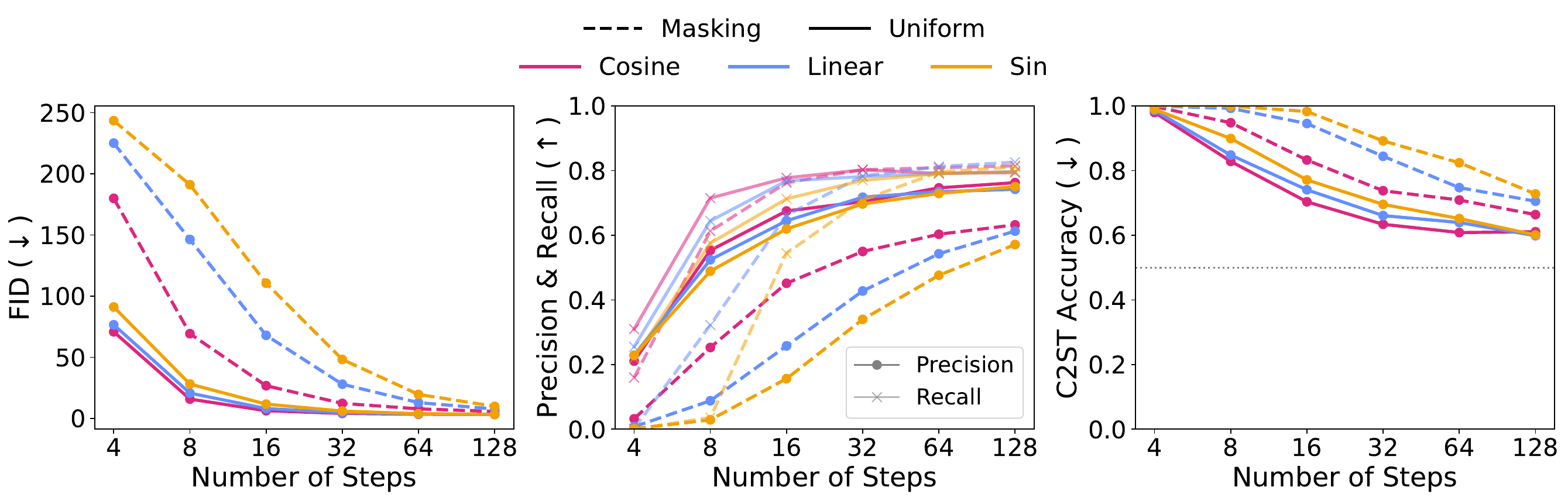}
    \caption{Sweep over the number of sampling steps for the conditional MNIST-trained model. Masking and uniform models are evaluated under three schedulers; sine (fast at first and then slow), linear (constant sped) and cosine (slow at first and then fast). From left to right, the FID, the Precision and Recall and the C2ST accuracy. }
    \label{fig: mnist FID ablation num steps for cond}
\end{figure}

We evaluate the generated images with three metrics, all computed on InceptionV3 features of 4096 generated samples compared to the training set. The Fréchet inception distance (FID) is the Fréchet distance between Gaussian fits to the generated and the real features, and measures overall quality and diversity (lower is better). Precision and recall are k-nearest-neighbor based: precision is the fraction of generated samples that lie within the manifold of the real data (fidelity), recall the fraction of real samples that lie within the manifold of the generated data (diversity). The C2ST is the held-out accuracy of a logistic regression classifier that separates real from generated features, where 0.5 means the two are indistinguishable.
\Cref{fig: mnist FID ablation guidance sweep} shows the effect of the guidance scale on these metrics. Conditional models (scale 1) are clearly better than unconditional ones (scale 0), and unconditional models that are trained separately (stars) are better than the unconditional mode of the jointly trained model. 
\Cref{fig: mnist FID ablation num steps for uncond,fig: mnist FID ablation num steps for cond} show the dependence on the number of sampling steps. The quality increases with the number of steps in all settings. For masking there is a clear ranking of the sampling schedules, with the cosine schedule, which is slow early and fast late, performing best. This is expected: early in the denoising process the tokens have to be placed carefully, while late steps are easy as most of the image is already fixed. For uniform noise the schedule matters less, thanks to the ability to correct earlier mistakes. The differences between schedules shrink as the number of steps grows, as a schedule is only a smarter way to use a given compute budget and not a better model.
Finally, the uniform model outperforms the masking model, especially for a low number of sampling steps. This comparison should be taken with care, as the training hyperparameters were set to standard values and not tuned per method, so they may happen to suit the uniform model better.

As the cosine scheduler (slow at first, fast at the end) outperforms all others all experiments use it during sampling. Furthermore, for maximal accuracy, all experiments were performed at $128$ steps. All metrics in this modality are computed on 4096 samples.
\subsection{Posterior Estimation}
\label{app: image - posterior estimation}
In the context of image generation, the posterior can be properly normalized by marginalizing over all classes. In practice, we achieve this by tracking the log-posterior for each class and subsequently applying a softmax over all classes. This results in a normalized and directly interpretable posterior distribution. Moreover, considering all classes jointly provides a clearer classification signal, as predictions are compared against one another rather than only against an unconditional prediction. Examples similar to the one presented in the main body in \Cref{subfig: Illustrative posterior estimations MNIST}, including an example in which the trajectory is misclassified, are shown in \Cref{fig (app): posterior approx}. In all examples, the posterior concentrates on the final class more rapidly under masking noise than under uniform noise.

\begin{figure}[h]
    \centering
    \begin{subfigure}[t]{0.48\textwidth}
        \centering
        \includegraphics[width=\linewidth]{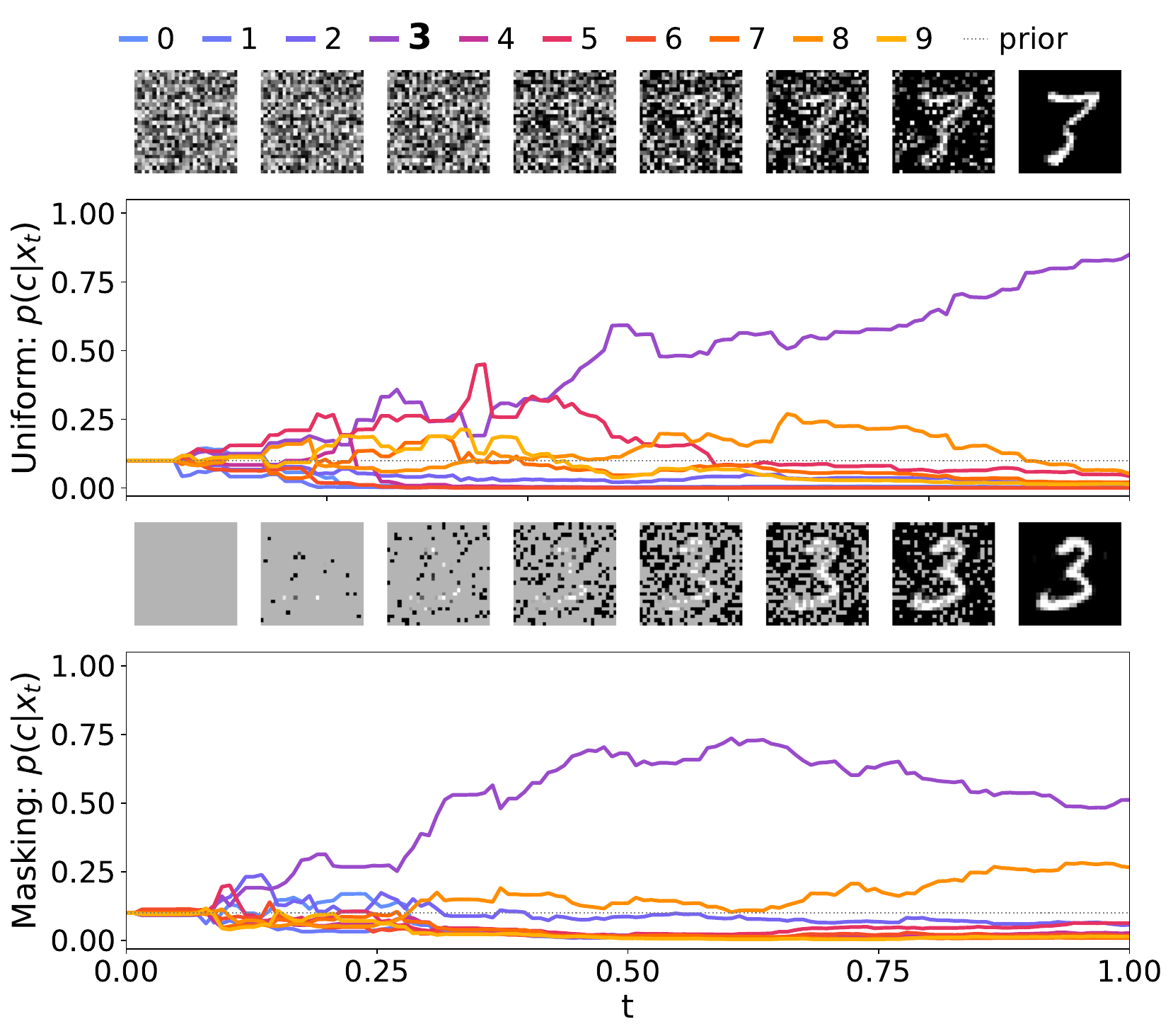}
        \caption{Successful example}
        \label{subfig (app): Illustrative posterior estimations - 3}
    \end{subfigure}
    \hfill
    \begin{subfigure}[t]{0.48\textwidth}
        \centering
        \includegraphics[width=\linewidth]{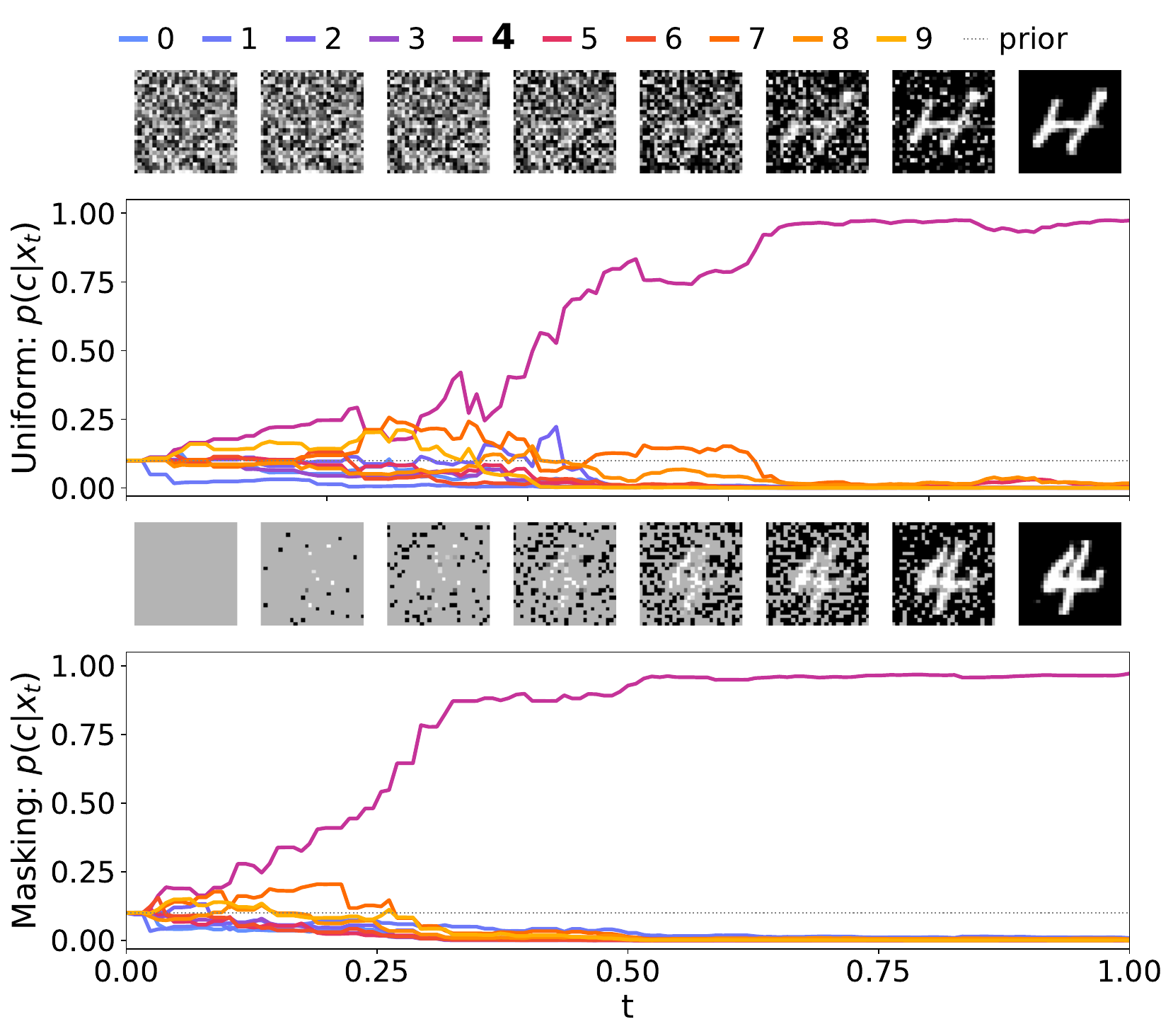}
        \caption{Successful example}
        \label{subfig (app): Illustrative posterior estimations - 4}
    \end{subfigure}
    \begin{subfigure}[t]{0.48\textwidth}
        \centering
        \includegraphics[width=\linewidth]{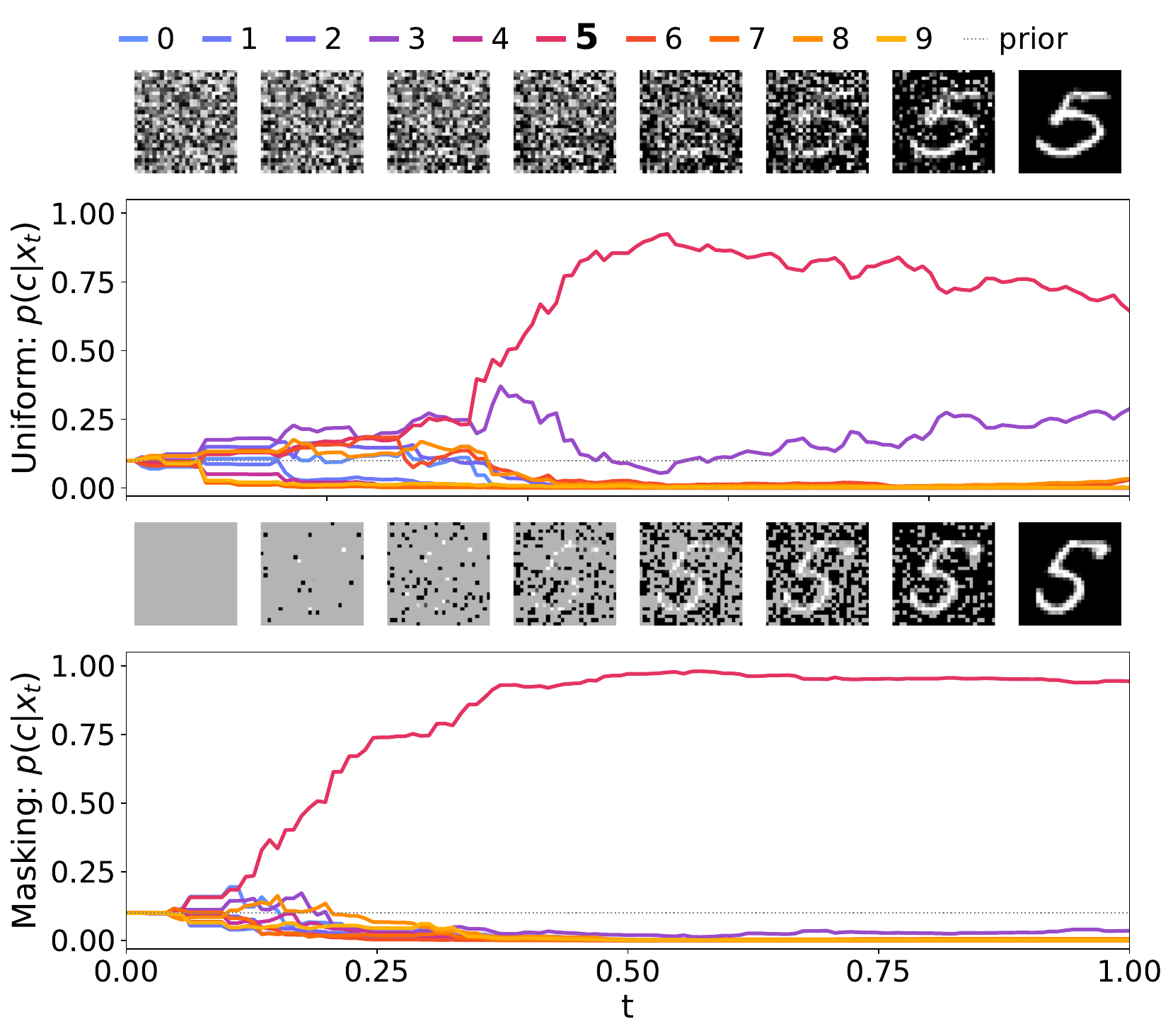}
        \caption{Successful example}
        \label{subfig (app): Illustrative posterior estimations - 5}
    \end{subfigure}
    \hfill
    \begin{subfigure}[t]{0.48\textwidth}
        \centering
        \includegraphics[width=\linewidth]{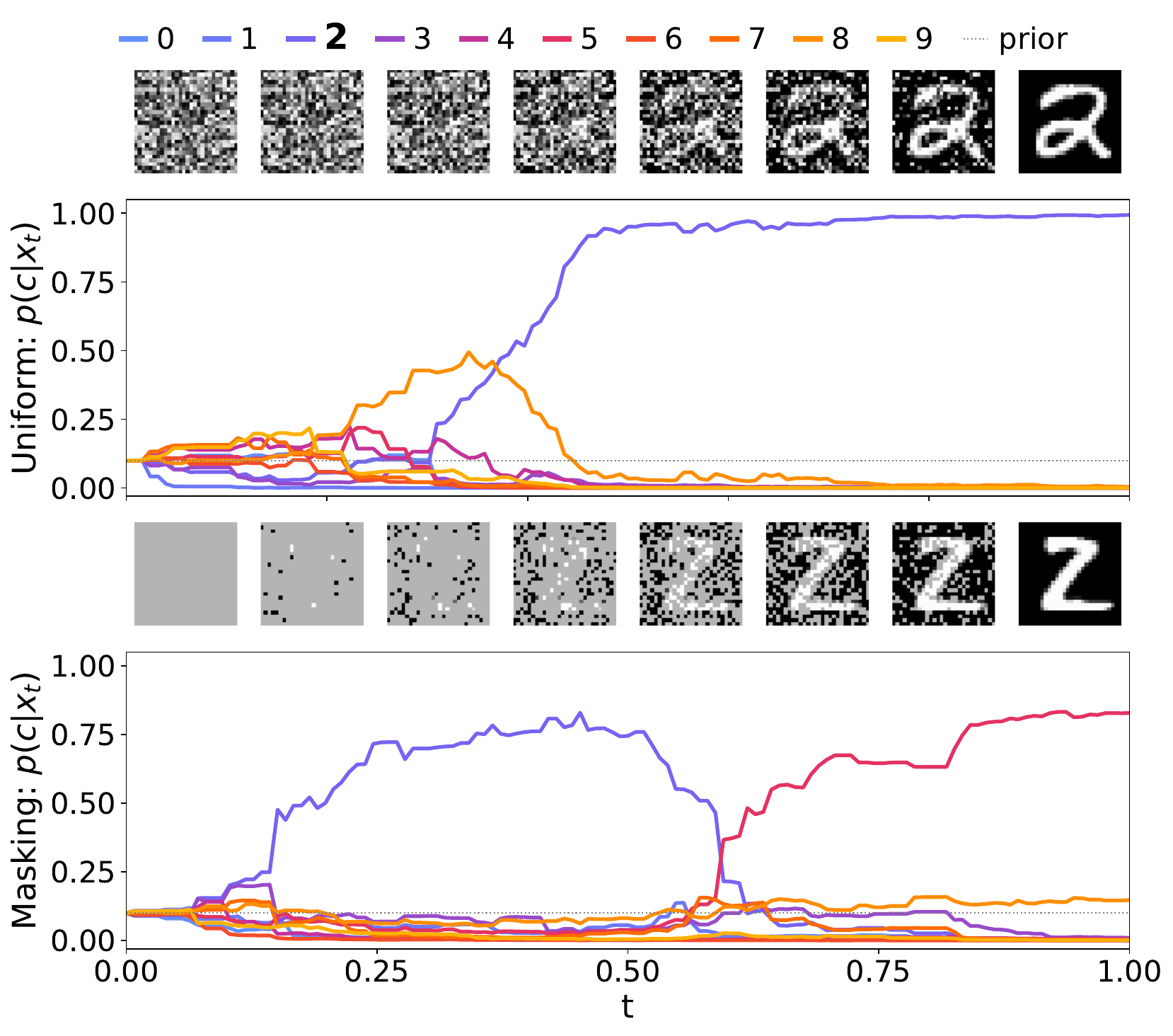}
        \caption{Unsuccessful example}
        \label{subfig (app): Illustrative posterior estimations - wrong 2}
    \end{subfigure}
    \caption{Illustration of the online posterior evaluation. 
    In (a-c), trajectories for which the online posterior estimate recovers the final class for both masking and uniform noise are shown.
    In (d), an example is shown for which the masking model does not recover the correct class label. All shown trajectories use $\tau=0.1$, a value taken from a calibration run on the \textsc{not} gate.} 
    \label{fig (app): Illustrative posterior approx}
\end{figure}
To quantify the effectiveness of the online posterior estimation, we consider two main metrics: one-vs-rest AUC and classification accuracy. Both metrics are evaluated as a function of denoising time, highlighting the main advantage of our approach: the posterior provides a classification signal before generation is complete. Ground-truth labels for the generated samples are obtained using a pretrained ViT classifier \citep{ViTClassifier}. The accuracy is computed as the agreement rate between the top posterior prediction and the class assigned by the ViT. For the AUC, we use the macro-averaged one-vs-rest (OvR) extension of the ROC AUC to the multiclass setting. For each class $c$, samples assigned class $c$ by the ViT are treated as positives and all remaining samples as negatives, while the posterior probability $p_t(c \mid \mathbf{x}_t)$ is used as the ranking score to compute the corresponding binary ROC AUC. The per-class AUCs are then averaged uniformly over the 10 classes. An AUC of $0.5$ corresponds to chance level, which is also the expected value at $t=0$, when no class-specific evidence has yet been accumulated.

\begin{figure}[h]
    \centering
    \begin{subfigure}[t]{0.48\textwidth}
        \centering
        \includegraphics[width=\linewidth]{fig/images/dfm_auc_softmax_vs_time.pdf}
        \caption{AUC}
        \label{subfig (app): AUC}
    \end{subfigure}
    \hfill
    \begin{subfigure}[t]{0.48\textwidth}
        \centering
        \includegraphics[width=\linewidth]{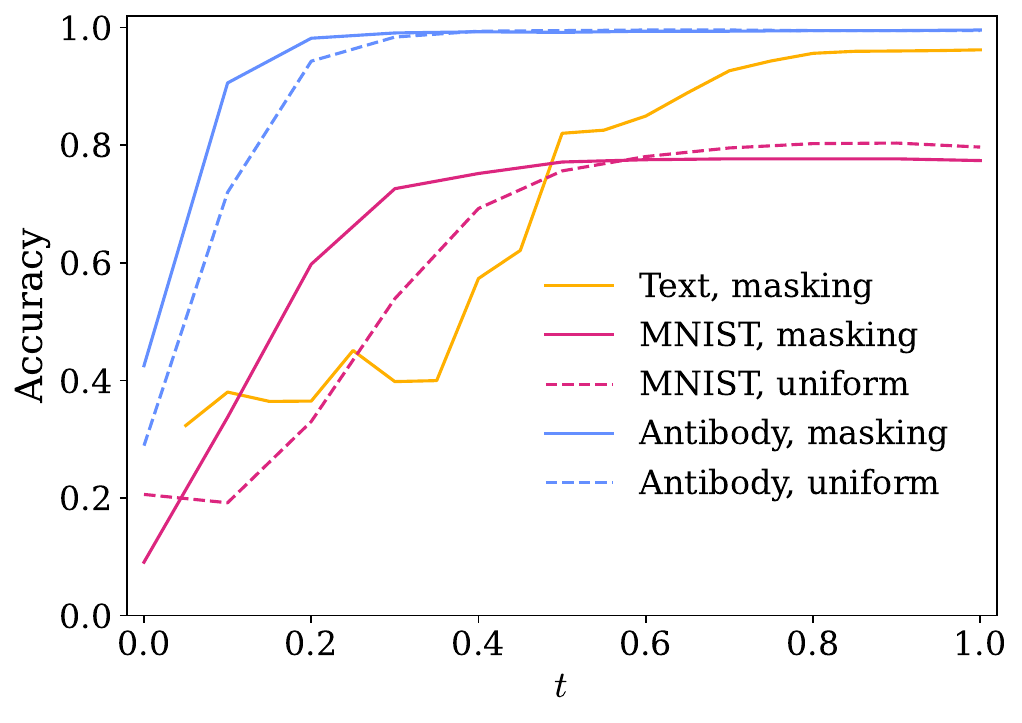}
        \caption{Accuracy}
        \label{subfig (app): Accuracy}
    \end{subfigure}
    \caption{Evaluation of the online posterior estimation for different modalities during denoising. In (a), the OvR AUC. In (b), the accuracy.}
    \label{fig (app): posterior estimation metrics}
\end{figure}

The evolution of both metrics throughout denoising, for the three modalities evaluated in this work, is shown in \cref{fig (app): posterior estimation metrics}. For a given noise schedule, the masking variant consistently accumulates class information more rapidly, which is expected from its inability to modify tokens once they have been unmasked. It is important to emphasize that we do not claim that our posterior provides a perfect classifier in this setting. Rather, its primary advantage is that it provides a proxy for the final class before generation is complete. As shown by both the AUC and accuracy curves, the classification signal largely saturates after approximately half of the generation process. This saturation is likely related to the out-of-distribution nature of the conditional models at low noise levels. As denoising progresses, the conditional marginals increasingly separate, such that evaluations of a model conditioned on one class become less meaningful for states belonging to another class. For example, a model trained exclusively on ``2''s cannot be expected to accurately predict the transition probabilities required to transform a partially denoised ``4'' into what it knows. This limitation occurs primarily once the sample has already been resolved, at which point the posterior has little additional utility. The posterior most informative precisely during the earlier stages of denoising, when class information is still emerging and an early estimate is most useful.
\subsection{Logical operators: \textsc{not} gate}
\label{app: image - logical gates}
\begin{figure}[H]
    \centering
    \includegraphics[width=0.9\linewidth]{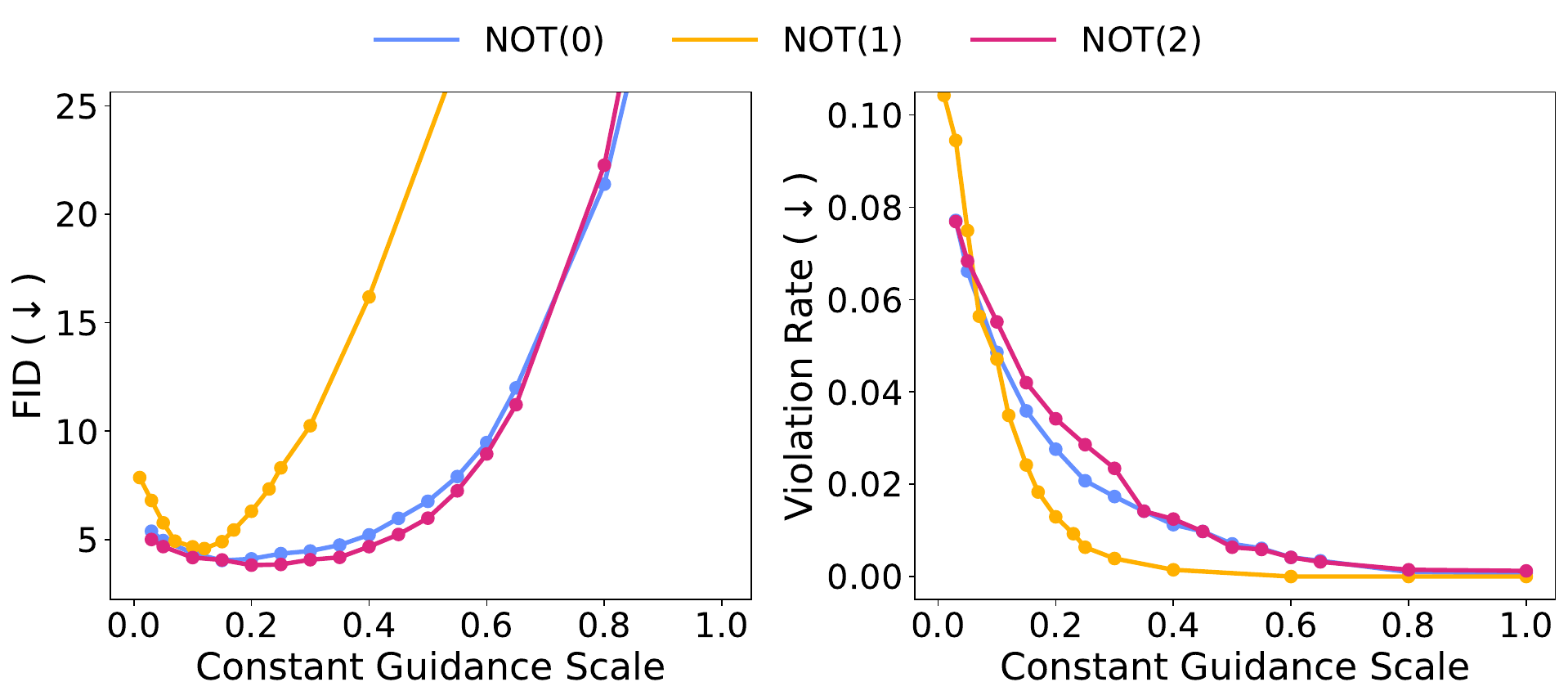}
    \caption{Static negative guidance (negative prompting): FID and violation rate as a function of the constant guidance scale $\lambda$, for \textsc{not}(0), \textsc{not}(1) and \textsc{not}(2).}
    \label{fig: mnist NOT baseline}
\end{figure}

For the \textsc{not} gate, avoiding the generation of a given digit is a trade-off between image quality and safety. We track quality with the FID with respect to reference data that excludes the unwanted digit, and safety with the violation rate, i.e., the fraction of generated images that a ViT classifier assigns to the unwanted digit.
As a static baseline we use negative prompting, a standard method from continuous diffusion, which is CFG with a negative guidance scale $-\lambda$. To stay consistent with the discrete literature, CFG is implemented as usual by an addition of logits.
\Cref{fig: mnist NOT baseline} shows FID and violation rate for a sweep over $\lambda$. A very small scale already suffices to push away from the unwanted digit, while larger scales quickly harm the image quality.

\begin{figure}[H]
    \centering
    \includegraphics[width=0.85\linewidth]{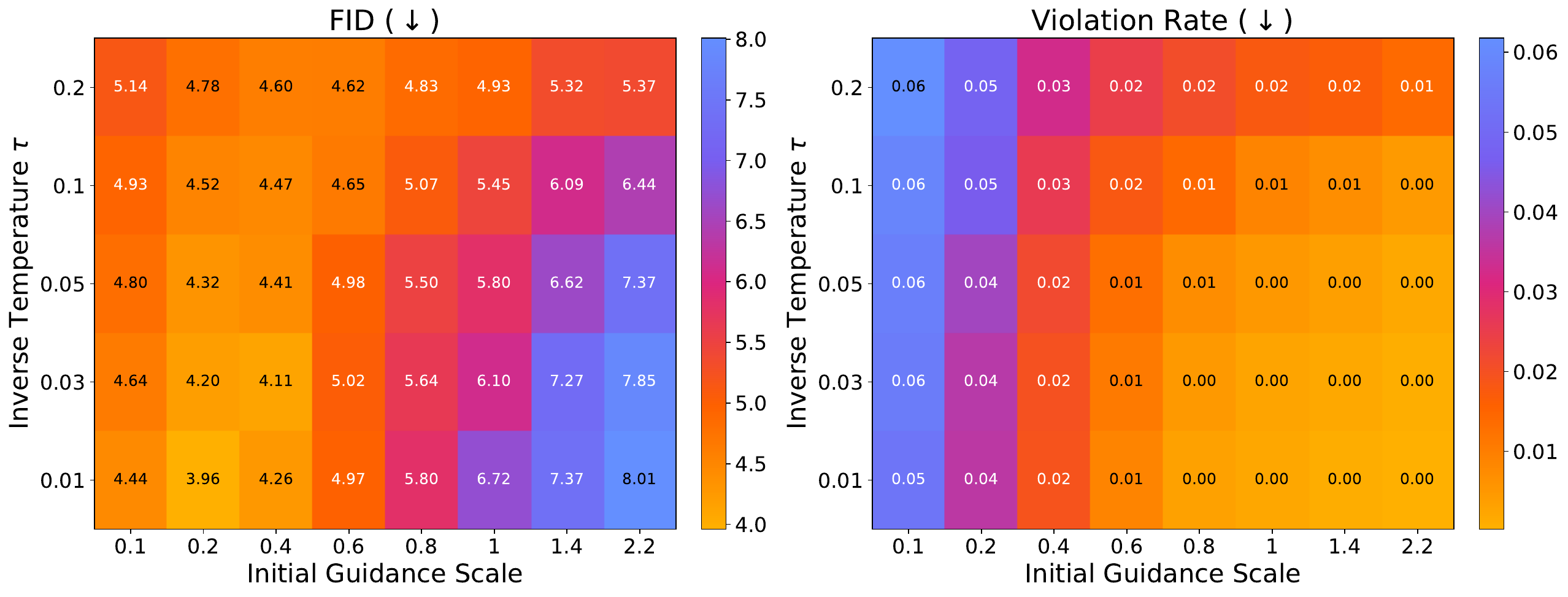}
    \includegraphics[width=0.85\linewidth]{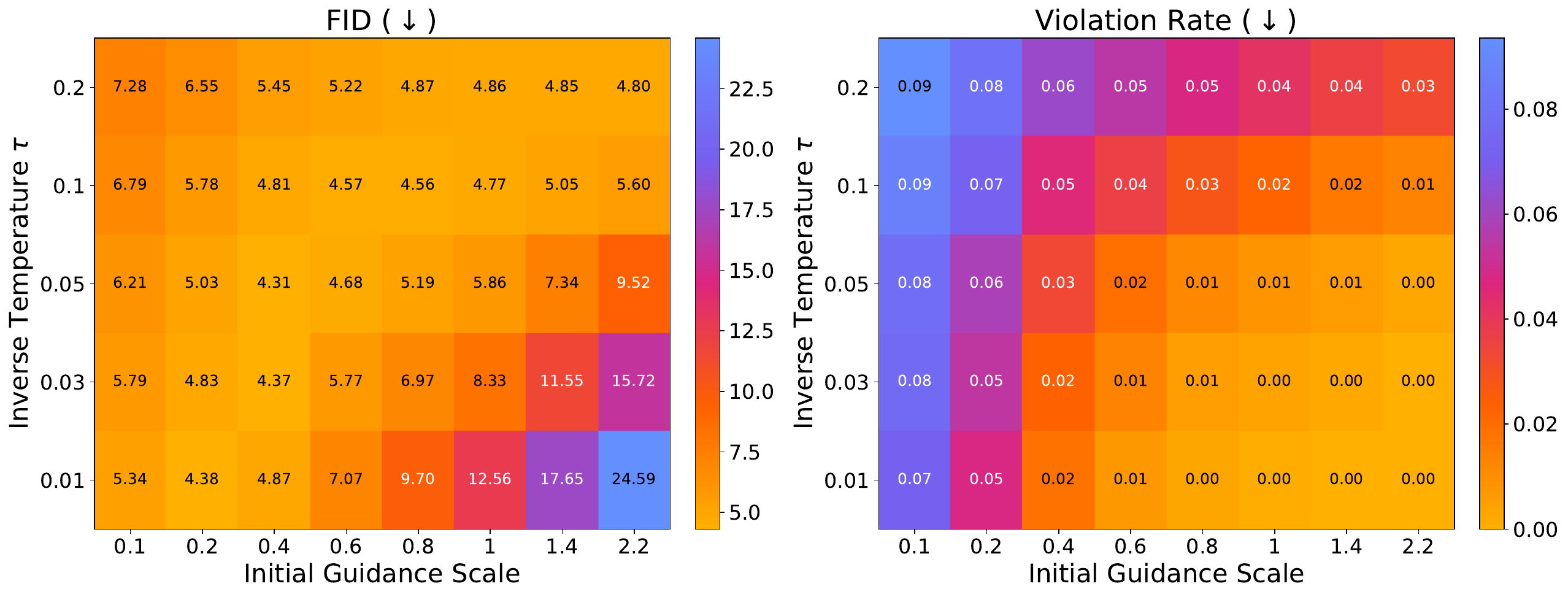}
    \includegraphics[width=0.85\linewidth]{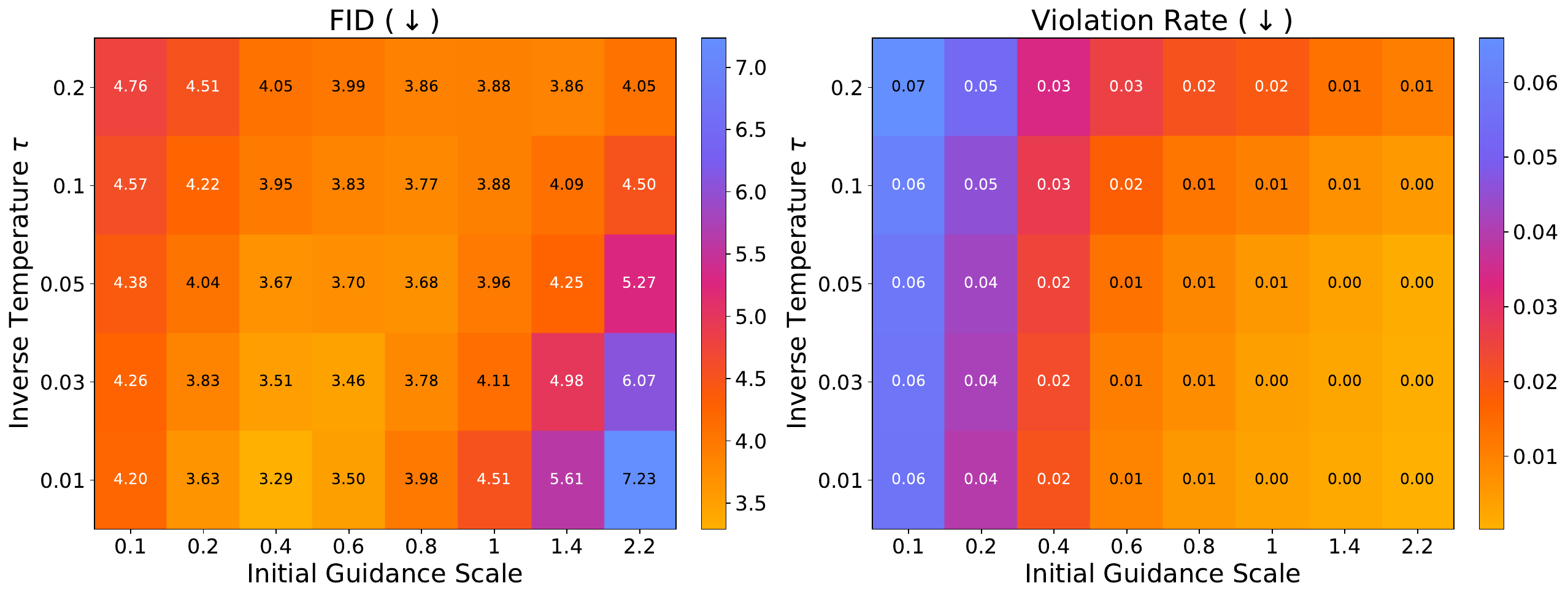}
    \caption{FID and violation rate of the dynamic \textsc{not} gate as a function of the initial guidance scale and the inverse temperature $\tau$, for \textsc{not}(0), \textsc{not}(1) and \textsc{not}(2) (top to bottom). Amber is best, blue is worst.}
    \label{fig: mnist NOT heatmaps}
\end{figure}

Our method has two hyperparameters: the initial guidance scale $s_0$, which mathematically is a parameter of the prior, and the inverse temperature $\tau$ with which the posterior updates are rescaled. In this setting the results were of sufficient quality that the offset parameter $\delta$ was not needed. We perform a grid search over both hyperparameters and analyze both metrics (\Cref{fig: mnist NOT heatmaps}). A larger $s_0$ increases the strength of the guidance, whereas a larger $\tau$ makes it fade faster. For small $\tau$ ($\leq 0.05$) the trade-off is almost independent of $\tau$, while for larger values the guidance fades too quickly and more violations remain. We therefore fix $\tau=0.01$ in all further experiments.

\begin{figure}[H]
    \centering
    \includegraphics[width=0.9\linewidth]{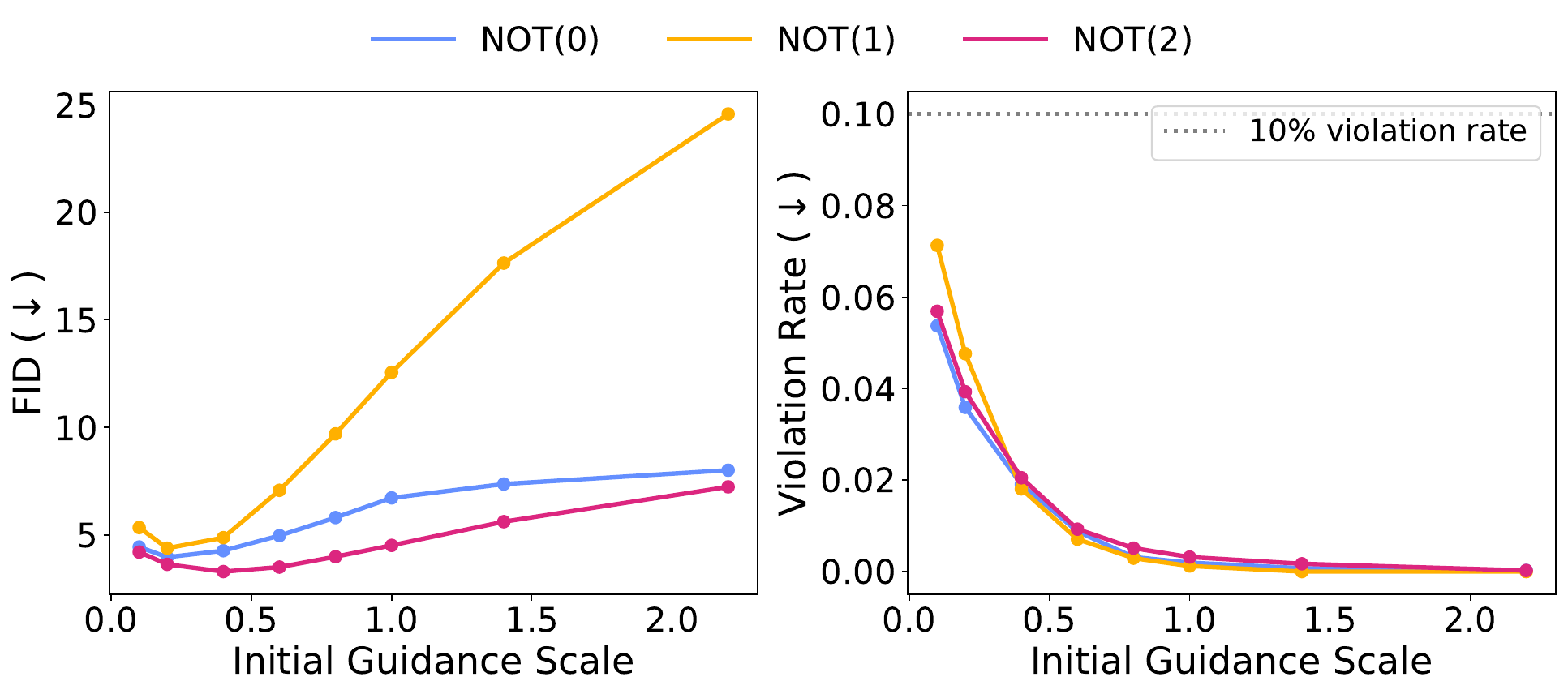}
    \caption{Dynamic negative guidance with $\tau=0.01$: FID and violation rate as a function of the initial guidance scale, for \textsc{not}(0), \textsc{not}(1) and \textsc{not}(2).}
    \label{fig: mnist NOT dynamic}
\end{figure}

\begin{figure}[H]
    \centering
    \includegraphics[width=0.9\linewidth]{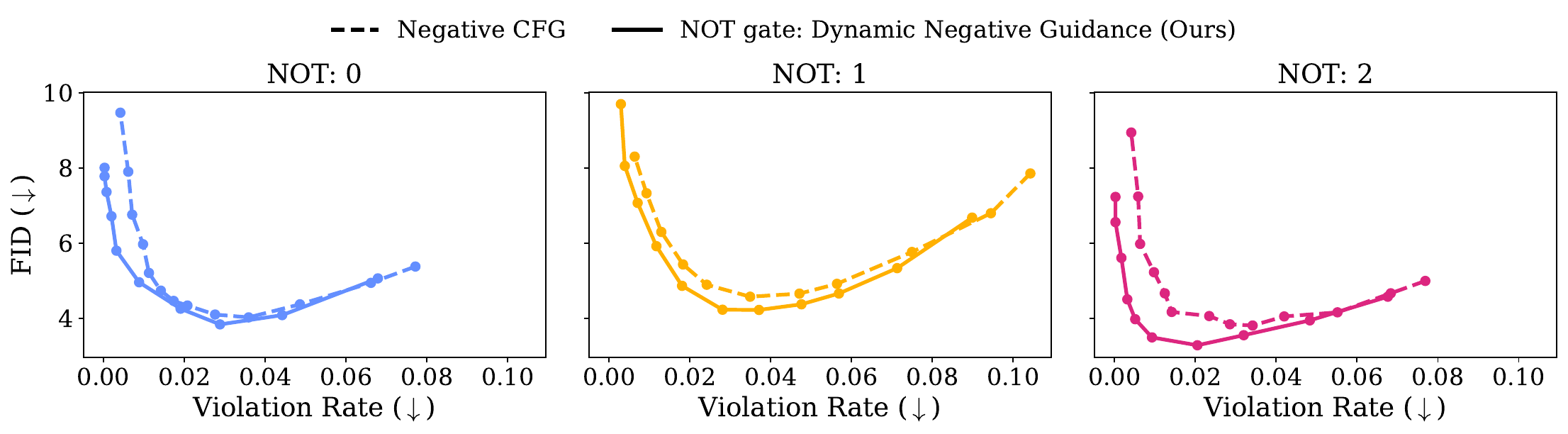}
    \caption{FID against violation rate for dynamic negative guidance (full) and the static baseline (dashed), for \textsc{not}(0), \textsc{not}(1) and \textsc{not}(2).}
    \label{fig (app): mnist NOT pareto}
\end{figure}

\begin{figure}[H]
    \centering
    \includegraphics[width=0.9\linewidth]{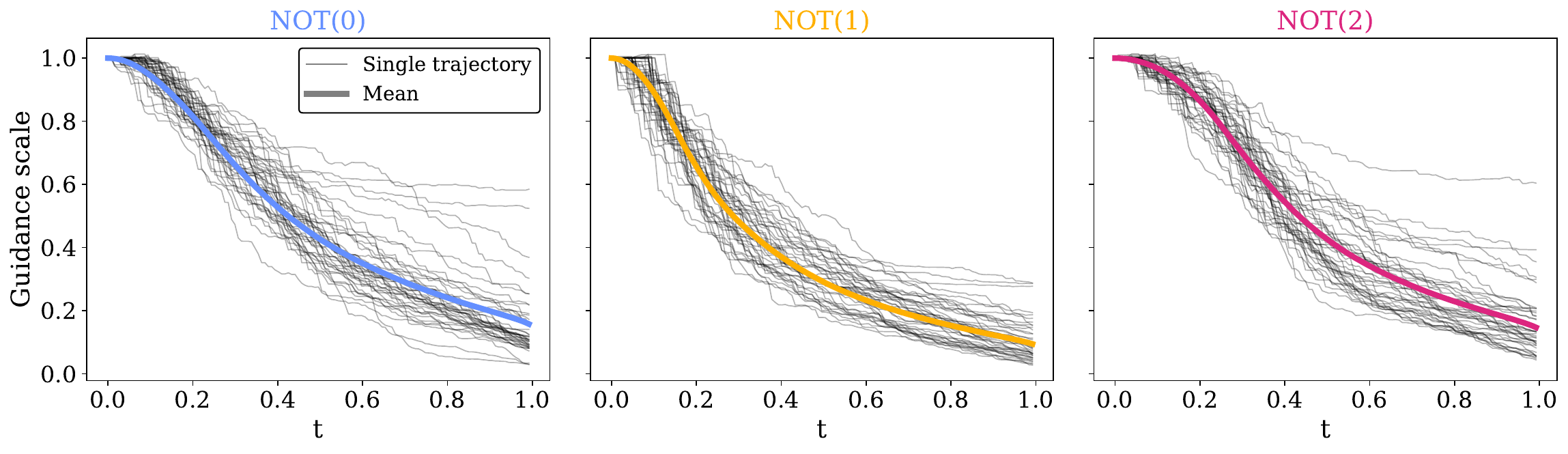}
    \caption{Guidance scale of the dynamic \textsc{not} gate over the sampling process ($\tau=0.01$, initial guidance scale 1): single trajectories (black) and their mean.}
    \label{fig: mnist NOT guidance profile}
\end{figure}

\Cref{fig: mnist NOT dynamic} shows FID and violation rate as a function of $s_0$. The initial guidance scale is an intuitive way of adding more guidance, and results in a trade-off profile very similar to that of the static baseline.
To compare both methods, \Cref{fig (app): mnist NOT pareto} (a copy of the figure in the main body) plots FID against violation rate. The dynamic approach outperforms the static baseline, at the cost of one additional hyperparameter.
Interestingly, the profile of the guidance scale is very consistent (\Cref{fig: mnist NOT guidance profile}). The guidance is mainly needed at the start of the generation and then decays roughly exponentially, where the trajectory only determines how fast. This suggests once more that for discrete image generation the first pixels are decisive, not only for the quality, as reflected by the choice of a cosine scheduler, but also for the class.

All results are shown for the uniform model. The masking model shows qualitatively similar behavior.

\section{Additional details regarding text evaluation}
In this appendix additional details and results on the text modality are provided. First, \Cref{sec:experimental_setup_details_for_text_evaluation} describes the experimental setup for the text modality. Second, \Cref{app:posterior_prompt_ablation} analyzes the posterior estimation in more detail. Finally, \cref{app:text_not} contains an in depth discussion of the results obtained using the \textsc{not} gate.

\subsection{Experimental setup details for text evaluation}
\label{sec:experimental_setup_details_for_text_evaluation}

In our analysis on the text modality, we use the Cognac \citep{chenControllableTextGeneration2022} dataset and benchmark. For the posterior evaluation, we use the dataset as a set of prompts to verify whether the answer is in a certain class. For the \textsc{not} gate evaluation, we use the benchmark, where the (guided) model needs to generate text about one topic (e.g., beverages), but it cannot generate anything related to a subcategory (e.g., wine), the class. An example of a prompt in this dataset is given in \Cref{fig:unconditional_prompt}. There is an instruction to generate about a specific \textcolor{yellow!50!black}{topic}, some \textcolor{purple!70!black}{examples} and finally some \textcolor{brown}{formatting instructions}. We extended the formatting instructions, as the DLM otherwise failed often to format correctly. Note that this is the prompt unconditional to the class, so it does not mention that class.

\begin{figure}[H]
    \centering
    \begin{tcolorbox}
        \textcolor{yellow!50!black}{Generate examples that are under the category of beverage.}\\
        \textcolor{purple!70!black}{
        == red wine is a type of wine made from dark colored grape varieties. == \\
        == pinot noir (french: [pino nwa\textipa{\textinvscr}]) is a red wine grape variety of the species vitis vinifera. == \\
        == turkish coffee is a style of coffee prepared in a cezve using very finely ground coffee beans without filtering. == \\
        }
        \textcolor{brown}{
        Now keep generating.\\
        Only output the continuation in == ... == format, no other text. \\
        Your continuation must be entirely original. \\
        }
        
    \end{tcolorbox}
    \caption{Neutral, unconditional prompt}
    \label{fig:unconditional_prompt}
\end{figure}

\textbf{Answer evaluation.} The evaluation uses WordNet \citep{millerWordNetLexicalDatabase1995} to evaluate the generated text. It leads to three possible outcomes: ``Class'', ``Other classes'' and ``Invalid''.  First, it checks whether the model generated something related to the topic at all, by finding literal matches of WordNet leaf hyponyms (i.e., words that fall under the topic). This method showed some limitations, as literal string matching on the semantically incomplete WordNet had quite some false negatives. To alleviate this issue, we introduce LLM as a judge with the LLM being Qwen3 8B \citep{yangQwen3TechnicalReport2025}. We find that the combination of both (so either a WordNet match or LLM as a judge match) works best for a fair evaluation. This is because WordNet catches niche words, while the LLM catches non-literal matches.
As an example, we take the dev-set ID 31, with topic snack food and class (subtopic) sandwich. WordNet has only four leaves that would be counted on topic, but not in the class: chip, corn chip, potato skin, tortilla chip. While catching niche snacks such as potato skin, it is clear that the category is broader. A generated sample ``== A chocolate bar is a popular snack food, typically made from cocoa and sugar. =='' was counted off topic by WordNet, but on topic by the LLM judge. As such, both methods truly complement each other.
To check whether the model not only generated something on topic, but also in the specified class (subtopic), again WordNet is used with string matching. For the class detection, we found much less false negatives and adding LLM as a judge the same way increased the false positives strongly. Hence, we keep the WordNet only approach for class detection. These two checks create three possible outcomes for generated text. The first one is ``Class", which means the text is on topic and the specified class was generated. The next is ``Other classes", where the text is on topic, but not the specified class. The last option is ``Invalid", where the model fails to generate something on topic at all. As the DLMs also showed some failure modes (such as not respecting the == == format), we count degenerate output also under the class of Invalid, no matter the Wordnet matches.


\textbf{Prompting the model.} As DLM, we choose the popular LLaDA 8B model \citep{LLaDa}. This model is prompted in three different ways. The unconditional prompt, as shown in \Cref{fig:unconditional_prompt}, does not mention the constraint at all. The negative prompt adds an instruction to surely mention the class, as shown in \Cref{fig:negative_prompt}. We call this the negative prompt, because for the \textsc{not} gate, we will guide away from this prompt and thus negatively guide on it. The positive prompt has the opposite instruction, where it is asked to not mention the subcategory. 

We would like to highlight that classical CFG, also in text, intends to strengthen the adherence to the entire prompt, by contrasting with the unconditional prompt filled with only [MASK] tokens. However, for our guidance setup, we argue that the unconditional model is the one that sees the neutral prompt. Just like in image generation, the model already ``knows'' about the task that it should generate natural images, or like our MNIST model knows it should generate numbers between 0-9, the diffusion model that generates inside the data distribution for our task, is the one that is prompted with a neutral prompt that explains the task at hand. As such, we do not use $p_{\text{MASK}}$ as the basis of all models, but rather the neutrally prompted $p_{\text{neutral prompt}}$. As this is our base model, our conditional models also build further on this prompt, only adding to the base prompt.

\textbf{Posterior evaluation setup.} For the posterior evaluation, we combine the three prompts in the main paper. That is: the neutral prompt is used to generate the text, while the posterior of the class is calculated on the negative versus the positive prompt. We use the positive prompt here as it gives an even better posterior signal than other prompt combinations, as shown in \Cref{app:posterior_prompt_ablation}.

For the \textsc{not} gate in text, we explore two different guidance setups and two baselines, with all setups being shown in \Cref{tab:DLM_guidance_methods}. The first setup is that of the \textsc{not} gate in probability space, as outlined in the theory of \Cref{subsec: logical operators}. This model only uses the neutral and negative prompt, so it does not use the ``power'' of the prompted \textsc{not}. In the second setup, negative CFG, we naively apply the well known logit-space CFG, but with a negative sign to guide away from the negative prompt. As baselines, we have the unconditional model, which is unaware of any constraints. Any samples that are already correct under this model, were at no risk of violation in the first place, As such the unconditional model sets a worst case lower bound. The second baseline model is the positive prompted model. This model is interesting to compare with, as it could be argued that this captures a very rough upper bound of how much the model ``knows'' not to generate. 

\begin{table}[H]
    \centering
    \begin{tabular}{|l|l|}
        \hline
        \textsc{not} gate (prob-space) & $p = p_{\text{neutral prompt}} - \lambda_t \cdot ( p_{\text{neg prompt}} - p_{\text{neutral prompt}} )$ \\
        Negative CFG (logit-space) & $l = l_{\text{neutral prompt}} - \lambda \cdot ( l_{\text{neg prompt}} - l_{\text{neutral prompt}} )$ \\
        \hline
        Unconditional & $p = p_{\text{neutral prompt}}$ \\
        Positive prompted & $p = p_{\text{pos prompt}}$ \\
         \hline
    \end{tabular}
    \caption{Contrastive signals used in this paper.}
    \label{tab:DLM_guidance_methods}
\end{table}

\begin{figure}[ht]
    \centering
    \begin{tcolorbox}
        \textcolor{yellow!50!black}{Generate examples that are under the category of beverage. \\}
        \textcolor{purple!70!black}{
        == red wine is a type of wine made from dark colored grape varieties. == \\
        == pinot noir (french: [pino nwa\textipa{\textinvscr}]) is a red wine grape variety of the species vitis vinifera. == \\
        == turkish coffee is a style of coffee prepared in a cezve using very finely ground coffee beans without filtering. == \\
        }
        \textcolor{red}{Now keep generating and include things in the category of wine. \\}
        \textcolor{brown}{
        Only output the continuation in == ... == format, no other text. \\
        Your continuation must be entirely original. \\
        }
    \end{tcolorbox}
    \caption{Negative prompt}
    \label{fig:negative_prompt}
\end{figure}

\subsection{Posterior estimation}
\label{app:posterior_prompt_ablation}
In this appendix, the posterior is analyzed deeper in the following ways: we analyze the impact of the prompt types to compare during posterior estimation, we determine classification accuracy in a few-class scenario and we give more examples on high impact tokens for some prompts.

\begin{figure}[H]
    \centering

    \begin{subfigure}[c]{0.32\textwidth}
        \centering
        \includegraphics[width=\linewidth]{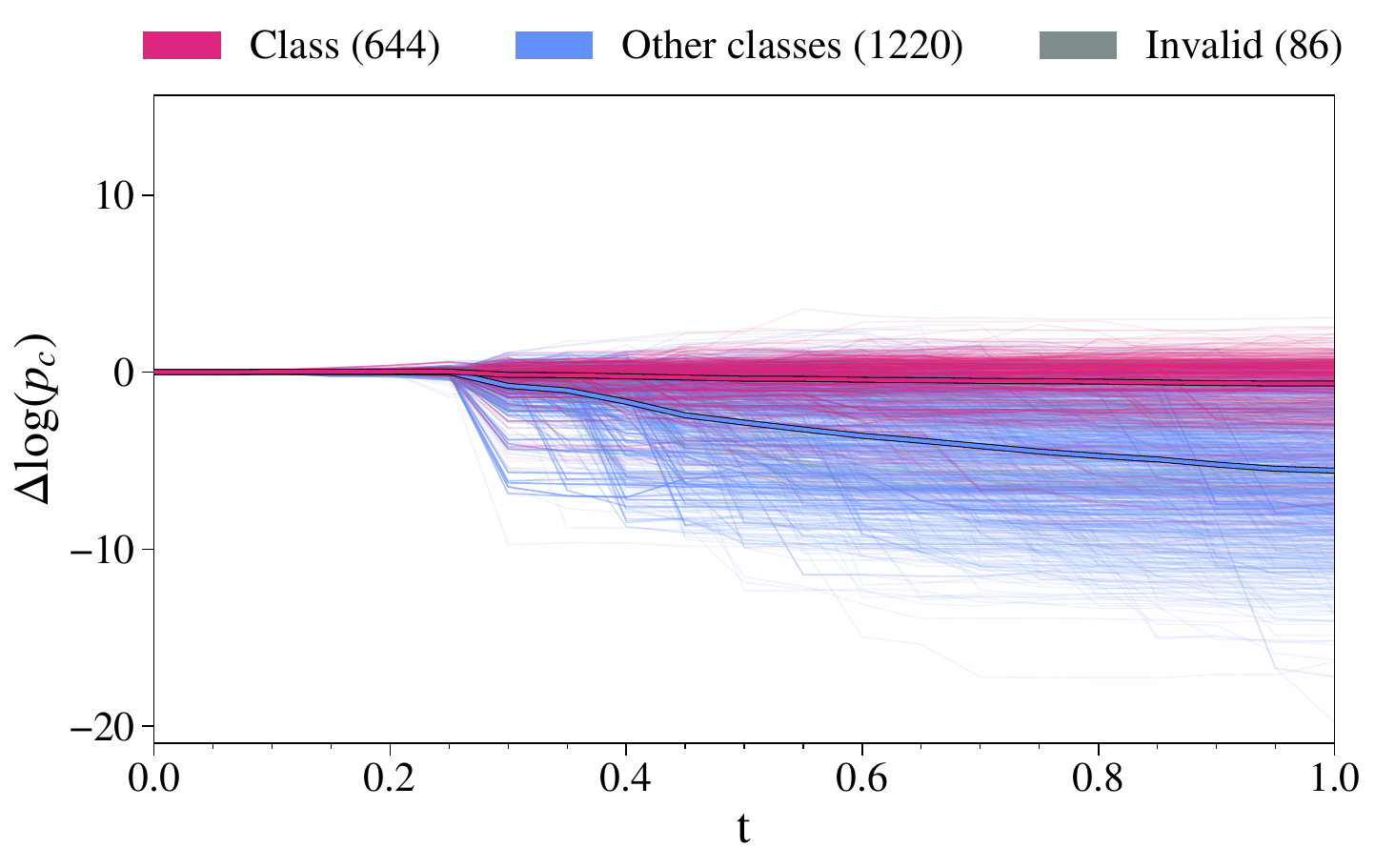}
        \caption{Negative vs neutral}
        \label{subfig:posterior_pos_base}
    \end{subfigure}
    \hfill
    \begin{subfigure}[c]{0.32\textwidth}
        \centering
        \includegraphics[width=\linewidth]{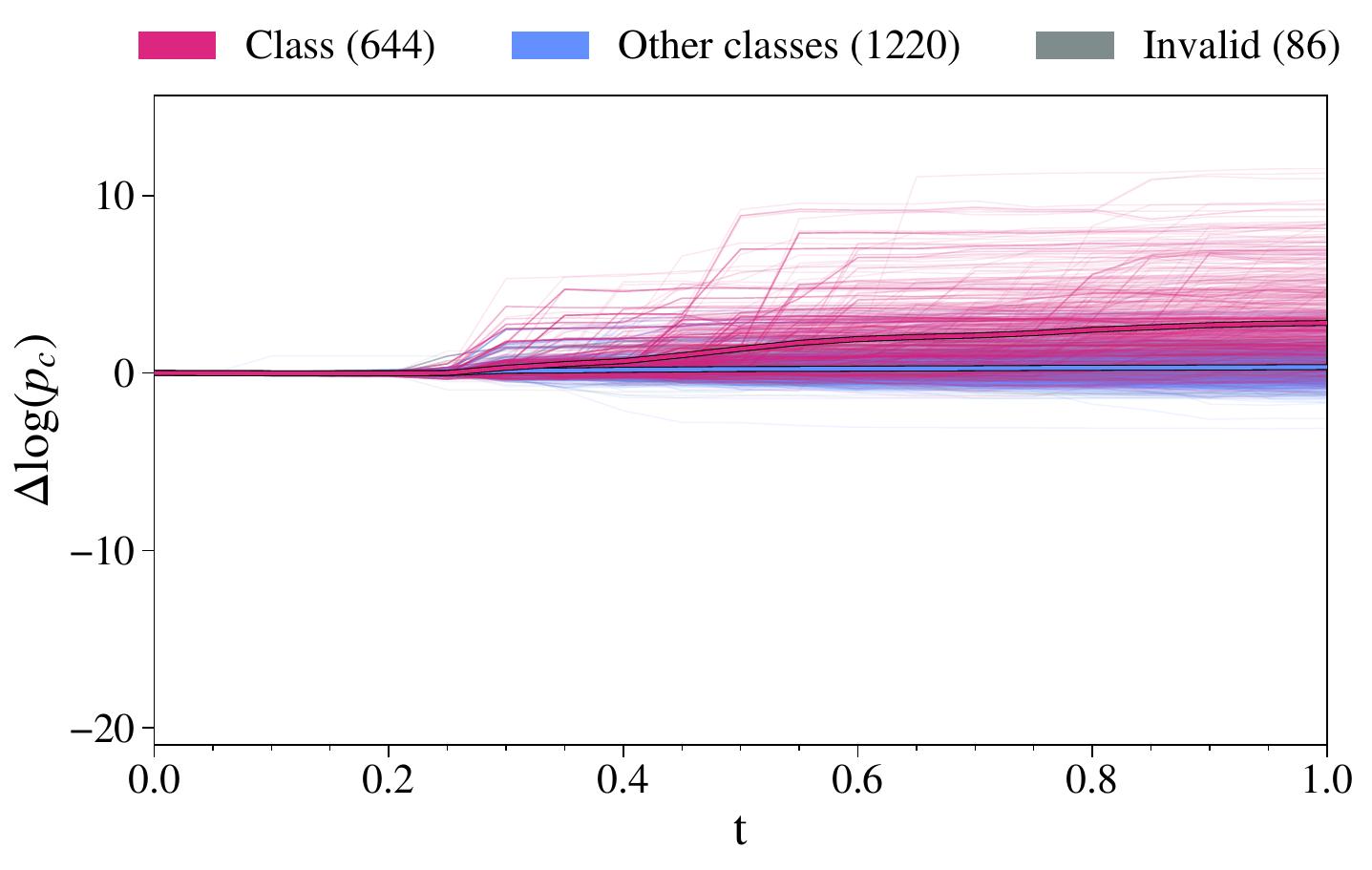}
        \caption{Neutral vs positive}
        \label{subfig:posterior_base_neg}
    \end{subfigure}
    \hfill
    \begin{subfigure}[c]{0.32\textwidth}
        \centering
        \includegraphics[width=\linewidth]{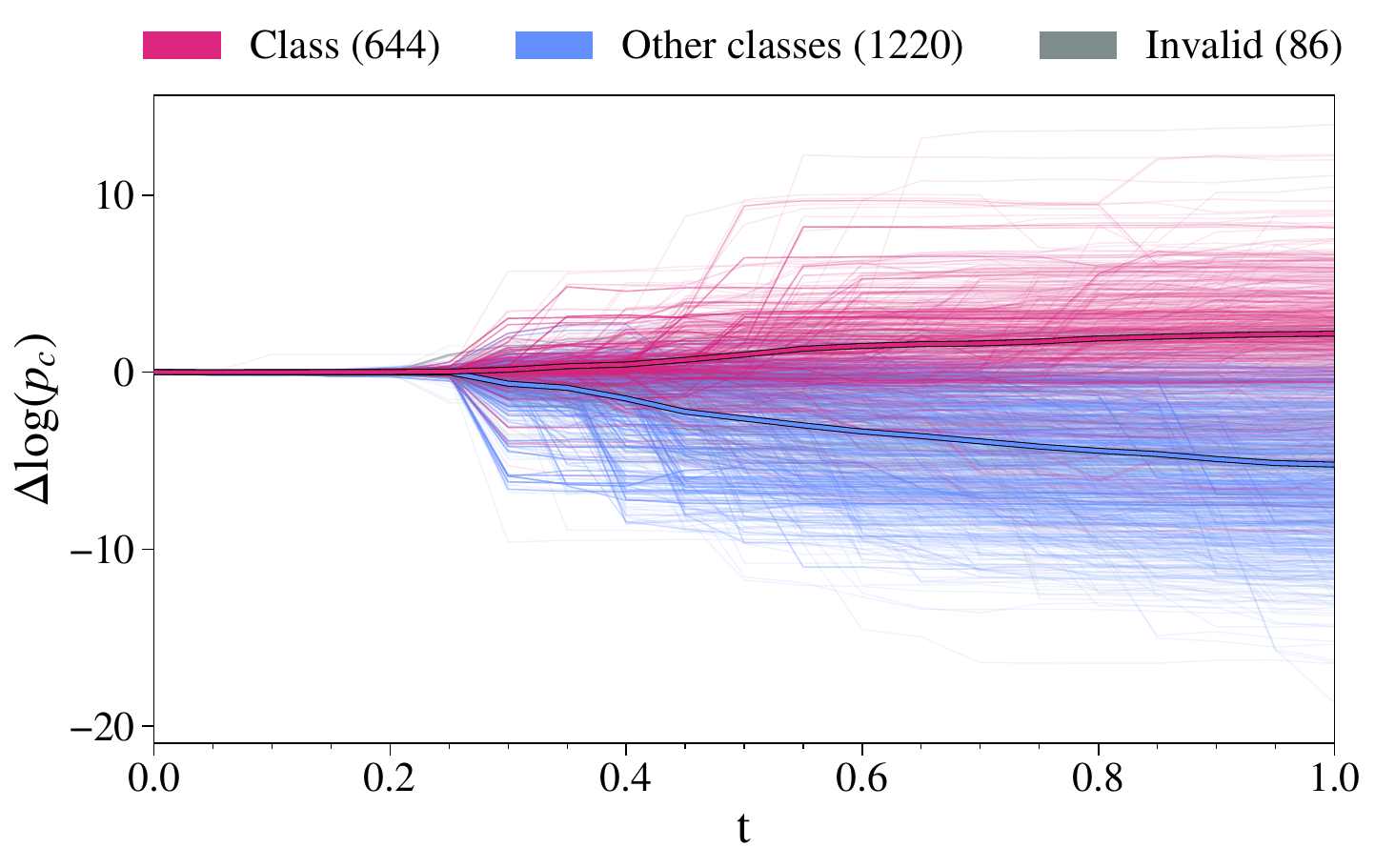}
        \caption{Negative vs positive}
        \label{subfig:posterior_neg_pos}
    \end{subfigure}
    \caption{Posterior trajectories with different prompt types.}
    \label{fig:text_posteriors}
\end{figure}

Our posterior approximation can be applied to any two models that give probabilities on the token being generated. 
This allows for three different couples of models in the case of ``\textsc{not}'' classification: negative vs positive prompt, negative vs neutral and neutral vs positive prompt. 
The posterior approximations for each of these runs are given in \Cref{fig:text_posteriors}. 
We note some interesting behavior, for example in \Cref{subfig:posterior_pos_base}, the posterior approximation does not move much for violated samples, but it goes down much for correct samples. 
We interpret it as follows: when the unconditional model was going to violate anyways, adding the negative part of the prompt does not change much. 
However, if the unconditional prompt wants to generate a correct sample, the negative part of the prompt adds a strong signal. 
Conversely, we see the opposite in \Cref{subfig:posterior_base_neg}, where the violations increase the posterior approximation and the correct samples stay rather flat. 
Adding both signals together by directly comparing negative vs positive, we see in \Cref{subfig:posterior_neg_pos}, that violations go up and correct samples go down. 

\begin{figure}[H]
    \centering

    \begin{subfigure}[c]{0.32\textwidth}
        \centering
        \includegraphics[width=\linewidth]{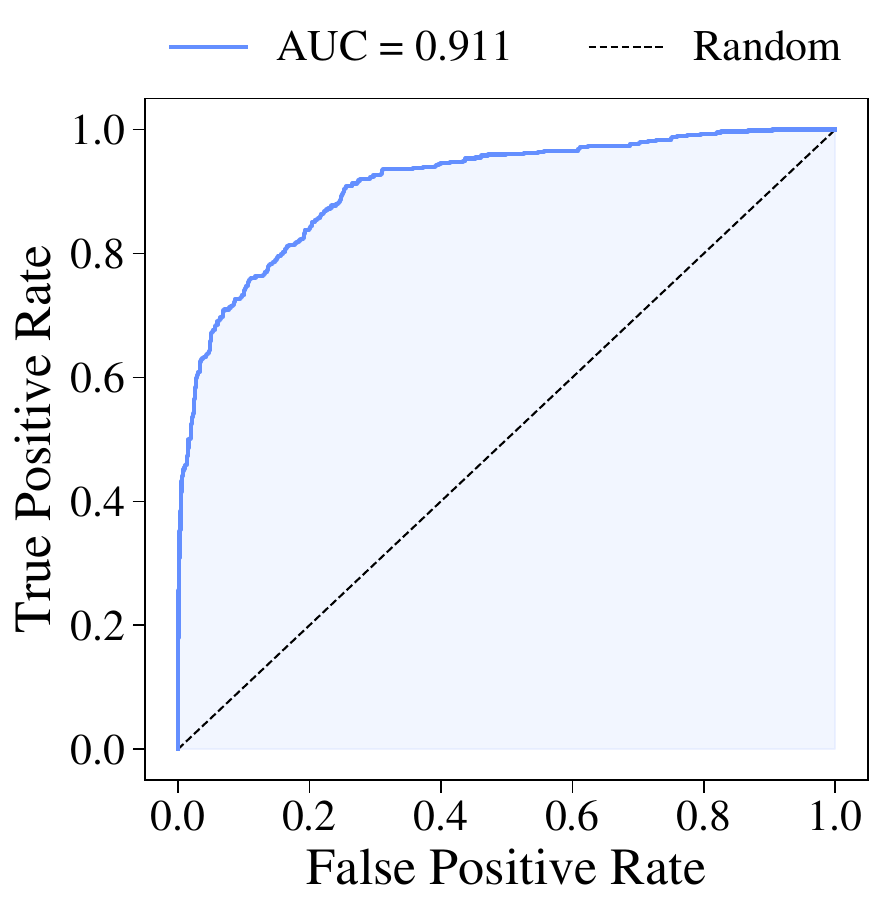}
        \caption{Negative vs neutral}
        \label{subfig:roc_pos_base}
    \end{subfigure}
    \hfill
    \begin{subfigure}[c]{0.32\textwidth}
        \centering
        \includegraphics[width=\linewidth]{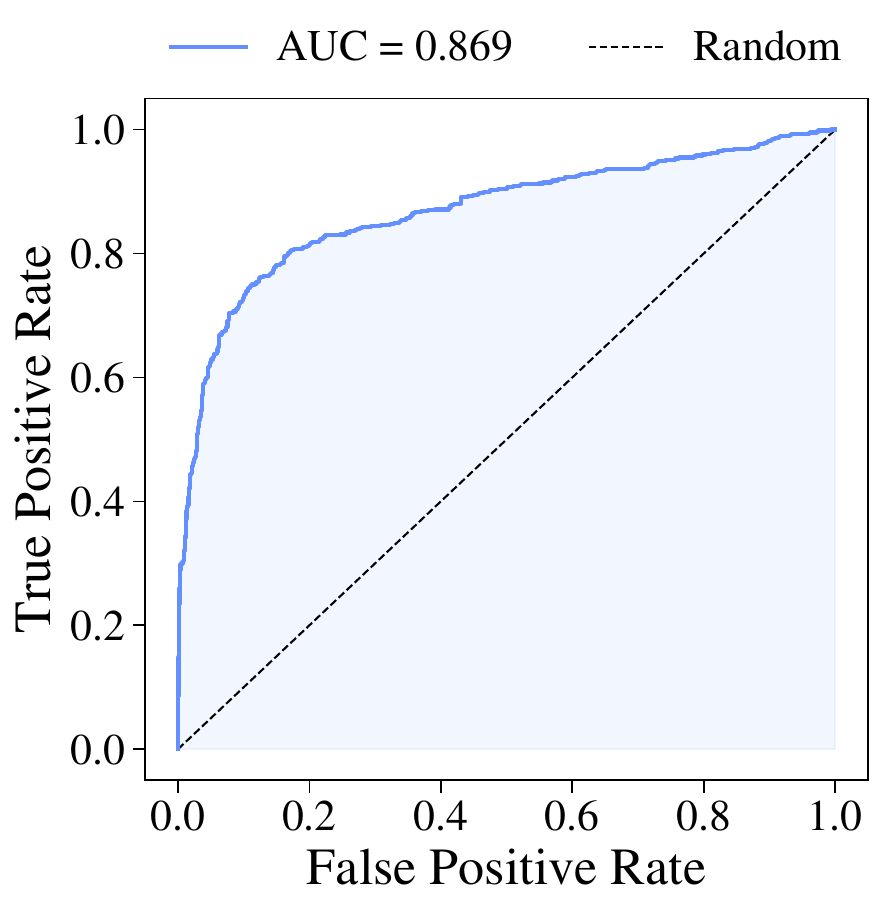}
        \caption{Neutral vs positive}
        \label{subfig:roc_base_neg}
    \end{subfigure}
    \hfill
    \begin{subfigure}[c]{0.32\textwidth}
        \centering
        \includegraphics[width=\linewidth]{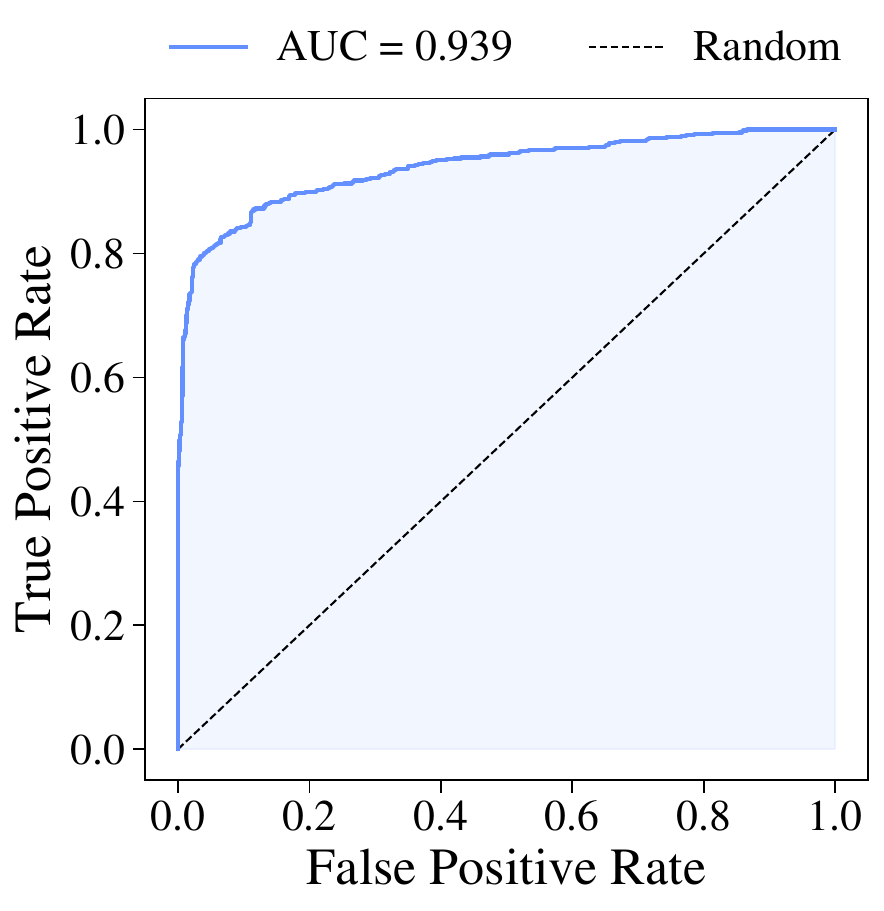}
        \caption{Negative vs positive}
        \label{subfig:roc_neg_pos}
    \end{subfigure}
    \caption{ROC curves for classification of correct vs violated using different prompt types.}
    \label{fig:text_ROCs}
\end{figure}

Do note in the typical \textsc{not} gate context (in other modalities), only the negative vs neutral signal is present and this is also the one used in \textsc{not} gate experiments. 
However, for the sake of finding the best posterior approximation, we find that negative vs positive is best at classifying samples, as shown in \Cref{fig:text_ROCs}. 
In more detail: negative vs positive yields the strongest classifier via the posterior approximation, with an ROC of 0.94. The other approximations perform significantly worse with p=0.05 in a bootstrap test. 
In conclusion, the best posterior approximation for classification is the negative vs positive setup.

\begin{figure}[H]
    \centering
    \includegraphics[width=0.8\linewidth]{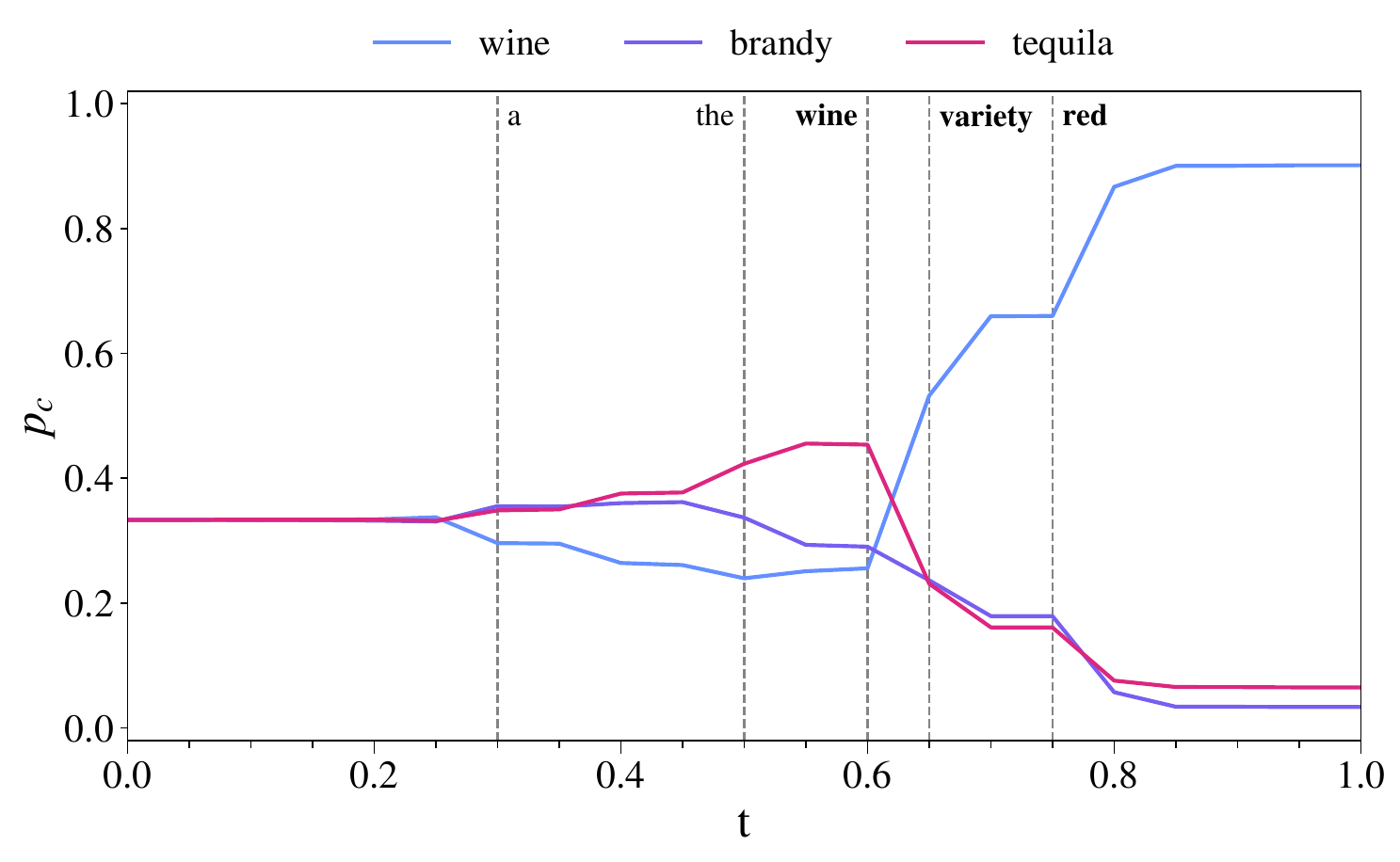}
    \caption{The posterior of multiple classes during generation of the prompt with ID 76 in the development set of Cognac. The answer of the model is ``== Cabernet Sauvignon is a type of red wine made from the Cabernet Sauvignon grape variety. ==''. Some tokens are highlighted with a dashed line at the time step of generation. The blue line, which represents the wine class, clearly jumps on the words ``wine'', ``variety'' and ``red'' being generated. On the other hand, the posteriors do not move much upon ``a'' or ``the'' being generated.}
    \label{fig:posterior_softmax_dp76_wine_timeline}
\end{figure}

While it is hard in free-text generation to define a fixed set of non-overlapping classes, we attempt to construct such a few-class setup, in order to determine the classification accuracy in such tasks. To this end, we take 1000 samples from the unconditional model per prompt and we select four prompts with reasonable variety (i.e., not the same topic being generated 99\% of the time). Then, the majority topics are selected per prompt, and these are defined as the classes. We use the softmax over the posteriors to determine class probability and classify the sample as the highest-probability class. \Cref{fig:posterior_softmax_dp76_wine_timeline} shows a single sample being generated, while \Cref{tab:text_posterior_classification} shows that the posterior estimations yield more than 95\% accuracy at the end of generation, while \Cref{subfig (app): Accuracy} shows the accuracy during generation.

In \Cref{tab:intuitive_token_posterior_full_appendix} we extend the list of intuitive tokens tokens with high impact on the posterior from the main paper.

\begin{table}[H]
    \centering
    \begin{tabular}{c|l|l|c}
    ID & Topic & Classes (\%) & Accuracy \\
    \hline
    76 & alcohol & wine 57\%, brandy 18\%, tequila 26\% & 99.0\% \\
    276 & beverage & wine 55\%, beer 28\%, liqueur 17\% & 95.1\% \\
    353 & liquor & gin 65\%, brandy 23\%, rum 12\% & 95.2\% \\
    418 & domestic animal & cat 16\%, dog 84\% & 95.6\% \\
    \hline
    \end{tabular}
    \caption{Accuracy of the posterior estimation, classifying which (manually defined) subtopic was generated using argmax of the posterior. The IDs refer to the Cognac dataset's development set.}
    \label{tab:text_posterior_classification}
\end{table}

\begin{table}[H]
    \centering
    \begin{tabular}{lll|lc|lc|lc}
    \toprule
    Topic & Sub-topic & & \multicolumn{2}{c|}{1st}
        & \multicolumn{2}{c|}{2nd}
        & \multicolumn{2}{c}{3rd} \\
    \cmidrule(lr){4-5}\cmidrule(lr){6-7}\cmidrule(lr){8-9}
    & (class c) & & Token & $\Delta\log p_c$ & Token & $\Delta\log p_c$
      & Token & $\Delta\log p_c$ \\
    \midrule
    \multirow{2}{*}{sweet} & \multirow{2}{*}{candy} & $+$ &  candy & +2.8 &  \textcolor{gray}{lollip}op & +0.8 &  are & +0.5 \\
     &  & $-$ &  \textcolor{gray}{mar}zip\textcolor{gray}{an} & -2.1 &  cookies & -0.7 &  sweet & -0.6 \\
    \cmidrule(lr){1-9}
    \multirow{2}{*}{dairy product} & \multirow{2}{*}{cheese} & $+$ &  cheddar & +2.4 &  Swiss & +1.4 &  cheese & +0.5 \\
     &  & $-$ &  yogurt & -6.8 &  cream & -0.5 &  product & -0.5 \\
    \cmidrule(lr){1-9}
    \multirow{2}{*}{dairy product} & \multirow{2}{*}{milk} & $+$ &  milk & +0.8 &  with & +0.4 &  whole & +0.4 \\
     &  & $-$ &  whey & -0.7 &  cheese & -0.7 &  liquid & -0.6 \\
    \cmidrule(lr){1-9}
    \multirow{2}{*}{alcohol} & \multirow{2}{*}{wine} & $+$ &  wine & +2.9 &  red & +0.6 &  Cabernet & +0.5 \\
     &  & $-$ &  abs\textcolor{gray}{inthe} & -5.5 &  te\textcolor{gray}{quila} & -4.6 &  \textcolor{gray}{li}que\textcolor{gray}{ur} & -4.2 \\
    \cmidrule(lr){1-9}
    \multirow{2}{*}{beverage} & \multirow{2}{*}{wine} & $+$ &  wine & +2.3 &  \textcolor{gray}{cab}ernet & +1.3 &  red & +1.1 \\
     &  & $-$ &  syrup & -5.5 &  ginger & -3.9 &  espresso & -3.2 \\
    \cmidrule(lr){1-9}
    \multirow{2}{*}{liquor} & \multirow{2}{*}{gin} & $+$ &  gin & +1.5 &  pink \textcolor{gray}{liqueur} & +0.8 &  floral \textcolor{gray}{liqueur} & +0.8 \\
     &  & $-$ &  vodka & -5.3 &  almonds & -4.8 &  coffee & -3.8 \\
    \cmidrule(lr){1-9}
    \multirow{2}{*}{domestic animal} & \multirow{2}{*}{dog} & $+$ &  dog & +4.0 & -sized & +2.0 &  coat & +0.9 \\
     &  & $-$ &  cat & -3.4 &  \textcolor{gray}{co}on & -0.5 &  breed & -0.2 \\
    \bottomrule
    \end{tabular}
    \caption{Tokens pushing most towards ($+$) and away from ($-$) the specified class, as estimated by our posterior estimation. (Sub)words in \textcolor{gray}{gray} are context to put the token (black) in context for the reader. Each of there gray contexts were the actual context for that token for 100\% of the samples with that token.}
    \label{tab:intuitive_token_posterior_full_appendix}
\end{table}

\subsection{Logical operators: \textsc{not} gate}
\label{app:text_not}
\begin{figure}
    \centering
    \includegraphics[width=1\linewidth]{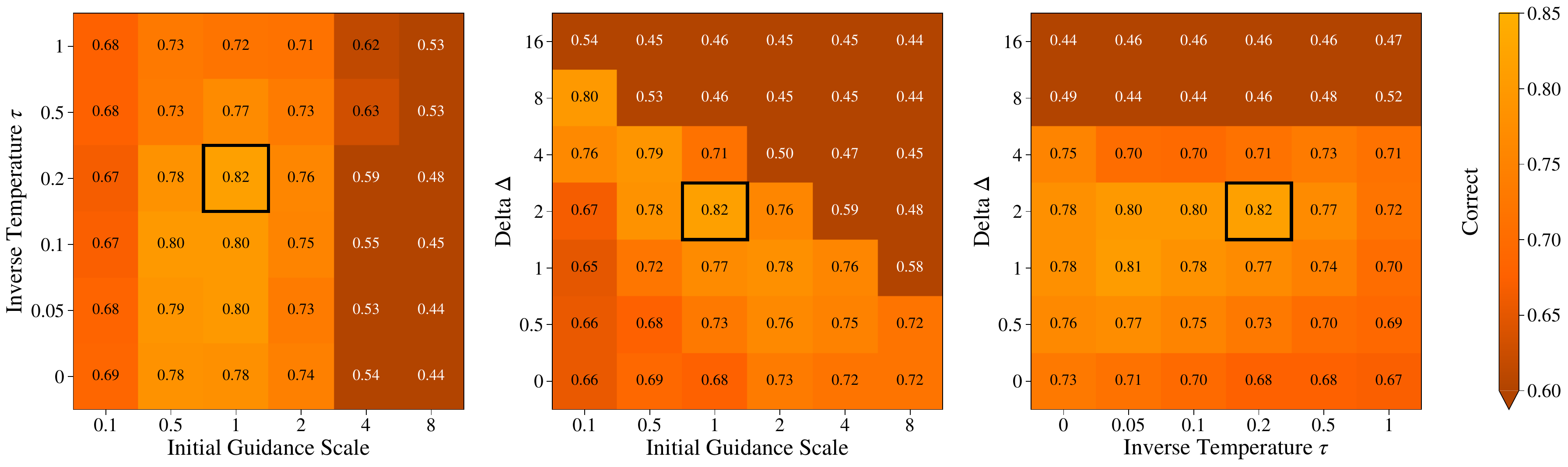}
    \caption{Grid search on the development set of the Cognac benchmark. The optimal hyperparameters that were used for the final experiment are indicated with a bold box. We found that the text modality needs the $\delta$ from \Cref{eq:iter_post_with_hyperparam}. We use $\Delta$ to indicate the cumulative $\delta$ over all steps. The first heatmap shows that $\tau$ is a little less sensitive to tune than the initial guidance scale. The second heatmap shows a downwards diagonal ridge, which seems to indicate that high initial guidance combined with high delta can make the guidance scale explode and yield poor results. The third heatmap again shows less sensitivity for $\tau$, and larger sensitivity for $\Delta$.}
    \label{fig:text_heatmap}
\end{figure}

The results of the \textsc{not} gate in language on the Cognac benchmark are shown in \Cref{tab:cognac_results}. Firstly, note that the unconditional model already answers correctly, i.e., any class other than the undesired class, 70\% of the times. This means that the model never intended to violate the constraint anyway in many cases. With this in mind, the positive prompted model, i.e., the one prompted not to violate the constraint, scores only 79\%. Interestingly, both negative CFG and the \textsc{not} gate yield very similar results. So while the \textsc{not} gate is more grounded in the theory, we would recommend negative CFG to practitioners, as there are not three hyperparameters (grid search shown in \Cref{fig:text_heatmap}) but just one hyperparameter to tune. The results of the \textsc{not} gate are also very similar to the positive prompted model. This validates that the not gate can work as well as prompting the model. In the case of language, this is rather silly, as we can just prompt the model, but this comparison serves more as a proof of concept in the one modality where we can actually compare with a prompted model. So these results are complementary to the other modalities, where simple prompting is not available.

\begin{table}[H]
    \centering
    \begin{tabular}{l|c|c|c}
        Method & Correct (Other classes) & Violated (Class) & Invalid \\
        \hline
        Unconditional & $0.70\,{\scriptstyle\pm0.03}$ & $0.23\,{\scriptstyle\pm0.03}$ & $0.07\,{\scriptstyle\pm0.02}$ \\
        Prompted ``\textsc{not}'' & $0.79\,{\scriptstyle\pm0.03}$ & $0.10\,{\scriptstyle\pm0.02}$ & $0.11\,{\scriptstyle\pm0.02}$ \\
        Negative CFG (logit-space) & $0.81\,{\scriptstyle\pm0.02}$ & $0.06\,{\scriptstyle\pm0.01}$ & $0.13\,{\scriptstyle\pm0.02}$ \\
        \textsc{not} gate (prob-space) & $0.79\,{\scriptstyle\pm0.03}$ & $0.08\,{\scriptstyle\pm0.02}$ & $0.14\,{\scriptstyle\pm0.02}$ \\
         \hline
    \end{tabular}
    \caption{Results of the \textsc{not} on the Cognac benchmark}
    \label{tab:cognac_results}
\end{table}

\section{Additional details regarding antibody sequence generation}
\label{app: antibody}
In this section additional details and results on antibody sequence generation are provided. First, \cref{app: antibody - background} introduces the task and the dataset for readers without a background in biology. Second, \cref{app: antibody - training details} describes the architecture and the training hyperparameters. Third, \cref{app: antibody - ablations} presents ablations on the model performance and introduces the metrics used. Finally, \cref{app: antibody - logical gates} covers the posterior estimation and the results obtained with the \textsc{not} gate. The structure deliberately mirrors that of the appendix on images (Sections~\ref{app: image - training details} and~\ref{app: image - model ablations}), which precedes this one, so that a reader familiar with it can focus on the differences, mainly the metrics used to evaluate the generated sequences.
\subsection{Dataset and background}
\label{app: antibody - background}
This work is aimed at a broad machine learning audience, which might not be familiar with proteins, let alone antibodies. In this section we therefore give the reader the minimum needed to understand the specificities of the task. Interested readers are referred to the protein language model literature, for instance \citet{ESM2, AMPLIFY}, and to \citet{IgLM} for antibodies in particular.

\paragraph{Antibodies.} Antibodies are Y-shaped proteins that are a key part of our immune system. Like every protein, an antibody is defined by its sequence of amino acids, of which 20 are commonly used and each is denoted by a letter. An antibody consists of two identical heavy chains and two identical light chains. At the tip of each chain sits a variable domain of roughly 120 residues, which recognises the target of the antibody, the antigen. This domain is built from four framework regions (FR1-4), which form a conserved structural scaffold, interleaved with three complementarity-determining regions (CDR1-3). The CDRs are hypervariable loops that make direct contact with the antigen and therefore determine what the antibody binds to. The CDR3 of the heavy chain (CDRH3) is the most diverse of them. During an immune response, the CDRs are fine-tuned towards the antigen in a process of somatic hypermutation and selection called affinity maturation.

\paragraph{V genes.} The variable domain of the heavy chain is encoded by recombining a V, a D and a J gene segment. The V gene encodes the largest part of the domain, namely the framework regions FR1-3 and CDR1-2, whereas CDR3 lies at the junction of the three segments and is therefore the most variable. The human heavy-chain V genes are commonly grouped into seven families, IGHV1 to IGHV7, whose members are highly similar in sequence. As the germline-encoded parts of a sequence are only mildly altered by mutations, the family of an antibody can be determined very reliably by aligning it to a database of germline genes. This makes the V-gene family a convenient class label for a conditional generative model: it is discrete and it can be assigned to every sequence. It should also be noted that the distribution of V-genes in humans is highly unbalanced, with IGHV3 accounting for 44\% of human antibodies, whereas IGHV6 and IGHV7 make up only about 0.6\% and 0.5\%, respectively (see \Cref{table: antibody classes}).

\paragraph{Dataset.} The Observed Antibody Space (OAS) \citep{OAS} is the largest and most widely used database of antibody sequences. As our goal is not to obtain a state-of-the-art antibody model, but to demonstrate the flexibility of our approach, we use an older partition of the database, namely the OAS95 set of 559M sequences clustered at 95\% identity, that was used to train IgLM \citep{IgLM}. IgLM is an autoregressive antibody language model, which we use to evaluate whether the generated sequences look reasonable (\cref{app: antibody - ablations}). To verify our theoretical claims, however, a conditional dataset is required. We therefore keep the human heavy chains of OAS95 that are complete variable domains (starting with the typical first residues of FR1 and without ambiguous residues), and classify them by V-gene family using the publicly available IgBLAST tool \citep{IgBLAST} with the IMGT database of human germline genes \citep{IMGT}. Exact duplicates are removed. This leaves 55.4M sequences, of which 2\% are held out for validation. The sequences are 122 residues long on average, and 99\% of them are between 113 and 132 residues long. As the family is mainly determined by the germline-encoded regions, this classification is essentially unambiguous.
\begin{table}[H]
\begin{center}
\begin{tabular}{|l|ccccccc|}
\hline
V-gene family & IGHV1 & IGHV2 & IGHV3 & IGHV4 & IGHV5 & IGHV6 & IGHV7 \\
Share of sequences (\%) & 24.9 & 2.5 & 44.1 & 25.3 & 2.2 & 0.6 & 0.5 \\
\hline
\end{tabular}
\end{center}
\caption{Class balance of the antibody dataset (human heavy chains of OAS95, measured on the held-out validation split of 1.1M sequences). The conditional model is trained on these seven classes.}
\label{table: antibody classes}
\end{table}
\subsection{Training details}
\label{app: antibody - training details}
\paragraph{Architecture.} The antibody models are based on the AMPLIFY protein language model \citep{AMPLIFY}, a bidirectional Transformer encoder with rotary position embeddings and SwiGLU feed-forward layers. As for images, we add adaptive layer normalization (AdaLN-Zero) as conditioning mechanism \citep{DiT}: the noise level $t$ and the V-gene family are embedded and summed into a single conditioning vector that modulates every block, where the label embedding has an additional entry for the unconditional case. The vocabulary has 27 tokens: 21 residue tokens (the 20 amino acids and the ambiguity code B) and 6 special tokens, of which one is the mask token used for masking diffusion. Every sequence is enclosed by a begin and an end token and padded to a fixed length of 256 tokens, well above the longest sequences. These special tokens are never corrupted and are excluded from the loss. As a consequence, the model does not have to determine the length of a sequence itself: when sampling, the padding pattern of a randomly chosen real training sequence is used as starting point, so that the lengths follow the empirical length distribution independently of the class. The resulting models have 50M parameters, of which 33.6M belong to the AMPLIFY backbone and 16.6M to the AdaLN modulation layers.

\paragraph{Training.} As for images, the model predicts the clean sequence with a cross-entropy loss, which for masking is computed on the masked residues only and for uniform noise on all residues. We use a linear noise schedule with stratified sampling of the noise level. Conditional models are trained with label dropout, so that the same network also acts as unconditional model when the label is replaced by the unconditional token. In contrast to the images, no separately trained unconditional model is used, and no exponential moving average of the weights is kept. The hyperparameters in \Cref{table: antibody training hyperparameters} were not tuned per noise type. Apart from the noise type, the masking and uniform models only differ in their number of training steps, 200k and 400k, which corresponds to roughly 0.9 and 1.9 epochs of the 54.3M training sequences.
\begin{table}[H]
\begin{center}
\begin{tabular}{|l|c|}
\hline
Architecture & AMPLIFY with AdaLN-Zero \\
Parameters & 50M \\
Depth / hidden size / heads & 12 / 480 / 20 \\
Sequence length / vocabulary & 256 / 27 (incl.\ mask token) \\
\hline
Optimizer & AdamW \\
Learning rate & $1\cdot10^{-4}$ \\
Betas / weight decay & (0.9, 0.95) / 0.01 \\
Warmup steps / schedule & 1000 / cosine decay \\
Batch size & 256 (2 GPUs $\times$ 128) \\
Training steps & 200k (masking) / 400k (uniform) \\
Gradient clipping & 1.0 \\
Numerical precision & bf16 \\
EMA & none \\
Label dropout & 0.15 \\
\hline
Noise schedule & linear \\
Noise level sampling & stratified in $[0,1]$ \\
Loss weighting & none \\
\hline
\end{tabular}
\end{center}
\caption{Hyperparameters of the antibody models.}
\label{table: antibody training hyperparameters}
\end{table}

\begin{figure}[H]
    \centering
    \includegraphics[width=0.9\linewidth]{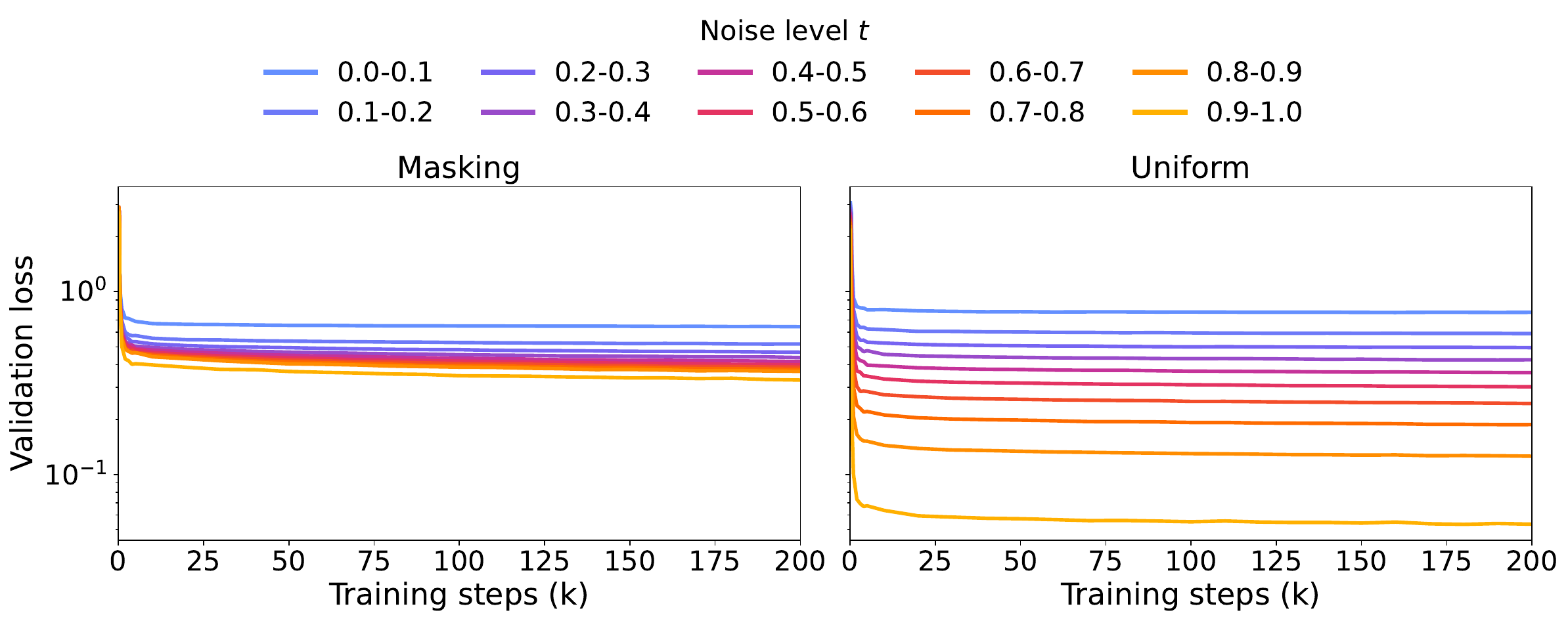}
    \caption{Validation loss during training of the antibody models, averaged over bins of noise levels $t\in[0,0.1],[0.1,0.2],\dots$, for masking (left) and uniform noise (right). 
    }
    \label{fig (app): antibody training curves}
\end{figure}

\paragraph{Training curves.} 
Training progress is shown per noise level in \Cref{fig (app): antibody training curves}. As expected, lower noise levels (closer to $t=1$) show a lower loss, particularly with uniform noise. This is beause when training a masked model the loss is only computed on the still masked tokes, while for the uniform model the loss is always computed on all tokens. In the latter, most tokens at low noise levels are already correct and should simply be returned back as output, a luxury the masking model can never have. This distinction also renders the comparison of loss curves between masking and uniform cases irrelevant. An interesting note in this setting is that the loss decreases very rapidly in the beginning and then only progresses very slowly. We hypothesize that this is due to the dual nature of antibody sequences, very easy framework regions, which are learnt fast and then highly unpredictable CDRs on which progress is slow. For the V-gene conditional task discussed in this paper the first part is most relevant, which is why our models are never evaluated on CDRs.
\subsection{Model ablations}
\label{app: antibody - ablations}
\begin{figure}[H]
    \centering
    \includegraphics[width=\linewidth]{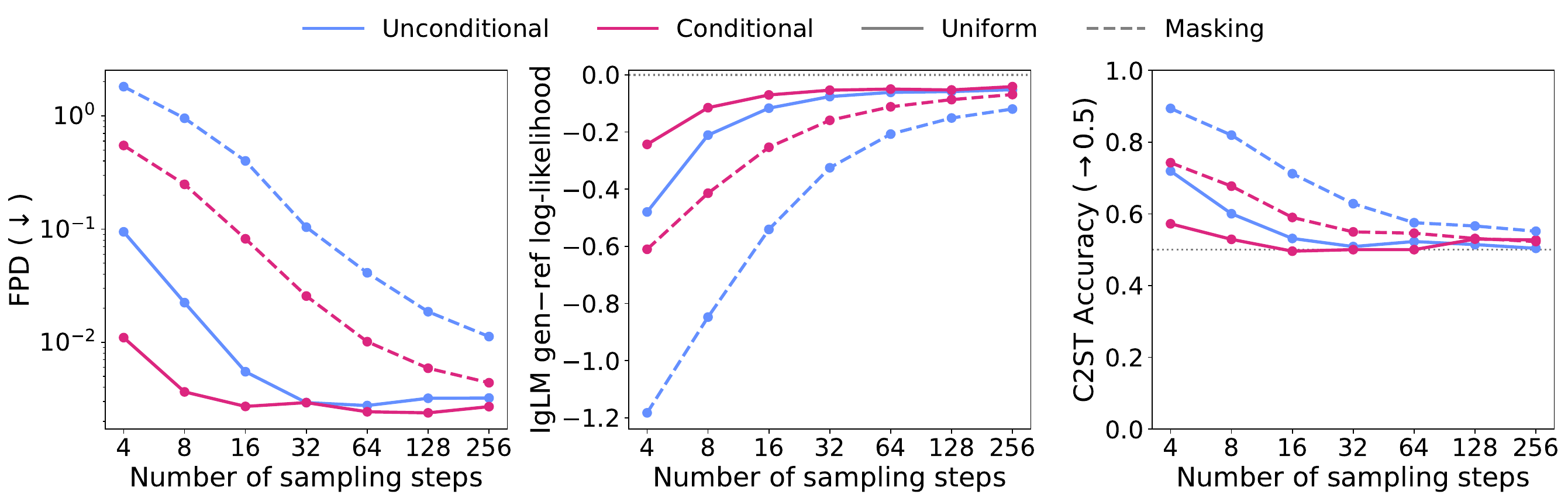}
    \caption{Sweep over the number of sampling steps for the uniform-noise antibody model, for the unconditional (label dropped) and the conditional model, with a linear sampling schedule and 4096 sequences per point. The uniform model in full and the masking model in dashed lines. Shown are, from left to right, the FPD, the gap between the IgLM log-likelihood of the generated and of real sequences, and the C2ST accuracy. Class labels for the conditional model are drawn from the empirical class frequencies. 
    }
    \label{fig (app): antibody ablation num steps}
\end{figure}

\paragraph{Metrics.} The image metrics (FID, precision, recall and C2ST) rely on features of a pretrained image classifier. For antibodies we replace them by their protein counterparts, and add a measure of how natural the sequences are. All metrics are computed on 4096 generated sequences, whose class labels are drawn from the empirical class frequencies of the training set (\Cref{table: antibody classes}), such that rare families remain rare, and whose lengths are set as explained above.
\begin{itemize}
    \item \textbf{IgLM log-likelihood.} IgLM \citep{IgLM} is an autoregressive language model trained on natural antibodies, conditioned on the chain type and the species (here heavy and human). We report the average log-likelihood per residue of the generated sequences under IgLM, which measures how antibody-like they are (higher is better). For reference, real training sequences reach $-0.57$ and random amino acid sequences of the same lengths $-3.80$. \Cref{fig (app): antibody ablation num steps} shows the gap between generated and real sequences, which should go to zero.
    \item \textbf{FPD.} The Fréchet Protein Distance is the protein counterpart of the FID: the Fréchet distance between Gaussian fits to the ESM-2 embeddings \citep{ESM2} of generated and real sequences. It measures quality and diversity together (lower is better). We compare to 20000 random training sequences.
    \item \textbf{C2ST.} The classifier two-sample test is the held-out accuracy of a logistic regression that separates real from generated embeddings (4096 of each, 80/20 split), where 0.5 is the best possible value, as then generated samples are indistinguishable from real ones. 
\end{itemize}
Additionally, we report the conditioning accuracy: the fraction of generated sequences that IgBLAST, the same tool that produced the training labels, assigns to the requested family. This is the counterpart of the classifier-based accuracy used for images.

These metrics are computed for both noise types as a function of the number of sampling steps. This is shown in \Cref{fig (app): antibody ablation num steps} where the unconditional model is compared to a conditional model with class labels from the true data priors. As expected the quality of all samples increases as the number of inference steps increases. At low number of sampling steps the conditional model outperforms the unconditional one, which can be attributed to the less diverse distribution the model is sampling from. In essence the conditioning signal is often a nearly deterministic signal for portions of the sequence, by using a conditional model, these sections will never be confused, which is not true for the unconditional model, particularly if multiple tokens are flipped simultaneously. Another interesting note is that all metrics seem to saturate at 32 steps, again indicating the strong determinism of the task at hand. In this regime, at every step around 4 tokens are flipped per step and still this is sufficient for perfect samples.  

All results are shown for the uniform-noise model, sampled with the linear schedule that was also used for training.

\subsection{Posterior estimation}
\label{app: antibody - posterior estimation} 
In this context the V-gene classes are so well separated that the online posterior's task is relatively easy, a few token flips is usually enough to determine the class fully. This is highlighted by the perfect accuracy and AUC scores the method obtains, not only at the end point, but very early on \Cref{fig (app): posterior estimation metrics}. We see this as a significant proof, that when a strong classification signal is available, the proposed online posterior distinguishes it reliably. 
\subsection{Logical operators: \textsc{not} gate}
\label{app: antibody - logical gates}
\paragraph{Metrics} For the \textsc{not} gate we use similar metrics to those used to perform the ablations on the model in \Cref{app: antibody - ablations}. In particular, the IgLM log-likelihood is used as a quality metric to verify sequence quality on an individual level. To measure distributional coverage, the FPD is again used with ESM2 embeddings, this time computed between generated samples and reference data which do not include the undesired class. The idea is that should a sequence of the unwanted class be generated, the FPD measured with respect to that set should increase as well. Further, as always, we measure safety or removal capacity of the method with a simple violation rate. This value is obtained by classifying samples using IgBLAST and computing the fraction of samples from the undesired class. All metrics are computed on 4096 samples.

\paragraph{Methods} To highlight the severe class imbalances present in the data, samples are first generated unconditionally. This gives a first worst-case scenario baseline. Then similarly to the image case, a static logit negative guidance version implemented. This is essentially standard discrete CFG with a negative guidance scale, that therefore pushes away from a certain condition instead of reinforcing it. Finally, the dynamic negative guidance approach, i.e. the \textsc{not} gate, is ablated. A key point of this method is that a properly scaled prior-weighted initial guidance scale is chosen. In other words, for a single hyperparameter setting a majority class receives more guidance than a minority class.

\paragraph{Challenge} Due to the severe class imbalance present in the data it should be intuitively clear that finding one setting that works best for all cases is no easy task. Majority class simply require much more guidance to be removed properly when compared to minority classes. This is particularly true in static approaches, such as standard CFG, where a fixed guidance scale is chosen for all classes regardless of their prevalence. Our dynamic approach and its prior-weighted initial guidance scale does not suffer the same flaw. By design the guidance scale start off at a multiple its theoretical value, i.e., at $r\frac{p(c_i)}{1-p(c_i)}$ with the factor $r$ defined by the value of the initial guidance scale $s_0$ through $r=(N-1)s_0$, see \Cref{app: hyperparameters posterior} for more details. This means that a single choice of hyperparameter will automatically apply more guidance to majority classes than to minority ones simply because the initial value is prior-weighted. This is an extra quality of our \textsc{not} gate. A single hyperparameter setting can be set, and work equally well at removing any class regardless of how imbalanced the dataset really is.

\begin{figure}
    \centering
    \includegraphics[width=0.9\linewidth]{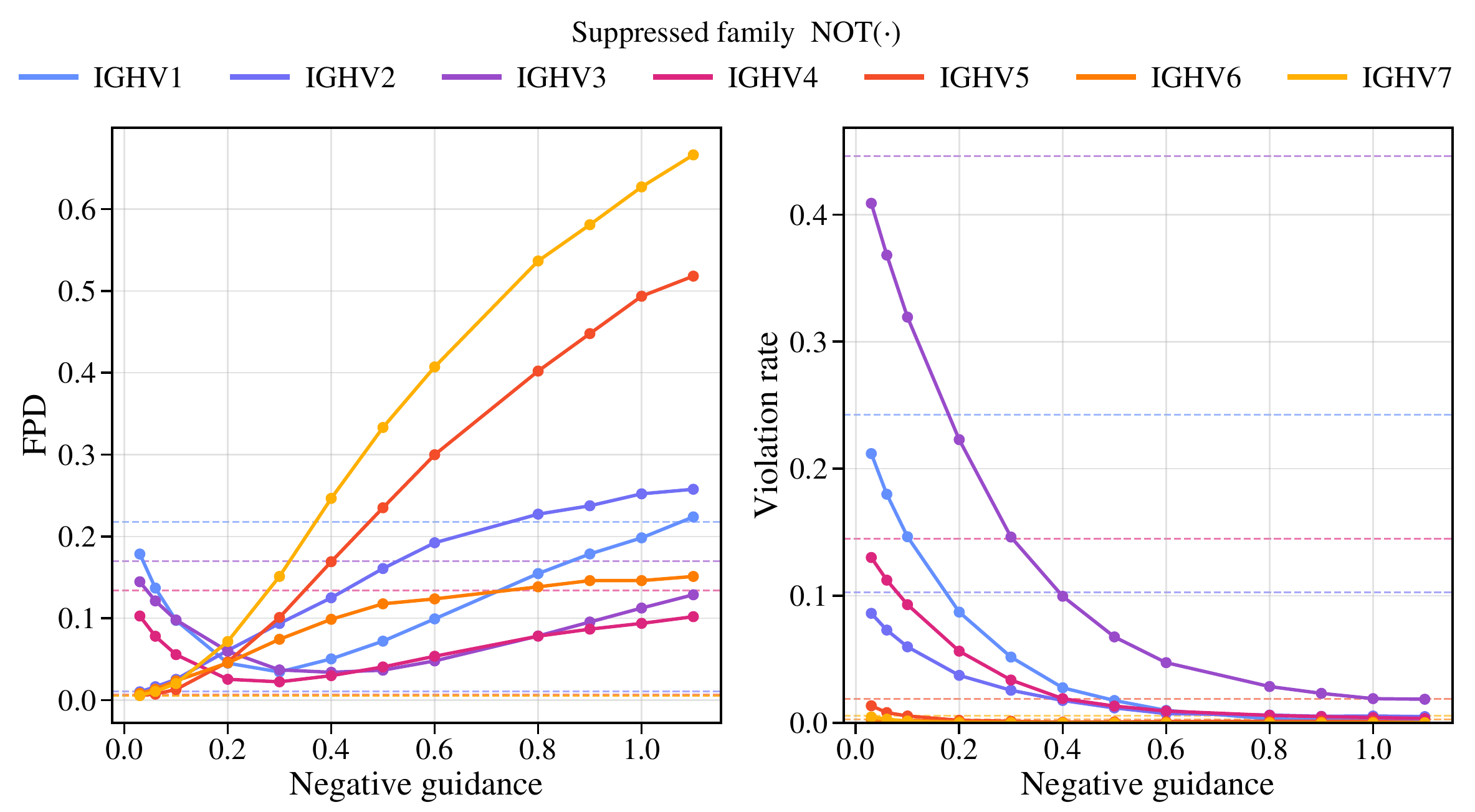}
    \caption{Negative CFG with static negative logit guidance. On the left, the Frechet Protein Distance measuring distributional coverage and sample quality, and on the right, the violation rate. Notice the strong differences between majority (IGHV1-4) and minority classes (IGHV5-7).}
    \label{fig (app): antibody - static baselie}
\end{figure}

\paragraph{Results} 
As outlined above, finding a single static guidance scale for  negative CFG, that is able to remove all classes regardless of their prevalence is impossible. This is shown in \Cref{fig (app): antibody - static baselie}. On the right the violation rate, decreasing for all classes as the negative guidance scale is increased, is shown. On the left, the FPD is shown, here a very different story emerges. Majority classes follow the usual trade-off, first decreasing and then increasing indicating that guidance is starting to become too strong. For the minority on the other hand every guidance scale value seems to be too strong in terms of FPD. This makes sense: the FPD is a distributional metric and in this static case guidance affects all trajectories just as much, and does not only apply to the targeted trajectories. One cannot hope the method to require the same hyperparameter for minority and majority classes.

\begin{figure}
    \centering
    \includegraphics[width=0.9\linewidth]{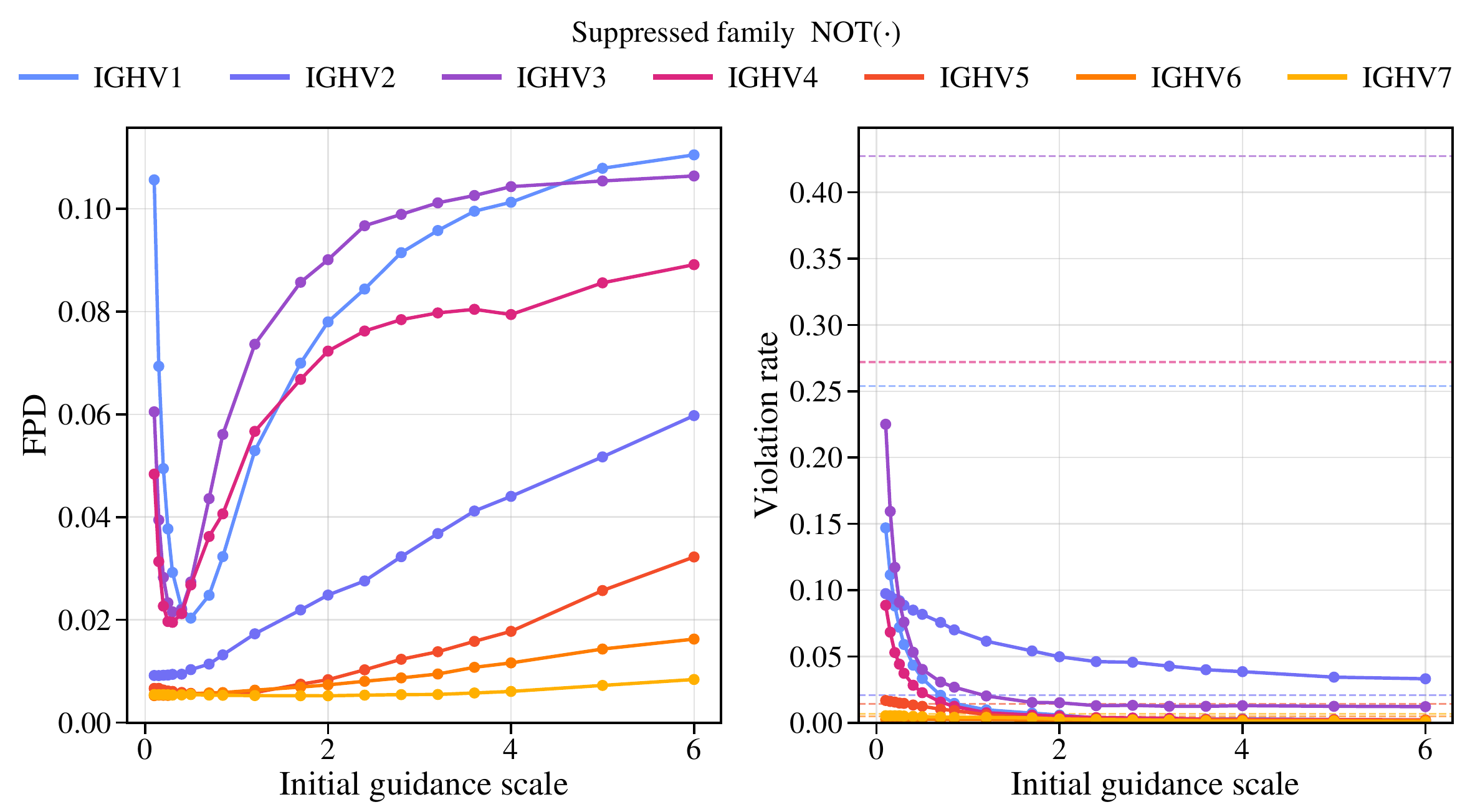}
    \caption{\textsc{not} gate with dynamic negative guidance. On the left, the Frechet Protein Distance measuring distributional coverage and sample quality, and on the rate the violation rate. Here, thanks to the prior weighted initial guidance scale, minority classes are much less harmed when applying guidance.}
    \label{fig (app): antibody - dynamic guidance}
\end{figure}

On the other hand, our \textsc{not} gate with dynamic negative guidance, that has a prior-weighted initial guidance is shown in \Cref{fig (app): antibody - dynamic guidance}. Here, we again can observe that the violation rate goes down as more guidance is applied, but crucially because guidance is not applied equally across classes, minority classes suffer much less from guidance. We view this prior weighted initial point as a key advantage of our theoretically grounded approach. 

\begin{figure}
    \centering
    \includegraphics[width=0.9\linewidth]{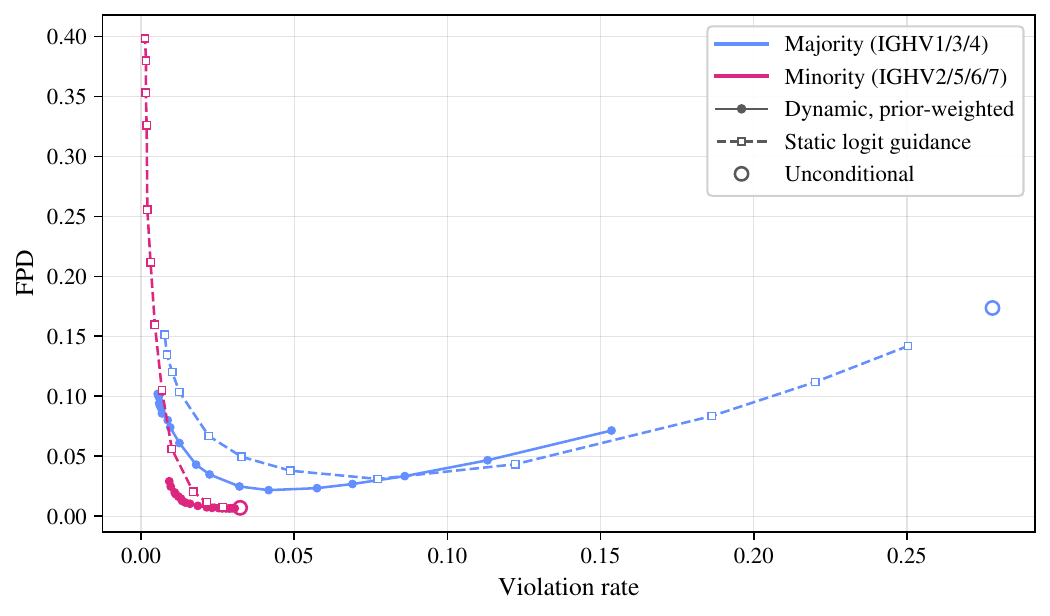}
    \caption{Pareto front between quality (FPD) and safety (violation rate) for majority and minority classes. Negative CFG with static logit guidance is compared to the \textsc{not} gate with dynamic negative guidance.}
    \label{fig (app): antibody - paretto}
\end{figure}

Similarly to how the two are compared in images, a Pareto curve comparing safety (violation rate) and diversity/quality (FPD) is shown in \Cref{fig (app): antibody - paretto}. To highlight the differences between majority and minority classes the two Pareto fronts are plotted. Similarly to what was obtained on images, the dynamic approach outperforms the static one. Even more so if a single global setting (one fixed set of hyperparameters) is desired.


\section{Posterior estimation hyperparameters}
\label{app: hyperparameters posterior}
The posterior estimation, when used as formulated by the theory, tends to be stochastic and overconfident, which is why prior work in the continuous domain introduced the two regularizing hyperparameters $\tau$ and $\delta$~\citep{DNG}. We also use an initial guidance scale to determine the strength of guidance with which generation starts.

By setting $\tau < 1$ in \Cref{eq:iter_post_with_hyperparam}, strong jumps in the posterior evaluation are avoided. All grid searches across modalities resulted in a value of $\tau$ that satisfies this intuition. 

The second hyperparameter $\delta$ gives a constant push upwards to the posterior. This is most important when interfering during generation with the \textsc{not} gate, as there the precise calibration of the posterior value is key in contrast to the AUC/Accuracy experiments that are mainly impacted by the correct ranking of classes. In practice we find that the posterior most often decreases, particularly for the \textsc{not} gate, this is shown for text in \Cref{subfig:posterior_pos_base}, even before guidance and \Cref{fig: mnist NOT guidance profile} after guidance for images. We argue this is because at best the conditional model agrees with the neutral model, yielding an update $\log p(x_t\mid x_{t-1},c) - \log p(x_t\mid x_{t-1})$ around 0. At worst the models disagree and we pick the token that the unconditional agrees most with, which means $\log p(x_t\mid x_{t-1})$ is relatively high and the difference will be strongly negative. So the posterior going up is a very rare case in practice, though it is clear intuitively that when the conditional model agrees, the posterior is going up. Thus, the $\delta$ gives a permanent upward push to actually increase the posterior when the models agree. 

The third hyperparameter is the initial guidance scale $s_0$, that works on the prior probability $p(c\mid x_{0})$, which is again important for the \textsc{not} gate. The idea of the initial guidance scale is to control the baseline strength of guidance, with three different cases merged in one formula. For this, we scale the priors in guidance scale-space, as if they were uniform: The guidance scale for a uniform prior would start at $\frac{p(c_i)}{1-p(c_i)}$, which is $\frac{1}{N-1}$, and thus we would need a scaling factor r of $s_0$$ \cdot (N-1)$ in order to achieve initial guidance of $s_0$. For all three cases, we take the real prior, convert it into guidance scale-space with $\frac{p(c_i)}{1-p(c_i)}$, multiply with r and invert back to probability-space. The first case is that of uniform priors, such as MNIST. In that case, the whole formula just boils down do using exactly $s_0$ as the strength for all classes. In the second case, we consider the prior to be unknown, for example in the text modality. As the prior is unknown and we want to start with a fixed guidance strength, the approach reduces to the uniform case with a flat guidance scale $s_0$. The third case is that of known but non-uniform priors. There, we again convert the prior to the guidance scale-space: $\frac{p(c_i)}{1-p(c_i)}$, we scale it by a factor r and then invert back into probability space. This way we scale all guidance strength equally, and the strength per class is still informed by its prior, rather than using a flat scale. We chose this parametrization as we found it easy to work with and sweep, but there is no fundamental reason to choose this parametrization over another.

\section{Reproduction details}
\label{app:reproduction}
We make our code available on 
\hyperlink{https://github.com/Gianni-Van-de-Velde/Online-Class-Posteriors-for-Interpretable-and-Controllable-Discrete-Diffusion}{https://github.com/Gianni-Van-de-Velde/Online-Class-Posteriors-for-Interpretable-and-Controllable-Discrete-Diffusion}. 
The current document is still under review and, as a consequence, both the text as the code are prone to minor chanes. 

All text and image experiments ran on NVIDIA A40 GPUs and 4 virtual cores of a Intel Xeon Silver 4310 CPU. For antibodies the experiments where performed on NVIDIA GeForce RTX 4090 GPUs.

\end{document}